\RequirePackage{fix-cm}
\documentclass[10pt,twocolumn,letterpaper]{article}
\usepackage[pagenumbers]{cvpr}
\usepackage{float}
\usepackage{placeins}
\usepackage{stfloats}
\usepackage[T1]{fontenc}
\usepackage[utf8]{inputenc}
\usepackage{colortbl}
\usepackage{multirow}
\definecolor{trainoverlap}{gray}{0.62}
\definecolor{gainrow}{HTML}{EAF4F7}
\newcommand{\tableevalmark}{\smash[t]{$^{\dagger}$}}
\makeatletter
\newcommand{\gainentry}[1]{%
  \raisebox{\dimexpr(\ht\@arstrutbox-\dp\@arstrutbox-\height+\depth)/2\relax}{\bfseries\boldmath#1}%
}
\newcommand{\resacell}[1]{%
  \begingroup
  \let\resa@originalunderline\underline
  \renewcommand{\underline}[1]{%
    \vphantom{##1}\smash{\resa@originalunderline{##1}}%
  }%
  \raisebox{\dimexpr(\ht\@arstrutbox-\dp\@arstrutbox-\height+\depth)/2\relax}{#1}%
  \endgroup
}
\makeatother
\definecolor{cvprblue}{rgb}{0.21,0.49,0.74}
\usepackage[pagebackref,breaklinks,colorlinks,allcolors=cvprblue]{hyperref}
\newcommand{\method}{ReSA\xspace}
\newcommand{\fg}{\mathrm{fg}}
\newcommand{\bg}{\mathrm{bg}}
\newcommand{\sigmoid}{\operatorname{sigmoid}}
\newcommand{\logit}{\operatorname{logit}}
\newcommand{\norm}{\operatorname{Norm}}
\DeclareMathOperator*{\argmax}{arg\,max}

\DeclareCaptionFont{currentcaptionlinks}{\hypersetup{allcolors=cvprblue}}
\newcommand{\UseCaptionStatus}[1]{%
  \captionsetup{font+={currentcaptionlinks,color=black},labelfont+={color=black},textfont+={color=black}}%
}

\makeatletter
\newcommand{\avgdividerfill}{%
  \ifx\CT@row@color\relax\else
    \rlap{\begingroup\let\CT@color\color\CT@row@color
      \vrule width6.35pt height\ht\@arstrutbox depth\dp\@arstrutbox
    \endgroup}%
  \fi}
\makeatother

\makeatletter
\newcommand{\avgdividerstroke}{\smash{\color{black}%
  \vrule width0.35pt height\ht\@arstrutbox depth\dp\@arstrutbox}}
\makeatother

\makeatletter
\newcommand{\tablegapfill}[1]{%
  \ifx\CT@row@color\relax\else
    \rlap{\begingroup\let\CT@color\color\CT@row@color
      \vrule width#1 height\ht\@arstrutbox depth\dp\@arstrutbox
    \endgroup}%
  \fi}
\makeatother


\definecolor{datasetlabel}{gray}{0.28}
\definecolor{maindivider}{gray}{0.68}
\newcommand{\datasetcategory}[1]{{\normalfont\fontsize{7.2}{8}\selectfont\color{datasetlabel}#1}}
\newcommand{\mainavgrule}{{\color{maindivider}\vrule width0.35pt}}
\newcommand{\mainavgheadrule}{\hspace{-0.35pt}\mainavgrule\hspace{0.35pt}}
\newcommand{\usecomparisonbox}[2]{\usebox{#1}}
\newlength{\comparisonmeasurewidth}
\newcommand{\extendcomparisonwidth}[2]{%
  \settowidth{\comparisonmeasurewidth}{#2}%
  \ifdim\comparisonmeasurewidth>#1\setlength{#1}{\comparisonmeasurewidth}\fi}

\newlength{\FigureCaptionGap}
\hypersetup{pdftitle={ReSA: Preserving Reference-Specified Foreground in In-Context Segmentation},pdfkeywords={in-context segmentation, few-shot segmentation, test-time adaptation, vision foundation models}}
\hypersetup{pdfauthor={Dezhao Su, Bingxu Hou, Jingdi Lei, Lin Qian, Likai Wang, Yunyun Wan}}
\def\confName{CVPR}
\def\confYear{2027}
\title{ReSA: Preserving Reference-Specified Foreground in In-Context Segmentation}
\author{%
  {\fontsize{11}{13}\selectfont
   Dezhao Su\textsuperscript{1}\thanks{Corresponding author.}, Bingxu Hou\textsuperscript{2},
   Jingdi Lei\textsuperscript{3}, Lin Qian\textsuperscript{4},
   Likai Wang\textsuperscript{5}, Yunyun Wan\textsuperscript{6}}\\[5pt]
  {\fontsize{9.5}{11.5}\selectfont
   \textsuperscript{1}Fudan University\quad
   \textsuperscript{2}Zhejiang University}\\
  {\fontsize{9.5}{11.5}\selectfont
   \textsuperscript{3}Nanyang Technological University\quad
   \textsuperscript{4}University of Manchester}\\
  {\fontsize{9.5}{11.5}\selectfont
   \textsuperscript{5}Tsinghua University\quad
   \textsuperscript{6}Beijing University of Posts and Telecommunications}
}
\makeatletter
\patchcmd{\@maketitle}{\vspace*{12pt}}{\vspace*{0pt}}{}%
  {\PackageError{ReSA}{Could not compact the gap above the teaser}{}}
\patchcmd{\@maketitle}{\vspace*{24pt}}{\vspace*{12pt}}{}%
  {\PackageError{ReSA}{Could not compact the title-to-author gap}{}}
\apptocmd{\@maketitle}{%
  \begin{minipage}{\textwidth}
  \centering
  \includegraphics[width=\linewidth]{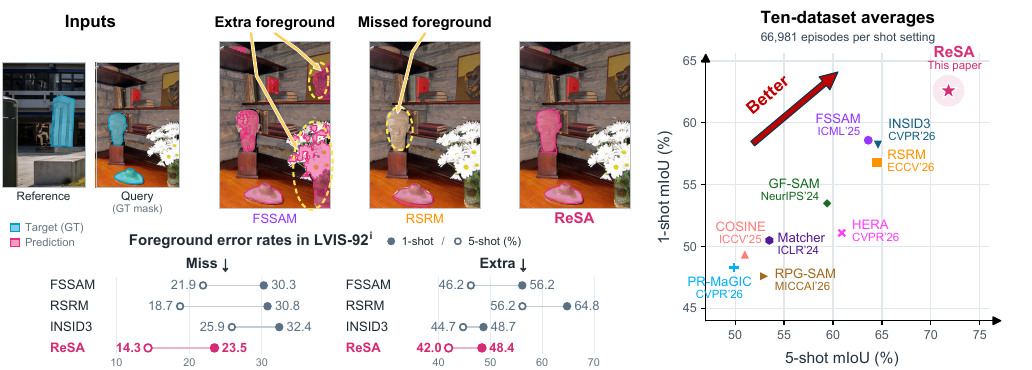}
  \captionsetup{hypcap=false}
  \UseCaptionStatus{fig:teaser}
  \captionof{figure}{\textbf{Foreground coverage and segmentation performance.} Left: a matched 5-shot example shows extra and missed foreground; one of five references is shown. Below: foreground miss $\mathrm{FN}/(\mathrm{TP}+\mathrm{FN})$ and extra $\mathrm{FP}/(\mathrm{TP}+\mathrm{FP})$ rates (\%) on LVIS-$92^i$, a natural-object segmentation benchmark with 23,000 test episodes per shot setting. Rates are averaged over classes and then folds; filled/open markers denote 1-shot/5-shot. Right: mean 1-shot versus 5-shot mIoU (\%) across ten benchmarks (Tables~\ref{tab:comparison_natural}--\ref{tab:comparison_parts_domains}); the upper right is better.}
  \label{fig:teaser}
  \end{minipage}\par\vspace{14pt}%
  \noindent\begin{minipage}{\columnwidth}
    \centerline{\large\bf Abstract}
  \end{minipage}\par\vspace{12pt}%
}{}{\PackageError{ReSA}{Could not attach first-page teaser}{}}
\patchcmd{\abstract}{\centerline{\large\bf Abstract}\vspace*{12pt}}{}{}%
  {\PackageError{ReSA}{Could not relocate the abstract heading}{}}
\makeatother
\begin{document}
\maketitle
\begin{abstract}
In-context segmentation uses annotated references to specify the target and its extent in a query image. Accurately transferring this reference-specified foreground requires recovering target regions while suppressing irrelevant predictions. We propose ReSA, a reference-guided framework that unifies query adaptation and mask selection through a shared reference kernel. ReSA leverages foreground and background reference prototypes to constrain prediction changes during adaptation and guide foreground selection through reference-weighted reconstruction. This shared guidance promotes consistent use of reference information across prediction refinement and mask formation. ReSA further selects among adapted predictions and aligns boundary refinement with the selected foreground threshold. We establish a theoretical bound connecting reference-response changes during adaptation and soft reconstruction. With a frozen visual encoder and lightweight test-time adaptation, ReSA requires no offline task training. Ablation studies demonstrate improved foreground retention and effective suppression of false-positive regions, while qualitative results show more complete masks with fewer isolated artifacts. Across ten object, part, and cross-domain segmentation benchmarks, ReSA achieves the highest average mIoU among the evaluated methods, reaching 62.6\% in 1-shot and 71.8\% in 5-shot.
\par
\end{abstract}

\section{Introduction}
\label{sec:intro}

In-context segmentation uses a few annotated reference images to specify the target in a query image.
The reference masks define both target identity and foreground extent, supporting object, part, and cross-domain segmentation~\cite{lvis,coco,fss1000,voc,pascalpart,paco,isic,suim,isaid,xray}.
Reference foreground and background provide examples of target and non-target content.
Segmentation must recover the specified extent while excluding surrounding distractors.

Existing approaches use learned decoders~\cite{seggpt,segic,sine} or pretrained mask generators~\cite{matcher,persam,gfsam}.
Decoder-based methods learn reference-conditioned segmentation from annotated training data.
This requires task-specific supervision; cross-domain research reports declining target-domain performance with continued source-domain training~\cite{loec}.
SAM-based approaches combine semantic correspondence with a mask generator, requiring prompt selection and alignment with the reference-specified mask granularity~\cite{gfsam}.
These dependencies have motivated segmentation directly from dense self-supervised features.

On frozen features, reference matching and feature affinities guide query-region selection and aggregation~\cite{insid3,rsrm}.
The conversion from correspondence to foreground regions determines mask quality.
Confident predictions can include distractors and omit target regions.
Figure~\ref{fig:teaser} illustrates two remaining errors: incomplete target coverage and extra foreground from unrelated regions.
Part segmentation further requires selecting the annotated portion of an object (Fig.~\ref{fig:qualitative}).
These errors motivate using reference foreground and background throughout mask inference.

We study foreground recovery through two connected decisions: query adaptation and mask selection.
Adaptation updates foreground probabilities; selection determines which regions enter the mask.
Both affect coverage and background inclusion.
Our central idea is to use the same reference-to-query similarities to guide both decisions.
The annotated context thus informs both the adaptation objective and the selection criterion.

We propose \method, a unified framework for reference-guided adaptation and mask selection (Fig.~\ref{fig:overview}).
A shared reference kernel represents similarities between query features and foreground/background reference prototypes.
Response-constrained adaptation (RCA) uses this kernel to regularize weighted foreground responses during entropy minimization.
Reference-weighted mask selection (RMS) reuses the kernel to fit candidate-specific thresholds through weighted probability reconstruction and selects a candidate by relative reconstruction gain.
Threshold-aligned refinement (TAR) uses the selected threshold to align the CRF unary scores before refining the mask boundaries.
The framework keeps the encoder frozen and adapts lightweight classifiers separately for each episode.

Our contributions are:
\begin{itemize}
\item We propose \method, an in-context segmentation framework that uses foreground and background reference cues to guide query adaptation and mask selection, improving target coverage and suppressing false positives.

\item We connect response regularization and reconstruction-based threshold fitting through a shared reference kernel, and select candidates by relative reconstruction gain. We derive a bound linking adaptation and soft reconstruction, and align boundary refinement with the selected threshold.

\item Across ten benchmarks, \method achieves the highest average mIoU among the evaluated methods: 62.6\% in 1-shot and 71.8\% in 5-shot. Ablations quantify the contributions of reference guidance, threshold fitting, and candidate selection; qualitative comparisons show more complete masks with fewer isolated false-positive regions.
\end{itemize}

\begin{figure}[!t]
\centering
\includegraphics[width=\columnwidth]{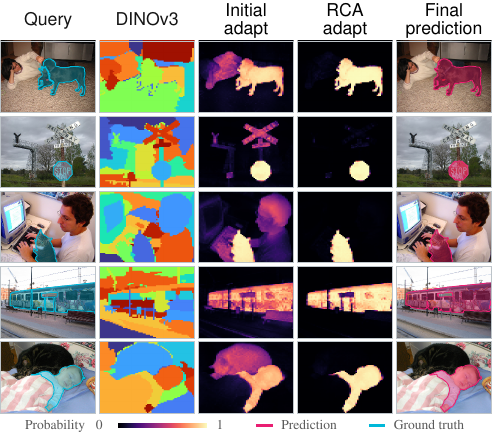}
\UseCaptionStatus{fig:mechanism}
\caption{\textbf{From reference fitting to query adaptation.} Columns show the query with its ground-truth target, DINOv3 feature clusters, initial probabilities $p_0$, probabilities after RCA, and the final mask. Probability maps share a $[0,1]$ scale. Cyan/magenta denote ground truth/predictions. Loss ablations are in Fig.~\ref{fig:module_validation}(a).}
\label{fig:mechanism}
\end{figure}

\begin{figure*}[!t]
\centering
\includegraphics[width=\textwidth]{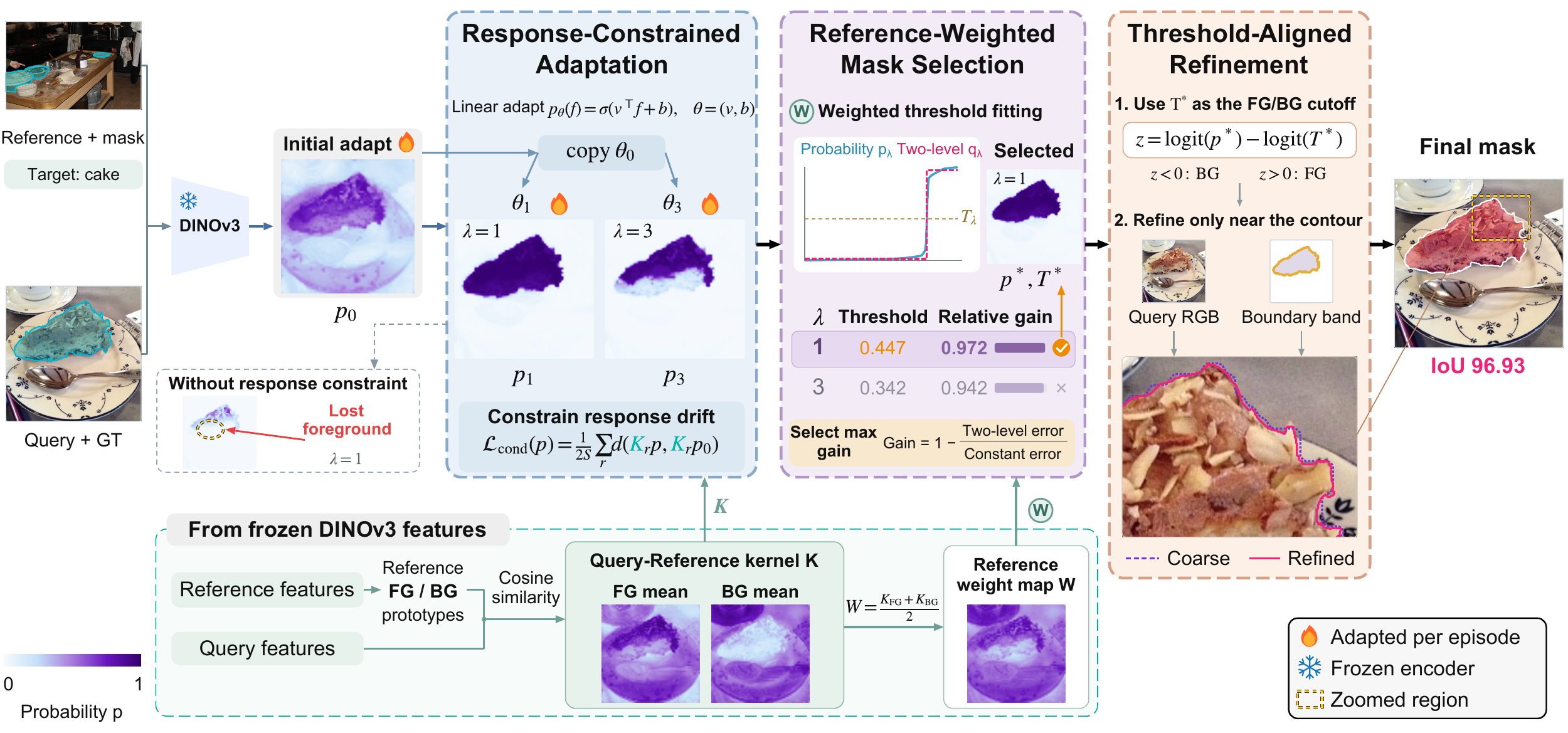}
\UseCaptionStatus{fig:overview}
\caption{\textbf{Reference guidance from predictions to masks.}
A recorded 1-shot example; query ground truth is shown for visualization. Frozen DINOv3 features form the reference kernel $K$.
Reference fitting produces $p_0$; RCA adapts two copies of the head with entropy weights $\lambda\in\{1,3\}$ while penalizing changes from $Kp_0$ to $Kp_\lambda$.
RMS uses the reference weight map $W$ (entries $w_i$) to fit thresholds and select $(p^*,T^*)$ by relative reconstruction gain.
TAR shifts the CRF unary scores to use $T^*$ and refines the coarse mask using query RGB and spatial cues.
The inset shows the matched $\lambda=1$ control without the response constraint. Probability maps, thresholds, gains, and masks are recorded outputs.}
\label{fig:overview}
\end{figure*}

\section{Related Work}
\label{sec:related}

\paragraph{Reference-guided segmentation.}
SegGPT, SegIC, and SINE learn reference-conditioned segmentation through in-context prediction, correspondence-derived instructions, and task-aware decoding~\cite{seggpt,segic,sine}.
DiffewS adapts latent-diffusion features~\cite{diffews,ldm}; FSSAM and SANSA adapt SAM~2~\cite{sam2} for few-shot segmentation~\cite{fssam,sansa}; LoEC emphasizes low-level features for cross-domain transfer~\cite{loec}.
Matcher, PerSAM, and GF-SAM use reference cues to guide pretrained mask generators~\cite{matcher,persam,gfsam,sam}; GSM improves semantic mining~\cite{gsm}.
REBASE projects reference and query features away from a reference-background subspace to improve matching and prompt localization~\cite{rebase}.

\paragraph{Constructing masks from frozen features.}
DINO representations support segmentation through feature grouping and reference matching~\cite{dinov2,dinov3}.
INSID3 selects a reference-related seed cluster and expands it through query-cluster aggregation~\cite{insid3}.
RSRM improves matching through feature fusion, query-guided support-feature enhancement, and global, regional, and pixel-level prototypes~\cite{rsrm}.
FROST estimates foreground and background feature densities from the references and applies a Bayes decision rule with equal class priors~\cite{frost}.

\paragraph{Test-time adaptation and foreground inference.}
Transductive methods use unlabeled query data to adapt predictions within a few-shot episode~\cite{repri}.
HERA combines feature-layer selection, attention regularization, and pixel-logit calibration for cross-domain segmentation~\cite{hera}.
Threshold selection converts query probabilities into a foreground mask.
RPG-SAM selects thresholds through geometric agreement among candidate masks~\cite{rpgsam}.
Bregman quantization groups continuous values by minimizing reconstruction error~\cite{banerjee2005}.
Dense CRFs refine boundaries using prediction scores, pixel positions, and appearance~\cite{densecrf,fwcrf}.
\method uses the same reference kernel to regularize query adaptation and fit mask thresholds, then uses the selected threshold in CRF refinement.

\section{Method}
\label{sec:method}

Given $S$ annotated references $\mathcal R=\{(I_s^r,M_s^r)\}_{s=1}^S$ and an unlabeled query $I^q$, \method predicts the foreground at the semantic granularity specified by the reference masks.
Figure~\ref{fig:overview} summarizes the inference pipeline.

A frozen encoder maps each image to a spatial grid of $N$ tokens with $D$-dimensional features.
Following INSID3~\cite{insid3}, we remove the positional subspace:
\begin{equation}
\widetilde F=\norm\!\left(F(I_D-BB^\top)\right),
\label{eq:debias}
\end{equation}
where $B$ is an orthonormal basis derived from a constant input and $\norm$ normalizes each row.
We use the resulting features for reference--query matching and foreground prediction.
Before classifier fitting, a reference--query matching check identifies episodes with no foreground candidates, which receive an empty mask.

\subsection{Reference kernel and responses}
\label{sec:responses}

For each reference image, we use masked average pooling on the token grid to compute foreground and background prototypes.
Let $c_r$, $r=1,\ldots,2S$, denote the normalized prototypes.
Applying a softmax to their cosine similarities with normalized query features $f_i^q$ gives the reference kernel:
\begin{equation}
K_{ri}=\frac{\exp(c_r^\top f_i^q/\tau)}
{\sum_j\exp(c_r^\top f_j^q/\tau)},\qquad \tau=0.05.
\label{eq:kernel}
\end{equation}
Each row of $K\in\mathbb R^{2S\times N}$ contains nonnegative weights that sum to one.
For a query foreground probability map $p\in[0,1]^N$, the reference response $(Kp)_r=\sum_iK_{ri}p_i$ is a weighted average that emphasizes query features similar to prototype $r$.
Foreground and background prototypes apply different weights to the same foreground probability map.
With 4,096 query tokens, $Kp$ contains two responses in 1-shot and ten in 5-shot.
We keep $K$ fixed and use it to compute RCA's responses and RMS's reconstruction weights.

\subsection{Response-constrained adaptation (RCA)}
\label{sec:adapt}

RCA regularizes query foreground responses against their values after reference fitting during entropy minimization.
We use a linear foreground classifier $p(f)=\sigmoid(v^\top f+b)$ with $D+1$ parameters and initialize $v$ and $b$ to zero for each episode.
We first fit it to the reference annotations by minimizing binary cross-entropy $\mathcal L_{\rm ref}$, weighted equally across foreground/background classes and references.
The fitted classifier produces the initial query probabilities $p_0$ and initializes two branches, which are adapted independently.
Writing $d(u,v)=D_{\rm KL}(\operatorname{Ber}(u)\|\operatorname{Ber}(v))$, we penalize deviations from the initial responses $Kp_0$:
\begin{equation}
\mathcal L_{\rm cond}(p)=\frac{1}{2S}\sum_{r=1}^{2S}d(K_rp,K_rp_0).
\label{eq:conditional}
\end{equation}
Each branch minimizes
\begin{equation}
\mathcal L_\lambda=\mathcal L_{\rm ref}+\frac{1}{S}
\left[\mathcal L_{\rm cond}(p)+\lambda\overline H(p)\right],
\label{eq:adaptation}
\end{equation}
where $\overline H(p)=N^{-1}\sum_i H(p_i)$ is mean Bernoulli entropy.
The factor $1/S$ reduces the weight of the query losses relative to reference supervision as the number of references increases; $1/(2S)$ averages the prototype-wise KL terms.
The branches use entropy weights $\lambda=1$ and $3$, with the same initialization and number of optimization steps.
RMS takes the resulting probability maps $p_1$ and $p_3$ as input.

\subsection{Reference-weighted mask selection (RMS)}
\label{sec:threshold}

\textbf{Threshold fitting.} RMS fits a separate threshold to each adapted probability map to account for differences in confidence scale, then selects a candidate mask.
For a threshold $t$, it approximates the probabilities using two fitted values:
\begin{equation}
q_{\lambda,i}(t,a,b)=a+(b-a)\mathbf1[p_{\lambda,i}>t],
\label{eq:reconstruction}
\end{equation}
with $0<a\leq b<1$.
Tokens with $p_{\lambda,i}\leq t$ receive the fitted value $a$, and those with $p_{\lambda,i}>t$ receive $b$.
The threshold partition defines the binary mask.
We average the rows of $K$ to obtain the reference weight map $W$, with entries $w_i=(2S)^{-1}\sum_rK_{ri}$ and $\sum_iw_i=1$, and minimize the weighted Bernoulli KL divergence:
\begin{equation}
D_\lambda(t,a,b)=\sum_iw_i d(p_{\lambda,i},q_{\lambda,i}).
\label{eq:distortion}
\end{equation}
The reference weights emphasize reconstruction accuracy at query locations similar to the annotated foreground or background.
Threshold fitting thus uses the same reference-to-query similarities as RCA.
For each nonempty split with positive group weights, the optimal values $a$ and $b$ are the weighted mean probabilities within the two groups.
Sorting the probabilities and evaluating prefix sums finds the best split in $O(N\log N)$ time.
This gives the minimum error $D_\lambda^*$ and a threshold $T_\lambda$ midway between the adjacent distinct probabilities at that split.

\textbf{Candidate selection.}
We score each candidate by the fraction of its best constant-reconstruction error removed by the two-level fit.
For $D_{\lambda,\mathrm{const}}>0$, we define
\begin{equation}
s_\lambda=1-\frac{D_\lambda^*}{D_{\lambda,\rm const}},\qquad
\lambda^*=\argmax_{\lambda\in\{1,3\}}s_\lambda.
\label{eq:selection}
\end{equation}
Here $D_{\lambda,\rm const}$ is the minimum error in Eq.~\ref{eq:distortion} for a constant map $q=c\mathbf1$, attained at $c=\sum_iw_ip_{\lambda,i}$.
Ties select $\lambda=3$.
The selected probability map $p^*=p_{\lambda^*}$ and threshold $T=T_{\lambda^*}$ define the coarse mask
\begin{equation}
\widehat m_i=\mathbf1[p_i^\star>T].
\label{eq:mask}
\end{equation}
TAR receives $(p^*,T)$.

\subsection{Threshold-aligned refinement (TAR)}
\label{sec:refinement}

TAR aligns the foreground/background preference of the CRF with the threshold selected by RMS.
We subtract $\operatorname{logit}(T)$ from the query logits:
\begin{equation}
z_i=\logit(p_i^\star)-\logit(T).
\label{eq:aligned}
\end{equation}
On the token grid, $z_i>0\iff p_i^\star>T$, so the unary scores agree with the coarse foreground/background labels.
For example, $p_i=0.4$ favors foreground when $T=0.3$.

We bilinearly upsample and binarize $\widehat m$ on the CRF grid, then use dilation and erosion to define a boundary band $\mathcal B$.
Interpolating $z$ to this grid gives the foreground/background scores
\begin{equation}
\ell_x^{\fg}=\eta\widetilde z_x/2,\qquad
\ell_x^{\bg}=-\eta\widetilde z_x/2,\quad x\in\mathcal B,
\label{eq:unary}
\end{equation}
with $\eta=8$.
Within the band, these scores define the unary terms.
The spatial pairwise term uses pixel positions, and the bilateral term uses both positions and RGB values.
We use the regularized Frank--Wolfe solver adopted by INSID3~\cite{insid3,densecrf,fwcrf} and retain the coarse mask's foreground core and exterior background in the final mask.

\begin{table*}[t]
\centering
\newsavebox{\mainnaturalbox}
\begin{lrbox}{\mainnaturalbox}%
\fontsize{8}{9}\selectfont%
\setlength{\tabcolsep}{2pt}%
\renewcommand{\arraystretch}{1.03}%
\newlength{\naturalMethodWidth}%
\setlength{\naturalMethodWidth}{0pt}%
\extendcomparisonwidth{\naturalMethodWidth}{\textbf{Method}}%
\extendcomparisonwidth{\naturalMethodWidth}{SegGPT\cite{seggpt}}%
\extendcomparisonwidth{\naturalMethodWidth}{SegIC\cite{segic}}%
\extendcomparisonwidth{\naturalMethodWidth}{SINE\cite{sine}}%
\extendcomparisonwidth{\naturalMethodWidth}{DiffewS\cite{diffews}}%
\extendcomparisonwidth{\naturalMethodWidth}{LDIS-1\tableevalmark\cite{ldis}}%
\extendcomparisonwidth{\naturalMethodWidth}{FSSAM\tableevalmark\cite{fssam}}%
\extendcomparisonwidth{\naturalMethodWidth}{COSINE\tableevalmark\cite{cosine}}%
\extendcomparisonwidth{\naturalMethodWidth}{PerSAM\tableevalmark\cite{persam}}%
\extendcomparisonwidth{\naturalMethodWidth}{Matcher\cite{matcher}}%
\extendcomparisonwidth{\naturalMethodWidth}{GF-SAM\cite{gfsam}}%
\extendcomparisonwidth{\naturalMethodWidth}{GSM+Matcher\cite{gsm}}%
\extendcomparisonwidth{\naturalMethodWidth}{RNS\tableevalmark\cite{rns}}%
\extendcomparisonwidth{\naturalMethodWidth}{PR-MaGIC\tableevalmark\cite{prmagic}}%
\extendcomparisonwidth{\naturalMethodWidth}{RPG-SAM\tableevalmark\cite{rpgsam}}%
\extendcomparisonwidth{\naturalMethodWidth}{INSID3\cite{insid3}}%
\extendcomparisonwidth{\naturalMethodWidth}{HERA\tableevalmark\cite{hera}}%
\extendcomparisonwidth{\naturalMethodWidth}{RSRM\tableevalmark\cite{rsrm}}%
\extendcomparisonwidth{\naturalMethodWidth}{FROST\cite{frost}}%
\extendcomparisonwidth{\naturalMethodWidth}{ReSA (ours)}%
\addtolength{\naturalMethodWidth}{0.3pt}%
\newlength{\naturalVenueWidth}%
\setlength{\naturalVenueWidth}{0pt}%
\extendcomparisonwidth{\naturalVenueWidth}{\textbf{Venue}}%
\extendcomparisonwidth{\naturalVenueWidth}{ICCV'23}%
\extendcomparisonwidth{\naturalVenueWidth}{ECCV'24}%
\extendcomparisonwidth{\naturalVenueWidth}{NeurIPS'24}%
\extendcomparisonwidth{\naturalVenueWidth}{AAAI'25}%
\extendcomparisonwidth{\naturalVenueWidth}{AAAI'26}%
\extendcomparisonwidth{\naturalVenueWidth}{ICML'25}%
\extendcomparisonwidth{\naturalVenueWidth}{ICCV'25}%
\extendcomparisonwidth{\naturalVenueWidth}{ICLR'24}%
\extendcomparisonwidth{\naturalVenueWidth}{CVPR'26}%
\extendcomparisonwidth{\naturalVenueWidth}{MICCAI'26}%
\extendcomparisonwidth{\naturalVenueWidth}{ECCV'26}%
\extendcomparisonwidth{\naturalVenueWidth}{arXiv'26}%
\extendcomparisonwidth{\naturalVenueWidth}{This paper}%
\addtolength{\naturalVenueWidth}{0.3pt}%
\newlength{\naturalBackboneWidth}%
\setlength{\naturalBackboneWidth}{0pt}%
\extendcomparisonwidth{\naturalBackboneWidth}{\textbf{Backbone}}%
\extendcomparisonwidth{\naturalBackboneWidth}{ViT-L}%
\extendcomparisonwidth{\naturalBackboneWidth}{DINOv2-L}%
\extendcomparisonwidth{\naturalBackboneWidth}{SD 2.1}%
\extendcomparisonwidth{\naturalBackboneWidth}{DINOv2+SAM2}%
\extendcomparisonwidth{\naturalBackboneWidth}{DINOv2-L+CLIP}%
\extendcomparisonwidth{\naturalBackboneWidth}{DINOv2-L+SAM-H}%
\extendcomparisonwidth{\naturalBackboneWidth}{DINOv3+SAM2.1}%
\extendcomparisonwidth{\naturalBackboneWidth}{DINOv2-L+SAM2-L}%
\extendcomparisonwidth{\naturalBackboneWidth}{DINOv3-L}%
\extendcomparisonwidth{\naturalBackboneWidth}{SD1.5+AlphaCLIP}%
\addtolength{\naturalBackboneWidth}{0.3pt}%
\newlength{\naturalParamsWidth}%
\setlength{\naturalParamsWidth}{0pt}%
\extendcomparisonwidth{\naturalParamsWidth}{\textbf{Params}}%
\extendcomparisonwidth{\naturalParamsWidth}{354M}%
\extendcomparisonwidth{\naturalParamsWidth}{310M}%
\extendcomparisonwidth{\naturalParamsWidth}{373M}%
\extendcomparisonwidth{\naturalParamsWidth}{890M}%
\extendcomparisonwidth{\naturalParamsWidth}{132M}%
\extendcomparisonwidth{\naturalParamsWidth}{691M}%
\extendcomparisonwidth{\naturalParamsWidth}{945M}%
\extendcomparisonwidth{\naturalParamsWidth}{528M}%
\extendcomparisonwidth{\naturalParamsWidth}{529M}%
\extendcomparisonwidth{\naturalParamsWidth}{303M}%
\extendcomparisonwidth{\naturalParamsWidth}{312M}%
\extendcomparisonwidth{\naturalParamsWidth}{1248M}%
\addtolength{\naturalParamsWidth}{0.3pt}%
\newlength{\naturalNumberWidth}%
\setlength{\naturalNumberWidth}{\dimexpr(\textwidth-\naturalMethodWidth-\naturalVenueWidth-\naturalBackboneWidth-\naturalParamsWidth-26\tabcolsep-0.35pt)/10\relax}%
\typeout{MAIN-NATURAL-METHOD-WIDTH=\the\naturalMethodWidth}%
\typeout{MAIN-NATURAL-VENUE-WIDTH=\the\naturalVenueWidth}%
\typeout{MAIN-NATURAL-BACKBONE-WIDTH=\the\naturalBackboneWidth}%
\typeout{MAIN-NATURAL-PARAMS-WIDTH=\the\naturalParamsWidth}%
\typeout{MAIN-NATURAL-NUMBER-WIDTH=\the\naturalNumberWidth}%
\xdef\naturalAvgOffset{\the\dimexpr 2\naturalNumberWidth+3\tabcolsep+0.35pt\relax}%
\begin{tabular}{@{}>{\raggedright\arraybackslash}p{\dimexpr\naturalMethodWidth+3pt\relax}@{\hspace{\dimexpr2\tabcolsep-2pt\relax}}>{\raggedright\arraybackslash}p{\naturalVenueWidth}>{\raggedright\arraybackslash}p{\naturalBackboneWidth}@{\hspace{\dimexpr2\tabcolsep-1pt\relax}}>{\centering\arraybackslash}p{\naturalParamsWidth}*{8}{>{\centering\arraybackslash}p{\naturalNumberWidth}}!{\mainavgrule}*{2}{>{\centering\arraybackslash}p{\naturalNumberWidth}}@{}}%
\toprule
\multirow{2}{*}[-1.4pt]{\textbf{Method}} & \multirow{2}{*}[-1.4pt]{\textbf{Venue}} & \multirow{2}{*}[-1.4pt]{\textbf{Backbone}} & \multirow{2}{*}[-1.4pt]{\textbf{Params}} & \multicolumn{2}{c}{\textbf{LVIS-92$^i$}} & \multicolumn{2}{c}{\textbf{COCO-20$^i$}} & \multicolumn{2}{c}{\textbf{FSS-1000}} & \multicolumn{2}{c}{\textbf{PASCAL-5$^i$}} & \multicolumn{2}{!{\mainavgheadrule}c}{\textbf{Avg (4)}} \\
\cmidrule(lr){5-6}\cmidrule(lr){7-8}\cmidrule(lr){9-10}\cmidrule(lr){11-12}\cmidrule(lr){13-14}
 & & & & 1-shot & 5-shot & 1-shot & 5-shot & 1-shot & 5-shot & 1-shot & 5-shot & 1-shot & 5-shot \\
\midrule
\multicolumn{14}{@{}l}{\textbf{Task-specific fine-tuning:} \textit{Semantic + mask supervision}} \\
\hspace*{3pt}SegGPT\cite{seggpt} & ICCV'23 & ViT-L & 354M & 18.6 & 25.4 & \textcolor{trainoverlap}{56.1} & \textcolor{trainoverlap}{67.9} & 85.6 & 89.3 & \textcolor{trainoverlap}{83.2} & \textcolor{trainoverlap}{89.8} & \textcolor{trainoverlap}{60.9} & \textcolor{trainoverlap}{68.1}\\
\hspace*{3pt}SegIC\cite{segic} & ECCV'24 & DINOv2-L & 310M & \textcolor{trainoverlap}{44.6} & \textcolor{trainoverlap}{46.8}\rlap{\tableevalmark} & \textcolor{trainoverlap}{76.1} & \textcolor{trainoverlap}{74.4}\rlap{\tableevalmark} & 86.8 & 87.3\rlap{\tableevalmark} & \textcolor{trainoverlap}{85.6} & \textcolor{trainoverlap}{88.4}\rlap{\tableevalmark} & \textcolor{trainoverlap}{73.3} & \textcolor{trainoverlap}{74.2}\\
\hspace*{3pt}SINE\cite{sine} & NeurIPS'24 & DINOv2-L & 373M & 31.2 & 35.5 & \textcolor{trainoverlap}{64.5} & \textcolor{trainoverlap}{66.1} & 84.3\rlap{\tableevalmark} & 85.0\rlap{\tableevalmark} & \textcolor{trainoverlap}{85.4} & \textcolor{trainoverlap}{86.2} & \textcolor{trainoverlap}{66.3} & \textcolor{trainoverlap}{68.2}\\
\hspace*{3pt}DiffewS\cite{diffews} & NeurIPS'24 & SD 2.1 & 890M & 31.4 & 35.4 & \textcolor{trainoverlap}{71.3} & \textcolor{trainoverlap}{72.2} & 87.8 & 88.0 & \textcolor{trainoverlap}{88.3} & \textcolor{trainoverlap}{87.8} & \textcolor{trainoverlap}{69.7} & \textcolor{trainoverlap}{70.9}\\
\hspace*{3pt}LDIS-1\tableevalmark\cite{ldis} & AAAI'25 & SD1.5+AlphaCLIP & 1248M & 15.9 & 20.4 & \textcolor{trainoverlap}{57.7} & \textcolor{trainoverlap}{63.8} & 67.5 & 69.3 & \textcolor{trainoverlap}{84.5} & \textcolor{trainoverlap}{86.0} & \textcolor{trainoverlap}{56.4} & \textcolor{trainoverlap}{59.9}\\
\hspace*{3pt}FSSAM\tableevalmark\cite{fssam} & ICML'25 & DINOv2+SAM2 & 132M & 37.2 & \underline{47.7} & \textcolor{trainoverlap}{62.3} & \textcolor{trainoverlap}{71.0} & \textbf{88.7} & \textbf{89.6} & \textcolor{trainoverlap}{79.5} & \textcolor{trainoverlap}{86.3} & \textcolor{trainoverlap}{66.9} & \textcolor{trainoverlap}{73.7}\\
\hspace*{3pt}COSINE\tableevalmark\cite{cosine} & ICCV'25 & DINOv2-L+CLIP & 691M & \textcolor{trainoverlap}{35.3} & \textcolor{trainoverlap}{40.8} & \textcolor{trainoverlap}{68.0} & \textcolor{trainoverlap}{71.5} & 87.2 & 87.1 & \textcolor{trainoverlap}{86.2} & \textcolor{trainoverlap}{88.3} & \textcolor{trainoverlap}{69.1} & \textcolor{trainoverlap}{71.9}\\
\midrule
\multicolumn{14}{@{}l}{\textbf{No offline task training:} \textit{Mask-supervised pre-training}} \\
\hspace*{3pt}PerSAM\tableevalmark\cite{persam} & ICLR'24 & DINOv2-L+SAM-H & 945M & 20.1 & 24.3 & 34.3 & 39.1 & 72.1 & 72.8 & 50.6 & 56.0 & 44.3 & 48.0\\
\hspace*{3pt}Matcher\cite{matcher} & ICLR'24 & DINOv2-L+SAM-H & 945M & 33.0 & 40.0 & 52.7 & 60.7 & 87.0 & \textbf{89.6} & 68.1 & 74.0 & 60.2 & 66.1\\
\hspace*{3pt}GF-SAM\cite{gfsam} & NeurIPS'24 & DINOv2-L+SAM-H & 945M & 35.2 & 44.2 & \underline{58.7} & \underline{66.8} & \underline{88.0} & 88.9 & \underline{72.1} & \underline{82.6} & \underline{63.5} & \underline{70.6}\\
\hspace*{3pt}GSM+Matcher\cite{gsm} & AAAI'26 & DINOv2-L+SAM-H & 945M & 34.3 & 41.9 & 55.9 & 62.7 & 87.3 & \textbf{89.6} & 71.3 & 75.7 & 62.2 & 67.5\\
\hspace*{3pt}RNS\tableevalmark\cite{rns} & CVPR'26 & DINOv3+SAM2.1 & 528M & 27.1 & 34.4 & 45.1 & 58.8 & 81.6 & 87.0 & 62.8 & 81.6 & 54.1 & 65.5\\
\hspace*{3pt}PR-MaGIC\tableevalmark\cite{prmagic} & CVPR'26 & DINOv2-L+SAM-H & 945M & 23.8 & 24.5 & 44.5 & 48.2 & 86.6 & 88.6 & 63.8 & 69.8 & 54.7 & 57.8\\
\hspace*{3pt}RPG-SAM\tableevalmark\cite{rpgsam} & MICCAI'26 & DINOv2-L+SAM2-L & 529M & 28.4 & 39.7 & 47.7 & 53.1 & 75.6 & 77.9 & 55.5 & 64.0 & 51.8 & 58.7\\
\midrule
\multicolumn{14}{@{}l}{\textbf{No offline task training:} \textit{Self-supervised pre-training}} \\
\hspace*{3pt}INSID3\cite{insid3} & CVPR'26 & DINOv3-L & 303M & \underline{41.8} & 47.2 & 57.6 & 65.1 & 85.8\rlap{\tableevalmark} & 87.4\rlap{\tableevalmark} & 67.3\rlap{\tableevalmark} & 80.1\rlap{\tableevalmark} & 63.1 & 70.0\\
\hspace*{3pt}HERA\tableevalmark\cite{hera} & CVPR'26 & DINOv3-L & 312M & 26.0 & 33.2 & 45.3 & 54.0 & 80.2 & 82.5 & 61.6 & 74.7 & 53.3 & 61.1\\
\hspace*{3pt}RSRM\tableevalmark\cite{rsrm} & ECCV'26 & DINOv3-L & 303M & 31.3 & 39.9 & 51.6 & 61.8 & 80.6 & 82.8 & 68.8 & 79.4 & 58.1 & 66.0\\
\hspace*{3pt}FROST\cite{frost} & arXiv'26 & DINOv3-L & 303M & 32.8 & 47.1 & 48.4 & 60.5 & 82.6\rlap{\tableevalmark} & 86.0\rlap{\tableevalmark} & 70.1 & 82.5 & 58.5 & 69.0\\
\rowcolor{gainrow}
\hspace*{3pt}\resacell{ReSA (ours)} & \resacell{This paper} & \resacell{DINOv3-L} & \resacell{303M} & \resacell{\textbf{44.9}} & \resacell{\textbf{53.0}} & \resacell{\textbf{58.8}} & \resacell{\textbf{68.4}} & \resacell{86.4} & \resacell{\underline{89.5}} & \resacell{\textbf{73.8}} & \resacell{\textbf{88.9}} & \resacell{\textbf{66.0}} & \resacell{\textbf{74.9}}\\
\bottomrule
\end{tabular}%
\end{lrbox}
\typeout{MAIN-COMPARISON-NATURAL-WIDTH=\the\wd\mainnaturalbox}
\typeout{MAIN-COMPARISON-TEXT-WIDTH=\the\textwidth}
\begin{minipage}{\textwidth}
\UseCaptionStatus{tab:comparison_natural}
\caption{\textbf{Natural-object segmentation.} Foreground mIoU (\%); Avg (4) weights these four datasets equally. Params: total model parameters (M), including frozen encoders. Gray marks task-training overlap (possible for COSINE); affected entries and averages are unranked. Best/second best: \textbf{bold}/\underline{underlined}, including ties. $^\dagger$: local evaluation (whole row when attached to a method); --: unavailable.}
\label{tab:comparison_natural}
\centering
\usecomparisonbox{\mainnaturalbox}{\naturalAvgOffset}
\end{minipage}
\end{table*}

\begin{table*}[t]
\centering
\newsavebox{\mainpartsdomainbox}
\begin{lrbox}{\mainpartsdomainbox}%
\fontsize{8}{9}\selectfont%
\setlength{\tabcolsep}{1.5pt}%
\renewcommand{\arraystretch}{1.08}%
\newlength{\partsMethodWidth}%
\setlength{\partsMethodWidth}{0pt}%
\extendcomparisonwidth{\partsMethodWidth}{\textbf{Method}}%
\extendcomparisonwidth{\partsMethodWidth}{SegGPT\cite{seggpt}}%
\extendcomparisonwidth{\partsMethodWidth}{SegIC\cite{segic}}%
\extendcomparisonwidth{\partsMethodWidth}{SINE\cite{sine}}%
\extendcomparisonwidth{\partsMethodWidth}{DiffewS\cite{diffews}}%
\extendcomparisonwidth{\partsMethodWidth}{LDIS-1\tableevalmark\cite{ldis}}%
\extendcomparisonwidth{\partsMethodWidth}{FSSAM\tableevalmark\cite{fssam}}%
\extendcomparisonwidth{\partsMethodWidth}{COSINE\tableevalmark\cite{cosine}}%
\extendcomparisonwidth{\partsMethodWidth}{PerSAM\tableevalmark\cite{persam}}%
\extendcomparisonwidth{\partsMethodWidth}{Matcher\cite{matcher}}%
\extendcomparisonwidth{\partsMethodWidth}{GF-SAM\cite{gfsam}}%
\extendcomparisonwidth{\partsMethodWidth}{RNS\tableevalmark\cite{rns}}%
\extendcomparisonwidth{\partsMethodWidth}{PR-MaGIC\tableevalmark\cite{prmagic}}%
\extendcomparisonwidth{\partsMethodWidth}{RPG-SAM\tableevalmark\cite{rpgsam}}%
\extendcomparisonwidth{\partsMethodWidth}{INSID3\cite{insid3}}%
\extendcomparisonwidth{\partsMethodWidth}{HERA\tableevalmark\cite{hera}}%
\extendcomparisonwidth{\partsMethodWidth}{RSRM\tableevalmark\cite{rsrm}}%
\extendcomparisonwidth{\partsMethodWidth}{FROST\cite{frost}}%
\extendcomparisonwidth{\partsMethodWidth}{ReSA (ours)}%
\addtolength{\partsMethodWidth}{0.3pt}%
\newlength{\partsVenueWidth}%
\setlength{\partsVenueWidth}{0pt}%
\extendcomparisonwidth{\partsVenueWidth}{\textbf{Venue}}%
\extendcomparisonwidth{\partsVenueWidth}{ICCV'23}%
\extendcomparisonwidth{\partsVenueWidth}{ECCV'24}%
\extendcomparisonwidth{\partsVenueWidth}{NeurIPS'24}%
\extendcomparisonwidth{\partsVenueWidth}{AAAI'25}%
\extendcomparisonwidth{\partsVenueWidth}{ICML'25}%
\extendcomparisonwidth{\partsVenueWidth}{ICCV'25}%
\extendcomparisonwidth{\partsVenueWidth}{ICLR'24}%
\extendcomparisonwidth{\partsVenueWidth}{CVPR'26}%
\extendcomparisonwidth{\partsVenueWidth}{MICCAI'26}%
\extendcomparisonwidth{\partsVenueWidth}{ECCV'26}%
\extendcomparisonwidth{\partsVenueWidth}{arXiv'26}%
\extendcomparisonwidth{\partsVenueWidth}{This paper}%
\addtolength{\partsVenueWidth}{0.3pt}%
\newlength{\partsNumberWidth}%
\setlength{\partsNumberWidth}{\dimexpr(\textwidth-\partsMethodWidth-\partsVenueWidth-30\tabcolsep-0.35pt)/14\relax}%
\typeout{MAIN-PARTS-METHOD-WIDTH=\the\partsMethodWidth}%
\typeout{MAIN-PARTS-VENUE-WIDTH=\the\partsVenueWidth}%
\typeout{MAIN-PARTS-NUMBER-WIDTH=\the\partsNumberWidth}%
\xdef\partsAvgOffset{\the\dimexpr 2\partsNumberWidth+3\tabcolsep+0.35pt\relax}%
\begin{tabular}{@{}>{\raggedright\arraybackslash}p{\dimexpr\partsMethodWidth+3pt\relax}@{\hspace{\dimexpr2\tabcolsep-2pt\relax}}>{\raggedright\arraybackslash}p{\partsVenueWidth}@{\hspace{\dimexpr2\tabcolsep-1pt\relax}}*{12}{>{\centering\arraybackslash}p{\partsNumberWidth}}!{\mainavgrule}*{2}{>{\centering\arraybackslash}p{\partsNumberWidth}}@{}}%
\toprule
\multirow{3}{*}[-1.8pt]{\textbf{Method}} & \multirow{3}{*}[-1.8pt]{\textbf{Venue}} & \multicolumn{4}{@{}c}{\datasetcategory{object parts}} & \multicolumn{2}{c}{\datasetcategory{dermoscopy}} & \multicolumn{2}{c}{\datasetcategory{underwater}} & \multicolumn{2}{c}{\datasetcategory{aerial}} & \multicolumn{2}{c}{\datasetcategory{chest X-ray}} & \multicolumn{2}{!{\mainavgheadrule}c}{\multirow{2}{*}[-0.7pt]{\textbf{Avg (6)}}} \\
 & & \multicolumn{2}{@{}c}{\textbf{PASCAL-Part}} & \multicolumn{2}{c}{\textbf{PACO-Part}} & \multicolumn{2}{c}{\textbf{ISIC}} & \multicolumn{2}{c}{\textbf{SUIM}} & \multicolumn{2}{c}{\textbf{iSAID}} & \multicolumn{2}{c}{\textbf{X-Ray}} & \multicolumn{2}{!{\mainavgheadrule}c}{} \\
\cmidrule(lr){3-4}\cmidrule(lr){5-6}\cmidrule(lr){7-8}\cmidrule(lr){9-10}\cmidrule(lr){11-12}\cmidrule(lr){13-14}\cmidrule(lr){15-16}
 & & 1-shot & 5-shot & 1-shot & 5-shot & 1-shot & 5-shot & 1-shot & 5-shot & 1-shot & 5-shot & 1-shot & 5-shot & 1-shot & 5-shot \\
\midrule
\multicolumn{16}{@{}l}{\textbf{Task-specific fine-tuning:} \textit{Semantic + mask supervision}} \\
\hspace*{3pt}SegGPT\cite{seggpt} & ICCV'23 & 35.8 & 42.8 & \textcolor{trainoverlap}{13.5} & \textcolor{trainoverlap}{14.1} & 37.5 & 45.2 & 34.9 & 33.7 & \textcolor{trainoverlap}{30.9} & \textcolor{trainoverlap}{35.9} & 87.5 & \underline{89.1} & \textcolor{trainoverlap}{40.0} & \textcolor{trainoverlap}{43.5}\\
\hspace*{3pt}SegIC\cite{segic} & ECCV'24 & 39.9 & 38.9\rlap{\tableevalmark} & 25.9 & 30.5\rlap{\tableevalmark} & 25.3 & 29.1\rlap{\tableevalmark} & 52.5 & 64.5\rlap{\tableevalmark} & 46.1 & 51.0\rlap{\tableevalmark} & 34.5 & 29.6\rlap{\tableevalmark} & 37.4 & 40.6\\
\hspace*{3pt}SINE\cite{sine} & NeurIPS'24 & 36.2 & 36.4 & 23.3 & 25.4 & 25.8 & 28.6 & 50.7 & 54.8 & 38.3 & 40.5 & 39.8 & 40.6 & 35.7 & 37.7\\
\hspace*{3pt}DiffewS\cite{diffews} & NeurIPS'24 & 34.0 & 39.7 & 22.8 & 26.1 & 27.8 & 32.7 & 48.9 & 49.8 & 47.5 & 48.0 & 41.6 & 45.1 & 37.1 & 40.2\\
\hspace*{3pt}LDIS-1\tableevalmark\cite{ldis} & AAAI'25 & 33.2 & 32.9 & 18.1 & 20.7 & 24.2 & 24.5 & 25.1 & 26.8 & 20.4 & 21.8 & 35.1 & 34.7 & 26.0 & 26.9\\
\hspace*{3pt}FSSAM\tableevalmark\cite{fssam} & ICML'25 & 43.8 & 52.7 & 33.0 & 38.3 & 53.9 & 65.1 & 56.7 & 61.9 & 50.0 & 53.5 & 80.5 & 70.0 & 53.0 & 56.9\\
\hspace*{3pt}COSINE\tableevalmark\cite{cosine} & ICCV'25 & 36.2 & 36.0 & \textcolor{trainoverlap}{27.7} & \textcolor{trainoverlap}{28.9} & 28.1 & 29.0 & \underline{58.7} & 61.9 & 36.4 & 36.9 & 30.3 & 30.0 & \textcolor{trainoverlap}{36.2} & \textcolor{trainoverlap}{37.1}\\
\midrule
\multicolumn{16}{@{}l}{\textbf{No offline task training:} \textit{Mask-supervised pre-training}} \\
\hspace*{3pt}PerSAM\tableevalmark\cite{persam} & ICLR'24 & 31.6 & 32.5 & 25.0 & 28.5 & 35.5 & 38.9 & 26.0 & 28.8 & 29.1 & 31.1 & 32.4 & 31.3 & 29.9 & 31.8\\
\hspace*{3pt}Matcher\cite{matcher} & ICLR'24 & 42.9 & 45.8 & 34.7 & 33.6 & 38.6 & 35.0 & 44.1 & 50.6 & 33.3 & 34.3 & 70.8 & 71.2 & 44.1 & 45.1\\
\hspace*{3pt}GF-SAM\cite{gfsam} & NeurIPS'24 & 44.5 & 51.2 & 36.3 & 41.9 & 48.7 & 55.2 & 53.1 & 58.1 & 47.1 & 52.4 & 51.0 & 52.9 & 46.8 & 52.0\\
\hspace*{3pt}RNS\tableevalmark\cite{rns} & CVPR'26 & 28.6 & 37.0 & 21.7 & 28.7 & 39.3 & 43.6 & 51.4 & 61.1 & 42.3 & 52.0 & 29.2 & 29.8 & 35.4 & 42.0\\
\hspace*{3pt}PR-MaGIC\tableevalmark\cite{prmagic} & CVPR'26 & 40.2 & 41.4 & 32.6 & 32.4 & 42.4 & 41.5 & 40.3 & 43.8 & 33.3 & 34.3 & 75.3 & 74.3 & 44.0 & 44.6\\
\hspace*{3pt}RPG-SAM\tableevalmark\cite{rpgsam} & MICCAI'26 & 41.6 & 48.3 & 36.8 & 42.5 & \textbf{58.8} & 63.7 & 39.7 & 44.0 & 41.3 & 47.4 & 50.4 & 49.6 & 44.8 & 49.2\\
\midrule
\multicolumn{16}{@{}l}{\textbf{No offline task training:} \textit{Self-supervised pre-training}} \\
\hspace*{3pt}INSID3\cite{insid3} & CVPR'26 & 50.5 & 57.1 & 38.7 & 46.8 & 54.4 & 63.9 & 54.9 & 61.7 & \underline{52.1} & 56.9 & 78.8 & 80.1 & 54.9 & 61.1\\
\hspace*{3pt}HERA\tableevalmark\cite{hera} & CVPR'26 & 45.9 & 59.4 & 32.4 & 45.5 & 52.9 & 61.9 & 48.5 & 58.0 & 44.5 & 52.8 & 74.0 & 86.8 & 49.7 & 60.7\\
\hspace*{3pt}RSRM\tableevalmark\cite{rsrm} & ECCV'26 & 51.1 & 60.2 & 38.0 & 48.5 & \underline{57.0} & 63.9 & 52.4 & 62.2 & 48.9 & 57.1 & \underline{87.8} & 89.0 & 55.9 & 63.5\\
\hspace*{3pt}FROST\cite{frost} & arXiv'26 & \underline{53.1}\rlap{\tableevalmark} & \underline{64.9}\rlap{\tableevalmark} & \underline{39.1} & \underline{52.1} & 52.6 & \underline{66.3} & 58.0 & \underline{68.8} & 50.5\rlap{\tableevalmark} & \underline{61.0}\rlap{\tableevalmark} & 84.7\rlap{\tableevalmark} & 85.5\rlap{\tableevalmark} & \underline{56.3} & \underline{66.4}\\
\rowcolor{gainrow}
\hspace*{3pt}\resacell{ReSA (ours)} & \resacell{This paper} & \resacell{\textbf{56.0}} & \resacell{\textbf{66.3}} & \resacell{\textbf{45.5}} & \resacell{\textbf{57.3}} & \resacell{\textbf{58.8}} & \resacell{\textbf{68.2}} & \resacell{\textbf{58.8}} & \resacell{\textbf{69.0}} & \resacell{\textbf{53.2}} & \resacell{\textbf{66.2}} & \resacell{\textbf{89.7}} & \resacell{\textbf{90.7}} & \resacell{\textbf{60.3}} & \resacell{\textbf{69.6}}\\
\bottomrule
\end{tabular}%
\end{lrbox}
\typeout{MAIN-COMPARISON-PARTS_DOMAINS-WIDTH=\the\wd\mainpartsdomainbox}
\typeout{MAIN-COMPARISON-TEXT-WIDTH=\the\textwidth}
\begin{minipage}{\textwidth}
\UseCaptionStatus{tab:comparison_parts_domains}
\caption{\textbf{Object-part and cross-domain segmentation.} Foreground mIoU (\%) in 1-shot and 5-shot. Avg (6) weights these six datasets equally. Backbones and notation follow Table~\ref{tab:comparison_natural}. Extended comparisons and ten-dataset averages are in Supplementary Tables~S3--S4.}
\label{tab:comparison_parts_domains}
\centering
\usecomparisonbox{\mainpartsdomainbox}{\partsAvgOffset}
\end{minipage}
\end{table*}

\subsection{Reference-response consistency}
\label{sec:analysis}

We analyze a fixed, row-stochastic $K$ for episodes with both reference classes present and a valid threshold fit.
The response penalty is zero whenever $K(p-p_0)=0$.
For initial responses strictly between zero and one, its local quadratic form in probability space has rank at most $2S$.
For RMS, joint convexity of KL gives
\begin{equation}
\frac{1}{2S}\sum_r d(K_rp_\lambda,K_rq_\lambda)\leq D_\lambda.
\label{eq:response_bound}
\end{equation}
Thus, the reconstruction error upper-bounds the average KL divergence between the original and reconstructed reference responses.
Define their root-mean-square difference as $E(u,v)=\|K(u-v)\|_2/\sqrt{2S}$.
With natural logarithms, Pinsker's inequality and the triangle inequality give
\begin{equation}
E(q_\lambda,p_0)\leq
\sqrt{\mathcal L_{\rm cond}(p_\lambda)/2}+\sqrt{D_\lambda/2}.
\label{eq:observable_chain}
\end{equation}
The bound combines adaptation and reconstruction errors to control the drift from the initial responses $Kp_0$ to the reconstructed responses $Kq_\lambda$.

Let $X\sim w$ be a sampled token index and $Y\mid X=i\sim\operatorname{Ber}(p_{\lambda,i})$ a model-generated label.
Define $Z^*=\mathbf1[p_{\lambda,X}>T_\lambda]$ as the threshold group of the sampled token.
At the optimal fitted levels and for $D_{\lambda,\rm const}>0$,
\begin{equation}
s_\lambda=\frac{I(Y;Z^*)}{I(Y;X)}.
\label{eq:information_gain}
\end{equation}
Thus, $s_\lambda$ measures the fraction of token--label mutual information retained by the threshold partition.

\section{Experiments}
\label{sec:experiments}

\subsection{Evaluation setting}

\paragraph{Benchmarks and metric.}
We evaluate 1-shot and 5-shot segmentation on ten datasets: LVIS-$92^i$, COCO-$20^i$, FSS, and PASCAL-$5^i$ for natural objects; PASCAL-Part and PACO-Part for parts; and ISIC, SUIM, iSAID, and X-Ray for specialized imaging domains~\cite{lvis,coco,fss1000,voc,pascalpart,paco,isic,suim,isaid,xray}.
The eight original benchmarks follow INSID3's class and fold protocols.
FSS uses Matcher's 240 test classes and 2,400 episodes per shot setting.
PASCAL-$5^i$~\cite{oslsm} uses VOC2012/SBD annotations and FSSAM's sampling pools, with 1,000 fixed episodes per fold and shot setting~\cite{sbd}.
The evaluation covers 133,962 episodes in total.
We pool intersections and unions across episodes within each class to compute foreground IoU, then average over classes and folds.
Cross-dataset averages weight all datasets equally.

\paragraph{Implementation.}
We use a frozen DINOv3 ViT-L/16 encoder with $1024\times1024$ inputs, producing $64\times64$ tokens.
Positional debiasing uses a rank-500 basis, and each classifier has 1,025 parameters.
Reference fitting and query adaptation use separate AdamW~\cite{adamw} optimizer states, with a learning rate of $0.004$ and betas $(0.9,0.999)$.
We use the same inference configuration across datasets and shot settings.
TAR runs five iterations on a $640\times640$ grid with $\eta=8$ and a five-pixel band radius.
Full implementation, optimization, sampling, and evaluation details are provided in Supplementary Sections~A and~C.

\paragraph{Comparisons.}
We group methods by offline task training and pretraining supervision: task-trained methods, methods using mask-supervised pretraining without task training, and methods using self-supervised features without task training.
INSID3, RSRM, HERA, FROST, and \method all use DINOv3-L/16.
Other backbones include CLIP~\cite{clip} and Alpha-CLIP~\cite{alphaclip}.
Supplementary Sections~D.2--D.4 document each method's inference configuration, result sources, and evaluation conventions.

\subsection{Main results}

\method achieves average mIoU scores of 62.6\% in 1-shot and 71.8\% in 5-shot across ten datasets (Tables~\ref{tab:comparison_natural}--\ref{tab:comparison_parts_domains}).
These scores exceed INSID3 by 4.4/7.2 points and FROST by 5.4/4.3 points.
In 5-shot, \method ranks first on three of the four natural-object datasets among methods without offline task training (Table~\ref{tab:comparison_natural}).

On PASCAL-Part and PACO-Part, \method averages 50.7/61.8 mIoU in 1-shot/5-shot, outperforming INSID3 by 6.1/9.8 points (Table~\ref{tab:comparison_parts_domains}).
The 5-shot gains over INSID3 on ISIC, SUIM, iSAID, and X-Ray are 4.3, 7.3, 9.3, and 10.6 points, respectively, using the same inference configuration.

\paragraph{Foreground coverage and background rejection.}
Compared with INSID3 in 1-shot, \method reduces both foreground miss and extra-foreground rates on PASCAL-Part and X-Ray (Supplementary Fig.~S1).
On PASCAL-Part, the miss rate falls from 32.6\% to 26.8\% and the extra-foreground rate from 32.3\% to 29.2\%; the corresponding changes on X-Ray are 8.9\% to 7.4\% and 14.4\% to 3.3\%.
On COCO-$20^i$, the miss rate decreases from 27.0\% to 21.0\%, while the extra-foreground rate increases from 27.3\% to 30.9\%.

\begin{figure*}[t]
\centering
\includegraphics[width=\textwidth]{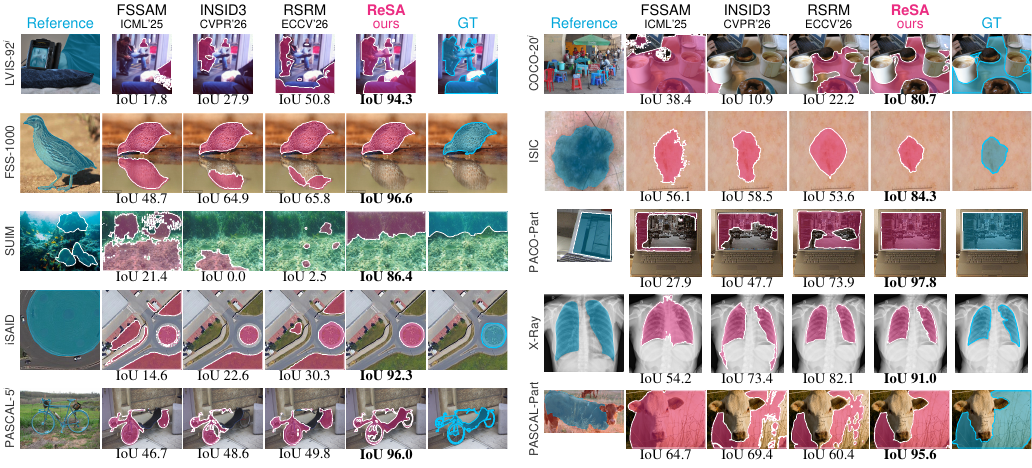}
\UseCaptionStatus{fig:qualitative}
\caption{\textbf{Five-shot qualitative comparisons across ten benchmarks.}
FSSAM (ICML'25), INSID3 (CVPR'26), RSRM (ECCV'26), and \method use the same five annotated references per query (one shown).
Cyan: reference masks and ground truth; magenta: predictions.
}
\label{fig:qualitative}
\end{figure*}

\label{sec:qualitative}
Figure~\ref{fig:qualitative} illustrates how \method follows the target extent specified by the references: it separates the bird from its reflection and segments the annotated cow torso and laptop screen.
Supplementary Fig.~S4 shows the progression from the initial prediction to more complete target masks with fewer isolated false-positive regions.

\subsection{Ablation and mechanism analysis}
\label{sec:ablation}

The full method outperforms the reference-fitted baseline (Base) by 5.3/3.9 mIoU points (Table~\ref{tab:progression}).
All ablations use the same evaluation episodes.

\begin{table}[!t]
\centering
\UseCaptionStatus{tab:progression}
\caption{\textbf{Module combinations.} Ten-dataset mean mIoU (\%). Base is $p_0$; $\Delta$ subtracts its displayed score. Bold marks the best per shot.}
\label{tab:progression}
\fontsize{8.5}{10}\selectfont
\setlength{\tabcolsep}{3pt}
\renewcommand{\arraystretch}{1.13}
\begin{tabular*}{\columnwidth}{@{\extracolsep{\fill}}*{3}{w{c}{2.4em}}*{4}{c}@{\hspace{2pt}}}
\toprule
\textbf{RCA} & \textbf{RMS} & \textbf{TAR} & \textbf{1-shot} & {\boldmath$\Delta$} & \textbf{5-shot} & {\boldmath$\Delta$}\\
\midrule
Base &  &  & 57.3 & -- & 67.9 & --\\
$\checkmark$ & -- & -- & 60.0 & $+2.7$ & 70.3 & $+2.4$\\
$\checkmark$ & $\checkmark$ & -- & 61.9 & $+4.6$ & 70.7 & $+2.8$\\
$\checkmark$ & -- & $\checkmark$ & 60.6 & $+3.3$ & 71.3 & $+3.4$\\
$\checkmark$ & $\checkmark$ & $\checkmark$ & \textbf{62.6} & $+5.3$ & \textbf{71.8} & $+3.9$\\
\bottomrule
\end{tabular*}
\end{table}

\begin{table}[!t]
\centering
\UseCaptionStatus{tab:ca_losses}
\caption{\textbf{RCA loss combinations.} Reference supervision is retained; ten-dataset mean mIoU (\%). $\Delta$ subtracts the displayed full-objective score; bold marks the best per shot.}
\label{tab:ca_losses}
\fontsize{8.5}{10}\selectfont
\setlength{\tabcolsep}{3pt}
\renewcommand{\arraystretch}{1.10}
\begin{tabular*}{\columnwidth}{@{\extracolsep{\fill}}l*{4}{c}@{\hspace{2pt}}}
\toprule
\textbf{Loss combination} & \textbf{1-shot} & {\boldmath$\Delta$} & \textbf{5-shot} & {\boldmath$\Delta$}\\
\midrule
$R$ & 56.8 & $-3.2$ & 68.3 & $-2.0$\\
$R+C$ & 55.2 & $-4.8$ & 67.5 & $-2.8$\\
$R+H$ & 53.9 & $-6.1$ & 69.7 & $-0.6$\\
$R+C+H$ & \textbf{60.0} & -- & \textbf{70.3} & --\\
\bottomrule
\end{tabular*}
\end{table}

\paragraph{Foreground retention and background rejection.}
Combining reference supervision ($R$), the response KL penalty ($C$), and query entropy ($H$) gives the highest average mIoU in both shot settings (Table~\ref{tab:ca_losses}).
All variants share $p_0$, the optimization budget, $\lambda=1$, and $T=0.5$, with RMS and TAR disabled.
The retained query losses use the same $1/S$ scaling.
In 1-shot, adding $C$ to $R+H$ reduces the foreground miss rate from 37.1\% to 26.5\%.
Adding $H$ to $R+C$ reduces the background false-positive rate from 7.8\% to 5.7\%.

RCA outperforms continued reference supervision alone ($R$-only) by 3.7/2.1 mIoU points with one candidate and the same threshold-fitting rule (Supplementary Table~S7).

\begin{figure}[!t]
\centering
\includegraphics[width=\columnwidth]{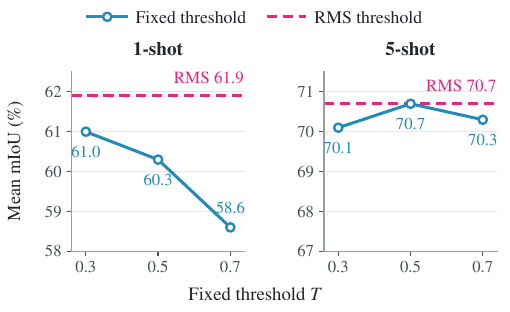}
\UseCaptionStatus{fig:thresholds}
\caption{Threshold fitting. Ten-dataset mean mIoU (\%) with RMS-selected probabilities fixed, without TAR.}
\label{fig:thresholds}
\end{figure}

\paragraph{Reference-weighted mask selection.}
Reference weighting improves average mIoU over uniform weighting by 0.3 points in 1-shot and 1.1 points in 5-shot under the same RCA objective, without TAR (Supplementary Table~S10).
RMS threshold fitting improves 1-shot mIoU by 1.6 points over $T=0.5$, reaching 61.9; both thresholds give 70.7 in 5-shot after rounding (Fig.~\ref{fig:thresholds}).

Relative-gain selection outperforms either fixed branch and random selection (Table~\ref{tab:branch_selection}).
It improves mIoU by 0.9/0.2 points over selecting the branch with the lowest reconstruction error.

\begin{table}[!t]
\centering
\UseCaptionStatus{tab:branch_selection}
\caption{\textbf{Branch-selection controls.} Ten-dataset mean mIoU (\%), without TAR, using fixed candidate probabilities and fitted thresholds. Min $D$ selects the branch with the lowest weighted reconstruction error; random selection averages 20 seeds. Bold marks the best per shot, including ties.}
\label{tab:branch_selection}
\fontsize{8.5}{10}\selectfont
\setlength{\tabcolsep}{3pt}
\renewcommand{\arraystretch}{1.10}
\begin{tabular*}{\columnwidth}{@{\extracolsep{\fill}}lcc@{\hspace{2pt}}}
\toprule
\textbf{Selection rule} & \textbf{1-shot} & \textbf{5-shot}\\
\midrule
Single branch: $\lambda=1$ & 61.5 & 70.2\\
Single branch: $\lambda=3$ & 61.0 & 70.6\\
Min $D$: $\lambda\in\{1,3\}$ & 61.0 & 70.5\\
Random $\lambda\in\{1,3\}$ & 61.2 & 70.4\\
RMS: $\lambda\in\{1,3\}$ & \textbf{61.9} & \textbf{70.7}\\
RMS: $\lambda\in\{1,2,3\}$ & 61.8 & \textbf{70.7}\\
\bottomrule
\end{tabular*}
\end{table}

\paragraph{Boundary recovery and threshold alignment.}
TAR improves mIoU by 0.7/1.1 points over the coarse mask when the input probabilities and coarse masks are held fixed.
Threshold alignment adds 0.4 points over unaligned unary scores in 1-shot; both variants round to 71.8 mIoU in 5-shot (Table~\ref{tab:controls}).
TAR also outperforms the \emph{No continuous evidence} control by 0.2/0.3 points.

Threshold alignment gives its largest Boundary IoU~\cite{boundaryiou} gain of 2.4 points when $|T-0.5|\geq0.2$ (Fig.~\ref{fig:threshold_alignment}).
The aligned and unaligned variants retain the same foreground core and exterior background, so the comparison isolates the threshold shift within the boundary band.

The 95\% benchmark-bootstrap intervals for threshold-fitting and alignment gains are above zero in 1-shot, while both include zero in 5-shot (Supplementary Table~S18).


\begin{figure}[!t]
\centering
\includegraphics[width=\columnwidth]{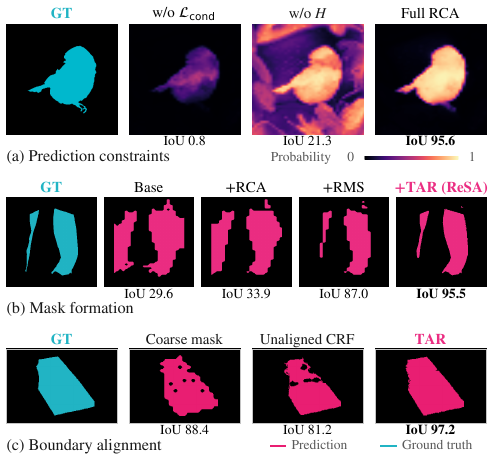}
\UseCaptionStatus{fig:module_validation}
\caption{\textbf{Changes in predictions, masks, and boundaries.}
(a) RCA probabilities with reference supervision retained: removing the response constraint, removing entropy, and using both.
(b) Masks after reference fitting (Base), then cumulative RCA, RMS, and TAR.
(c) The coarse mask, CRF without threshold alignment, and TAR.
Each row uses a common display region.
Numbers are full-query IoUs (\%); cyan/magenta denote annotations/predictions.}
\label{fig:module_validation}
\end{figure}

\begin{figure}[!t]
\centering
\includegraphics[width=\columnwidth]{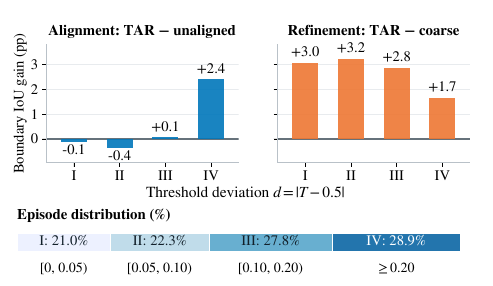}
\UseCaptionStatus{fig:threshold_alignment}
\caption{\textbf{Boundary recovery and threshold alignment.}
Boundary IoU~\cite{boundaryiou} changes across four $d=|T-0.5|$ groups: TAR--unaligned isolates alignment; TAR--coarse measures total refinement. The lower strip shows the distribution of 133,690 valid-threshold episodes. Shots are pooled within each dataset before averaging datasets.}
\label{fig:threshold_alignment}
\end{figure}

\begin{table}[!t]
\centering
\UseCaptionStatus{tab:controls}
\caption{\textbf{Matched refinement controls.} Ten-dataset mean mIoU (\%) with probabilities and coarse masks fixed. Refiners share the band, RGB/spatial terms, $\eta=8$, and five iterations. Parentheses subtract Full TAR; bold marks the best, including ties.}
\label{tab:controls}
{\fontsize{8.5}{10}\selectfont
\newcommand{\controldelta}[1]{{\fontsize{6}{7}\selectfont\ensuremath{(#1)}}}
\setlength{\tabcolsep}{2pt}
\renewcommand{\arraystretch}{1.10}
\begin{tabular*}{\linewidth}{@{\extracolsep{\fill}}lr@{\,\extracolsep{0pt}}l@{\extracolsep{\fill}}r@{\,\extracolsep{0pt}}l@{\hspace{2pt}}}
\toprule
\textbf{Configuration} & \multicolumn{2}{c}{\textbf{1-shot}} & \multicolumn{2}{c@{\hspace{2pt}}}{\textbf{5-shot}}\\
\midrule
No TAR (RCA+RMS) & 61.9 & \controldelta{-0.7} & 70.7 & \controldelta{-1.1}\\
Unaligned evidence & 62.2 & \controldelta{-0.4} & \textbf{71.8} & \controldelta{+0.0}\\
No continuous evidence & 62.4 & \controldelta{-0.2} & 71.5 & \controldelta{-0.3}\\
Full TAR & \textbf{62.6} &  & \textbf{71.8} & \\
\bottomrule
\end{tabular*}
\par}
\end{table}

\begin{table}[!t]
\centering
\UseCaptionStatus{tab:backbones}
\caption{\textbf{Matched backbone controls.} Ten-dataset mean mIoU (\%), 1-shot/5-shot. Base is first-stage $p_0$; Full uses the same encoder and episodes. Gains subtract displayed Base scores.}
\label{tab:backbones}
\fontsize{8.5}{10}\selectfont
\setlength{\tabcolsep}{3pt}
\renewcommand{\arraystretch}{1.15}
\begin{tabular*}{\columnwidth}{@{\extracolsep{\fill}}lccc@{\hspace{2pt}}}
\toprule
\textbf{Backbone} & \textbf{Base} & \textbf{Full method} & \textbf{Gain}\\
\midrule
DINOv2-L/14 & 50.5/60.8 & 57.0/67.5 & +6.5/+6.7\\
DINOv3-S/16 & 50.5/61.0 & 55.4/64.2 & +4.9/+3.2\\
DINOv3-B/16 & 53.8/64.8 & 59.6/68.9 & +5.8/+4.1\\
DINOv3-L/16 & 57.3/67.9 & 62.6/71.8 & +5.3/+3.9\\
\bottomrule
\end{tabular*}
\end{table}

\subsection{Encoder controls and inference cost}
ReSA improves on Base across DINOv3-S/B/L and DINOv2-L in both shot settings (Table~\ref{tab:backbones}).
\par

On an H100, two-branch inference takes 0.205/0.420 seconds per episode in 1-shot/5-shot, averaged over 100 episodes each.
Branches share feature extraction and reference fitting; TAR refines the selected prediction.
A third candidate takes 0.210/0.450 seconds and gives 62.6/71.7 mIoU (Supplementary Section~G).

\section{Conclusion}
\label{sec:conclusion}

We presented \method, a framework connecting query adaptation and mask selection through a shared reference kernel.
Foreground and background reference cues regularize query responses and weight probability reconstruction for threshold fitting.
Relative reconstruction gain selects a candidate whose threshold aligns CRF unary scores for boundary refinement.
Our analysis links adaptation and soft reconstruction through a bound on reference-response drift.
With a frozen encoder and lightweight adaptation, \method leads the evaluated methods across ten benchmarks at 62.6\%/71.8\% mean mIoU in 1-shot/5-shot.
Ablations quantify adaptation and selection gains; visualizations show more complete masks and fewer isolated false positives.
\label{sec:body_end}
\section*{LLM Usage Statement}
\label{sec:llm_usage}
We used OpenAI Codex to assist with methodological discussions, code development and debugging, experimental and statistical analysis, figure and table preparation, and drafting, editing, and translating the main paper and supplement. GPT-6 Pro provided pre-submission manuscript feedback.
The authors take responsibility for the methodology, results, and final content.

{\small
\bibliographystyle{ieeenat_fullname}
\bibliography{references}
}
\clearpage
\appendix
\setcounter{table}{0}
\setcounter{figure}{0}
\setcounter{equation}{0}
\renewcommand{\thetable}{S\arabic{table}}
\renewcommand{\thefigure}{S\arabic{figure}}
\renewcommand{\theequation}{S\arabic{equation}}
\renewcommand{\theHtable}{supp.\arabic{table}}
\renewcommand{\theHfigure}{supp.\arabic{figure}}
\renewcommand{\theHequation}{supp.\arabic{equation}}
\captionsetup[figure]{skip=10pt}
\renewcommand{\topfraction}{0.9}
\renewcommand{\bottomfraction}{0.8}
\renewcommand{\textfraction}{0.1}
\renewcommand{\dbltopfraction}{0.9}
\setcounter{topnumber}{2}
\setcounter{bottomnumber}{2}
\setcounter{totalnumber}{3}
\setcounter{dbltopnumber}{1}
\setlength{\textfloatsep}{10pt plus 2pt minus 2pt}
\setlength{\dbltextfloatsep}{10pt plus 2pt minus 2pt}
\setlength{\floatsep}{8pt plus 2pt minus 2pt}
\setlength{\dblfloatsep}{8pt plus 2pt minus 2pt}
\setlength{\intextsep}{8pt plus 2pt minus 2pt}

\twocolumn[{
  \centering
  {\Large\bfseries Supplementary Material\par}
  \vspace{18pt}
}]
\newcommand{\suppdelta}[1]{\,{\scriptsize\ensuremath{(#1)}}}
\newcommand{\suppdiscussion}[1]{\par\smallskip{\normalfont\normalsize\noindent #1\par}}
\section{Inference details and fixed configurations}
\label{sec:supp_inference}

ReSA carries reference guidance through adaptation and mask selection using the same kernel: RCA constrains reference responses, and RMS uses the kernel to weight threshold fitting and candidate selection. TAR uses the selected threshold for boundary refinement. Tables~\ref{tab:episodes} and~\ref{tab:supp_parameters} give the evaluation episodes and fixed inference settings.

\begin{table}[tbp]
\centering
\begin{minipage}[t]{\columnwidth}
\vspace{0pt}
\centering
\UseCaptionStatus{tab:episodes}
\caption{\textbf{Evaluation episodes per shot.}
Counts are identical in the 1-shot and 5-shot settings.
FSS uses a single test-class pool.}
\fontsize{8.5}{10}\selectfont
\setlength{\tabcolsep}{7pt}
\renewcommand{\arraystretch}{1.13}
\begin{tabular*}{\columnwidth}{@{\extracolsep{\fill}}lcc@{}}
\toprule
\multicolumn{1}{@{}l}{\textbf{Dataset}} & \textbf{Folds/pools} & \textbf{Episodes}\\
\midrule
LVIS-$92^i$ & 10 & 23,000\\
COCO-$20^i$ & 4 & 4,000\\
PASCAL-Part & 4 & 8,459\\
PACO-Part & 4 & 10,000\\
ISIC & 1 & 600\\
SUIM & 1 & 3,859\\
iSAID & 3 & 10,063\\
X-Ray & 1 & 600\\
FSS & 1 & 2,400\\
PASCAL-$5^i$ & 4 & 4,000\\
\midrule
Total & 33 & 66,981\\
\bottomrule
\end{tabular*}
\label{tab:episodes}

\end{minipage}
\end{table}

\begin{table}[tbp]
\centering
\UseCaptionStatus{tab:supp_parameters}
\caption{\textbf{Main inference parameters.} Shared across all ten datasets and both shot settings.}
\label{tab:supp_parameters}
\fontsize{8.5}{10}\selectfont
\setlength{\tabcolsep}{5pt}
\renewcommand{\arraystretch}{1.10}
\begin{tabular*}{\columnwidth}{@{\extracolsep{\fill}}ll@{}}
\toprule
\textbf{Parameter} & \textbf{Value}\\
\midrule
Frozen encoder & DINOv3 ViT-L/16\\
Input resolution & $1024\times1024$\\
Positional basis rank & $500$\\
\midrule
Optimizer & AdamW~\cite{adamw}\\
Learning rate & $0.004$\\
AdamW $(\beta_1,\beta_2)$ & $(0.9,\,0.999)$\\
Weight decay & $5\times10^{-4}$\\
Reference-fitting steps & $400$\\
RCA adaptation steps (per branch) & $100$\\
Kernel temperature $\tau$ & $0.05$\\
Entropy weights $\lambda$ & $\{1,3\}$\\
\midrule
TAR grid resolution & $640\times640$\\
Aligned-evidence strength $\eta$ & $8$\\
Boundary-band radius & $5$ CRF-grid pixels\\
TAR iterations & $5$\\
\bottomrule
\end{tabular*}
\end{table}

\subsection{Response-constrained adaptation}

\paragraph{Balanced reference supervision.}
Sampling each reference mask on the feature grid gives foreground/background sets $\mathcal R_s^+$ and $\mathcal R_s^-$.
The reference loss assigns equal weight to foreground and background classes within each reference:
\begin{equation}
\mathcal L_{\rm ref}=\frac{1}{2S}\sum_{s=1}^S
\left(\operatorname{CE}(\mathcal R_s^+)+\operatorname{CE}(\mathcal R_s^-)\right),
\label{eq:supp_reference}
\end{equation}
where $\operatorname{CE}(\mathcal R_s^+)=-|\mathcal R_s^+|^{-1}\sum_{j\in\mathcal R_s^+}\log p_{sj}$ and the background term averages $-\log(1-p_{sj})$.

\paragraph{Loss reductions and shot scaling.}
Using the main-paper definitions of the fixed kernel $K$, initial prediction $p_0$, and Bernoulli KL $d$, the query part of the RCA objective expands to
\begin{equation}
\frac{\mathcal L_{\rm cond}+\lambda\overline H}{S}
=\frac{1}{2S^2}\sum_{r=1}^{2S}d(K_rp,K_rp_0)
+\frac{\lambda}{SN}\sum_{i=1}^N H(p_i).
\label{eq:supp_loss_reductions}
\end{equation}
The inner $1/(2S)$ gives each reference equal weight, split between foreground and background independently of mask area; entropy averages the $N$ query tokens.
The outer $1/S$ scales the mean query losses relative to reference supervision. Both branches and all objective ablations retain these reductions.

\subsection{Reference-weighted mask selection}

For the main-paper RMS objective, let $w_i=(2S)^{-1}\sum_rK_{ri}$ and $q_{\lambda,i}=a+(b-a)\mathbf1[p_{\lambda,i}>t]$:
\begin{equation}
D_\lambda(t,a,b)=\sum_iw_i d(p_{\lambda,i},q_{\lambda,i}).
\label{eq:supp_pixel_readout}
\end{equation}

\paragraph{Closed-form levels and threshold search.}
For a fixed nonempty split with positive group weights, let $G_0=\{i:p_{\lambda,i}\leq t\}$ and $G_1=\{i:p_{\lambda,i}>t\}$, and define
\begin{equation}
W_g=\sum_{i\in G_g}w_i,\qquad
\mu_g=\frac{\sum_{i\in G_g}w_ip_{\lambda,i}}{W_g}.
\end{equation}
The derivative of a group's loss with respect to its level $c$ is
$W_g(c-\mu_g)/(c(1-c))$.
Convexity of Bernoulli KL in its second argument therefore gives $a^*=\mu_0$ and $b^*=\mu_1$; sorting ensures their order.
At these levels, writing $H(u)=-u\log u-(1-u)\log(1-u)$, the distortion is
\begin{equation}
D_\lambda(t)=\sum_{g=0}^1W_gH(\mu_g)-\sum_iw_iH(p_{\lambda,i}).
\end{equation}
A stable sort followed by prefix sums evaluates every nonempty split between distinct probabilities in $O(N\log N)$ time and $O(N)$ auxiliary space.
The minimum supplies $D_\lambda^*$ and a threshold $T_\lambda$ at the midpoint between the two adjacent distinct probabilities.

\paragraph{Relative gain and the constant baseline.}
For the constant family $q=c\mathbf1$, the optimum is
$\mu_\lambda=\sum_iw_ip_{\lambda,i}$, giving
\begin{equation}
D_{\lambda,\rm const}=H(\mu_\lambda)-\sum_iw_iH(p_{\lambda,i}).
\end{equation}
The relative-gain score is $s_\lambda=1-D_\lambda^*/D_{\lambda,\mathrm{const}}$. The fitted levels determine reconstruction error, while the selected pair $(p_{\lambda^*},T_{\lambda^*})$ defines the coarse mask and supplies TAR's aligned evidence. Ties select $\lambda=3$.

\subsection{Episode-local optimization}
\label{sec:supp_uniform_budgets}
Each episode initializes a zero-weight, zero-bias head and fits the reference annotations once. Both query branches copy the fitted head and reset their AdamW states. They share detached features, a fixed reference kernel, and fixed initial responses.

\subsection{Reference-mask sampling}

Reference masks are bilinearly resized to the token grid and thresholded at $0.5$, following INSID3~\cite{insid3}. Candidate matching, dense evidence, and head supervision use this same sampling.

\subsection{Numerical implementation and boundary refinement}

For AdamW, $\epsilon=10^{-8}$ and the exact weight decay is $(1-(1-0.002\cdot0.0005)^2)/0.004$; the other settings are in Table~\ref{tab:supp_parameters}.
Initial responses $Kp_0$ are clipped to $[10^{-4},1-10^{-4}]$; conditional responses and entropy probabilities are clipped to $[10^{-6},1-10^{-6}]$.

Threshold search uses float64 arithmetic, clips logarithm arguments and fitted levels to $[10^{-8},1-10^{-8}]$, and clips numerical gain to $[0,1]$. A valid threshold fit requires at least two probabilities, $\max_i p_i-\min_i p_i\geq10^{-12}$, $D_{\lambda,\rm const}>10^{-12}$, and a split with both group weights above $10^{-12}$. Equal minimum distortions select the first split in ascending probability order. If the gain-selected fit is invalid, the implementation uses an unweighted between-class variance threshold for each branch and selects the larger foreground--background mean probability difference. Constant maps use $T=0.5$; an empty threshold group gives contrast zero, and ties select $\lambda=3$.

\paragraph{Threshold-aligned boundary evidence.}
On the CRF grid specified in Table~\ref{tab:supp_parameters}, the spatial scales are $\alpha=15$ and $\gamma=5$ pixels, the color scale is $\beta=0.03$, and spatial and bilateral weights are $3$ and $20$, respectively. Potts compatibility, Frank--Wolfe step size, and $\ell_2$ regularization weight are all $1$~\cite{densecrf,fwcrf}.
Circular erosion/dilation defines the band. Foreground priors are $0.95$ in the eroded core, $0.05$ outside the dilation, and $0.5$ in the band.
Channel 0 denotes background and channel 1 foreground; the corresponding input scores are the negative and positive logit of the prior.
TAR retains these scores outside the band and sets its band scores to $(\ell_x^{\rm bg},\ell_x^{\rm fg})=(-\eta\widetilde z_x/2,+\eta\widetilde z_x/2)$, where $\widetilde z$ is the interpolated aligned logit.
The solver subtracts these scores in its unary energy, so positive $\widetilde z_x$ favors foreground. The final band label is foreground when the channel-1 output exceeds channel 0.
The probability and selected threshold are clipped to $[10^{-6},1-10^{-6}]$ before taking logits.
After optimization, the core foreground and exterior background are explicitly restored.

\subsection{Reference-based foreground check}

Reference--query correspondence identifies candidate foreground tokens before probability fitting.
For a query token, the foreground and background matching scores are log-mean-exp similarities to the respective reference token sets, with inverse temperature $\beta=8$.
A reference votes foreground when its foreground score is larger.
Majority voting across references is intersected with positive similarity to the equally weighted reference foreground prototype. If no query token has positive similarity, the prototype gate instead retains similarities above their 90th percentile. A reference containing only one class on the token grid votes for that class.

If the candidate set is empty, the coarse mask is empty. If candidates exist but any reference lacks foreground or background tokens, reference-conditioned region aggregation produces the coarse mask. These cases bypass head adaptation and threshold fitting and use neutral-evidence CRF refinement with a five-pixel band radius and ten iterations. Their predictions are included in the full evaluation.

\subsection{Execution and computational cost}

The two RCA branches share feature extraction and reference fitting; each branch adds query adaptation and $O(N\log N)$ threshold search. TAR runs once on the selected output. All-pairs correspondence scales quadratically with token count; Section~\ref{sec:supp_inference_efficiency} reports measured runtimes.

\section{Mathematical analysis}
\label{sec:supp_analysis}

Throughout, $K$ is fixed and row-stochastic, $w_i=(2S)^{-1}\sum_rK_{ri}$, and logarithms are natural. The derivations assume foreground and background reference tokens and exact probability arithmetic; numerical clipping is specified in Section~A.5.

\subsection{Adaptation: constrain reference responses, allow local changes}

\paragraph{Complementary gradients of entropy and conditional KL.}
For the fixed reference kernel $K$, writing $\ell=\operatorname{logit}$, the probability-space gradients for interior probabilities are
\begin{equation}
\begin{aligned}
\nabla_p\overline H&=-\frac1N\ell(p),\\
\nabla_p\mathcal L_{\rm cond}&=\frac1{2S}K^\top\!\left[\ell(Kp)-\ell(Kp_0)\right].
\end{aligned}
\label{eq:supp_ca_gradients}
\end{equation}
Entropy sharpens individual probabilities; conditional KL responds to drift in their reference-weighted averages. The shared classifier couples token updates. Section~\ref{sec:supp_ca_controls} tests their complementary effects on foreground retention and background rejection.

\paragraph{Local geometry of the adaptation constraint.}
Let $\Delta p=p-p_0$ and $z_0=Kp_0$, with all $z_{0,r}\in(0,1)$.
If $K\Delta p=0$, the conditional penalty is exactly zero, even when token probabilities change.
For small reference-response changes, a Taylor expansion in the first argument of Bernoulli KL gives
\begin{equation}
\begin{aligned}
\mathcal L_{\rm cond}(p)
&=\frac{1}{4S}\Delta p^\top K^\top M_0 K\Delta p
  +o(\|K\Delta p\|_2^2),\\
M_0&=\operatorname{diag}\!\left(\frac{1}{z_{0,r}(1-z_{0,r})}\right).
\end{aligned}
\label{eq:supp_local_geometry}
\end{equation}
The positive semidefinite matrix $K^\top M_0 K$ has rank at most $\min(2S,N)$ and shares the null space of $K$.
Within the updates realizable by the classifier, RCA penalizes reference-response drift and permits local changes that preserve those responses.

\subsection{Mask selection: control reference-response changes}

Bernoulli KL is the Bregman divergence generated by $\phi(p)=p\log p+(1-p)\log(1-p)$; RMS is a reference-weighted, two-group instance of Bregman quantization~\cite{banerjee2005}.

\paragraph{Reference-weighted reconstruction and response error.}
For any row-stochastic $K$, Bernoulli mixtures satisfy
$\sum_iK_{ri}\operatorname{Ber}(p_i)=\operatorname{Ber}(K_rp)$.
Joint convexity of KL implies, for each reference prototype,
\begin{equation}
d(K_rp,K_rq)\leq\sum_iK_{ri}d(p_i,q_i).
\end{equation}
Averaging over the $2S$ prototypes yields
\begin{equation}
\frac{1}{2S}\sum_r d(K_rp,K_rq)\leq\sum_iw_id(p_i,q_i).
\end{equation}
The RMS objective bounds changes in the same reference responses constrained by RCA, connecting adaptation to threshold fitting through $K$.

\paragraph{Combining adaptation and reconstruction errors.}
Define the root-mean-square reference-response difference
\begin{equation}
E(u,v)=\left[\frac{1}{2S}\sum_r(K_ru-K_rv)^2\right]^{1/2}.
\end{equation}
Bernoulli Pinsker gives $d(x,y)\geq2(x-y)^2$.
Applying this inequality to the RCA penalty and the preceding reconstruction bound yields
\begin{equation}
\begin{aligned}
E(p_\lambda,p_0)&\leq\sqrt{\mathcal L_{\rm cond}(p_\lambda)/2},\\
E(q_\lambda,p_\lambda)&\leq\sqrt{D_\lambda/2}.
\end{aligned}
\end{equation}
The Euclidean triangle inequality after applying $K$ then gives
\begin{equation}
E(q_\lambda,p_0)\leq
\sqrt{\mathcal L_{\rm cond}(p_\lambda)/2}+\sqrt{D_\lambda/2}.
\label{eq:supp_observable_chain}
\end{equation}
Equation~\eqref{eq:supp_observable_chain} bounds reference-response drift across adaptation and two-level reconstruction. The RCA penalty bounds the change from $Kp_0$ to $Kp_\lambda$, and the RMS reconstruction error bounds the change from $Kp_\lambda$ to $Kq_\lambda$, both through the same reference kernel.

\subsection{Branch comparison: relative information retention}

\paragraph{Information interpretation of relative gain.}
For one fixed branch, define a random token $X\sim w$, a model-generated label $Y\mid X=i\sim\operatorname{Ber}(p_{\lambda,i})$, and the threshold group $Z=\mathbf1[p_{\lambda,X}>t]$.
At the optimal group levels, $q_{\lambda,i}=\Pr(Y=1\mid Z=Z(i))$.
The expected cross-entropy with these levels is $H(Y\mid Z)$, whereas the original expected Bernoulli entropy is $H(Y\mid X)$; hence
\begin{equation}
\begin{aligned}
D_\lambda(t)&=H(Y\mid Z)-H(Y\mid X),\\
D_{\lambda,\rm const}&=H(Y)-H(Y\mid X)=I(Y;X).
\end{aligned}
\end{equation}
For the minimizing partition $Z^*$ and positive constant distortion,
\begin{equation}
s_\lambda
=\frac{D_{\lambda,\rm const}-D_\lambda^*}{D_{\lambda,\rm const}}
=\frac{I(Y;Z^*)}{I(Y;X)}.
\label{eq:supp_information_gain}
\end{equation}
Thus, $s_\lambda$ measures the fraction of each branch's model-generated label information retained by its threshold partition. Normalization compares each partition against its branch-specific constant baseline. Since $Z^*$ is a function of $X$, $s_\lambda\in[0,1]$.

\section{Evaluation protocol}

\subsection{Datasets, folds, and aggregation}

The eight INSID3-protocol datasets contribute 60,581 episodes per shot across 28 folds; FSS and PASCAL-$5^i$ complete the counts in Table~\ref{tab:episodes}.

For a fold $f$ with classes $\mathcal C_f$, let $I_e$ and $U_e$ be foreground intersection and union for episode $e$ after excluding ignored pixels.
The reported metric is
\begin{equation}
\operatorname{mIoU}_f=
\frac{100}{|\mathcal C_f|}\sum_{c\in\mathcal C_f}
\frac{\sum_{e\in\mathcal E_{f,c}}I_e}
     {\max(\sum_{e\in\mathcal E_{f,c}}U_e,1)}.
\end{equation}
Dataset scores average predefined folds equally.
Avg in each main benchmark table equally weights its displayed datasets (four natural-object datasets or six part/cross-domain datasets). The supplementary benchmark tables and ablation means equally weight all ten datasets.
The supplementary group-wise table separately averages the four natural-object, two part, and four imaging-domain datasets.
Task-training-overlap scores remain included in the relevant means.

The original eight datasets retain their frozen INSID3-based sampling and $1024\times1024$ scoring.
FSS scores at $518\times518$.
PASCAL-$5^i$ restores the prediction to the original annotation resolution and ignores label 255.
Encoder inputs and the CRF grid follow Table~\ref{tab:supp_parameters}; the DINOv2 control uses the matched token grid specified in Section~\ref{sec:supp_ablations}. Comparison methods retain their native input conventions below.

\subsection{FSS}

FSS follows the public Matcher test split and sampling convention~\cite{matcher,fss1000}.
There are 240 test classes with ten query images each.
Support sampling uses seed zero and excludes the query image.
The five-reference sampler permits repeated support images, as in the adopted public code.
The \method and official-code INSID3 evaluations use the same frozen episode lists.
Ground-truth masks use nearest-neighbor resizing; binary predictions are bilinearly resized to the scoring grid and thresholded at $0.5$.

\subsection{\texorpdfstring{PASCAL-$5^i$}{PASCAL-5i}}

PASCAL-$5^i$~\cite{oslsm} uses the four-fold validation pools from the official FSSAM implementation, VOC2012 images, and augmented SBD masks~\cite{fssam,voc,sbd}.
The 20 classes are partitioned into five classes per fold.
A single-process sampler freezes 1,000 tasks per fold and shot using seed 321, excluding the query from its support set, removing duplicate support images, and preserving the adopted support-input reversal.

\section{Comparison methods and model size}

\paragraph{Benchmark conventions.}
Benchmark tables combine published results with marked local evaluations.
Local runs use the shared episode and scoring protocol with method-specific preprocessing and adaptation settings; published entries retain their source protocols.
A method-level dagger marks a locally evaluated row; a cell-level dagger marks that measurement. Rankings use displayed scores and include ties. Gray overlap cells contribute to averages but are excluded from ranking, as are averages containing them.

\subsection{Model parameter counts}

Table~1 in the main paper reports total model parameters, including frozen backbones and auxiliary networks.
Totals come from checkpoints, instantiated modules, or published model counts~\cite{insid3,fssam,sansa}; registered buffers are excluded from local counts.
Values are rounded to the nearest million.

The DINOv3-L backbone has 303.23M parameters.
The two episode-local heads in \method have 1,025 parameters each (2,050 adapted coefficients in total) and share this frozen backbone.
HERA's 312M total includes its 8.40M fusion head, and RNS includes SAM2.1-L~\cite{sam2} with its text branch disabled.
FSSAM uses DINOv2-B and SAM2-S; SANSA uses SAM2-L.
GSM+Matcher uses DINOv2-L and SAM-H; semantic mining adds no learned network.
PerSAM uses DINOv2-L localization with SAM-H in both shot settings (945M parameters).
LDIS-1 includes the SD1.5 U-Net and VAE~\cite{ldm} plus the AlphaCLIP-L/14 visual encoder~\cite{alphaclip} (1,248.13M in total); the unused text tower is excluded.
COSINE includes DINOv2-L/14, CLIP ConvNeXt-L~\cite{clip,convnext}, and its learned segmentation modules (690.94M); shared module aliases are counted once.
FROST uses only DINOv3-L.
RPG-SAM combines DINOv2-L (304.37M) and the original SAM2 Hiera-L (224.43M), totaling 528.80M parameters.

\begin{table*}[tbp]
\centering
\newsavebox{\suppcomparisononeshotbox}
\begin{lrbox}{\suppcomparisononeshotbox}%
\fontsize{8.5}{10}\selectfont%
\setlength{\tabcolsep}{1.65pt}%
\renewcommand{\arraystretch}{1.00}%
\begin{tabular}{@{}l@{\tablegapfill{7.7pt}\hspace{7.7pt}}l>{\centering\arraybackslash}p{29.53192pt}>{\centering\arraybackslash}p{29.53192pt}>{\centering\arraybackslash}p{29.53192pt}>{\centering\arraybackslash}p{36.33192pt}>{\centering\arraybackslash}p{36.33192pt}>{\centering\arraybackslash}p{29.53192pt}>{\centering\arraybackslash}p{29.53192pt}>{\centering\arraybackslash}p{29.53192pt}>{\centering\arraybackslash}p{29.53192pt}>{\centering\arraybackslash}p{29.53192pt}@{\avgdividerfill\hspace{4pt}{\color{black}\vrule width0.35pt}\hspace{2pt}}>{\centering\arraybackslash}p{29.53192pt}@{}}%
\toprule
\multirow{3}{*}[-2.7pt]{\textbf{Method}} & \multirow{3}{*}[-2.7pt]{\textbf{Venue}} & \multicolumn{4}{c}{\textbf{Natural objects}} & \multicolumn{2}{c}{\textbf{Object parts}} & \multicolumn{4}{c}{\textbf{Cross-domain}} & \multirow{3}{*}[-2.7pt]{\textbf{Avg}} \\
\cmidrule(lr){3-6}\cmidrule(lr){7-8}\cmidrule(lr){9-12}
 &  & \textbf{LVIS-} & \textbf{COCO-} & \textbf{FSS-} & \textbf{PASCAL-} & \textbf{PASCAL-} & \textbf{PACO-} & \multirow{2}{*}{\textbf{ISIC}} & \multirow{2}{*}{\textbf{SUIM}} & \multirow{2}{*}{\textbf{iSAID}} & \multirow{2}{*}{\textbf{X-Ray}} &  \\
 & & \textbf{92$^i$} & \textbf{20$^i$} & \textbf{1000} & \textbf{5$^i$} & \textbf{Part} & \textbf{Part} & & & & & \\
\midrule
\multicolumn{13}{@{}l}{\textbf{Task-specific fine-tuning:} \textit{Semantic + mask supervision}} \\
\hspace{2pt}SegGPT\cite{seggpt} & ICCV'23 & 18.6 & \textcolor{trainoverlap}{56.1} & 85.6 & \textcolor{trainoverlap}{83.2} & 35.8 & \textcolor{trainoverlap}{13.5} & 37.5 & 34.9 & \textcolor{trainoverlap}{30.9} & 87.5 & \textcolor{trainoverlap}{48.4}\\
\hspace{2pt}SegIC\cite{segic} & ECCV'24 & \textcolor{trainoverlap}{44.6} & \textcolor{trainoverlap}{76.1} & 86.8 & \textcolor{trainoverlap}{85.6} & 39.9 & 25.9 & 25.3 & 52.5 & 46.1 & 34.5 & \textcolor{trainoverlap}{51.7}\\
\hspace{2pt}SINE\cite{sine} & NeurIPS'24 & 31.2 & \textcolor{trainoverlap}{64.5} & 84.3\rlap{\tableevalmark} & \textcolor{trainoverlap}{85.4} & 36.2 & 23.3 & 25.8 & 50.7 & 38.3 & 39.8 & \textcolor{trainoverlap}{48.0}\\
\hspace{2pt}DiffewS\cite{diffews} & NeurIPS'24 & 31.4 & \textcolor{trainoverlap}{71.3} & 87.8 & \textcolor{trainoverlap}{88.3} & 34.0 & 22.8 & 27.8 & 48.9 & 47.5 & 41.6 & \textcolor{trainoverlap}{50.1}\\
\hspace{2pt}LDIS-1\tableevalmark\cite{ldis} & AAAI'25 & 15.9 & \textcolor{trainoverlap}{57.7} & 67.5 & \textcolor{trainoverlap}{84.5} & 33.2 & 18.1 & 24.2 & 25.1 & 20.4 & 35.1 & \textcolor{trainoverlap}{38.1}\\
\hspace{2pt}FSSAM\tableevalmark\cite{fssam} & ICML'25 & 37.2 & \textcolor{trainoverlap}{62.3} & \textbf{88.7} & \textcolor{trainoverlap}{79.5} & 43.8 & 33.0 & 53.9 & 56.7 & 50.0 & 80.5 & \textcolor{trainoverlap}{58.6}\\
\hspace{2pt}COSINE\tableevalmark\cite{cosine} & ICCV'25 & \textcolor{trainoverlap}{35.3} & \textcolor{trainoverlap}{68.0} & 87.2 & \textcolor{trainoverlap}{86.2} & 36.2 & \textcolor{trainoverlap}{27.7} & 28.1 & \underline{58.7} & 36.4 & 30.3 & \textcolor{trainoverlap}{49.4}\\
\hspace{2pt}SANSA\cite{sansa} & NeurIPS'25 & 36.2 & \textcolor{trainoverlap}{76.4} & 88.2 & -- & 40.4 & 29.2 & -- & -- & -- & -- & --\\
\midrule
\multicolumn{13}{@{}l}{\textbf{No offline task training:} \textit{Mask-supervised pre-training}} \\
\hspace{2pt}PerSAM\tableevalmark\cite{persam} & ICLR'24 & 20.1 & 34.3 & 72.1 & 50.6 & 31.6 & 25.0 & 35.5 & 26.0 & 29.1 & 32.4 & 35.7\\
\hspace{2pt}Matcher\cite{matcher} & ICLR'24 & 33.0 & 52.7 & 87.0 & 68.1 & 42.9 & 34.7 & 38.6 & 44.1 & 33.3 & 70.8 & 50.5\\
\hspace{2pt}GF-SAM v2\cite{gfsam} & NeurIPS'24 & 35.2 & \underline{58.7} & 88.0 & \underline{72.1} & 44.5 & 36.3 & 48.7 & 53.1 & 47.1 & 51.0 & 53.5\\
\hspace{2pt}GF-SAM v3\cite{gfsam} & NeurIPS'24 & 31.8 & 54.8 & \underline{88.3}\rlap{\tableevalmark} & 71.8\rlap{\tableevalmark} & 44.9 & 34.4 & 50.9 & 50.5 & 46.7 & 56.1 & 53.0\\
\hspace{2pt}GSM+Matcher\cite{gsm} & AAAI'26 & 34.3 & 55.9 & 87.3 & 71.3 & -- & -- & 42.7 & -- & 36.8 & -- & --\\
\hspace{2pt}RNS\tableevalmark\cite{rns} & CVPR'26 & 27.1 & 45.1 & 81.6 & 62.8 & 28.6 & 21.7 & 39.3 & 51.4 & 42.3 & 29.2 & 42.9\\
\hspace{2pt}PR-MaGIC\tableevalmark\cite{prmagic} & CVPR'26 & 23.8 & 44.5 & 86.6 & 63.8 & 40.2 & 32.6 & 42.4 & 40.3 & 33.3 & 75.3 & 48.3\\
\hspace{2pt}RPG-SAM\tableevalmark\cite{rpgsam} & MICCAI'26 & 28.4 & 47.7 & 75.6 & 55.5 & 41.6 & 36.8 & \textbf{58.8} & 39.7 & 41.3 & 50.4 & 47.6\\
\midrule
\multicolumn{13}{@{}l}{\textbf{No offline task training:} \textit{Self-supervised pre-training}} \\
\hspace{2pt}INSID3\cite{insid3} & CVPR'26 & \underline{41.8} & 57.6 & 85.8\rlap{\tableevalmark} & 67.3\rlap{\tableevalmark} & 50.5 & 38.7 & 54.4 & 54.9 & \underline{52.1} & 78.8 & \underline{58.2}\\
\hspace{2pt}HERA\tableevalmark\cite{hera} & CVPR'26 & 26.0 & 45.3 & 80.2 & 61.6 & 45.9 & 32.4 & 52.9 & 48.5 & 44.5 & 74.0 & 51.1\\
\hspace{2pt}RSRM\tableevalmark\cite{rsrm} & ECCV'26 & 31.3 & 51.6 & 80.6 & 68.8 & 51.1 & 38.0 & \underline{57.0} & 52.4 & 48.9 & \underline{87.8} & 56.8\\
\hspace{2pt}FROST\cite{frost} & arXiv'26 & 32.8 & 48.4 & 82.6\rlap{\tableevalmark} & 70.1 & \underline{53.1}\rlap{\tableevalmark} & \underline{39.1} & 52.6 & 58.0 & 50.5\rlap{\tableevalmark} & 84.7\rlap{\tableevalmark} & 57.2\\
\rowcolor{gainrow}
\multicolumn{1}{@{}>{\columncolor{gainrow}[0pt][\tabcolsep]}l}{\resacell{\hspace{2pt}ReSA (ours)}} & \resacell{This paper} & \resacell{\textbf{44.9}} & \resacell{\textbf{58.8}} & \resacell{86.4} & \resacell{\textbf{73.8}} & \resacell{\textbf{56.0}} & \resacell{\textbf{45.5}} & \resacell{\textbf{58.8}} & \resacell{\textbf{58.8}} & \resacell{\textbf{53.2}} & \resacell{\textbf{89.7}} & \multicolumn{1}{@{}>{\columncolor{gainrow}[0pt][0pt]}c@{}}{\makebox[29.53192pt][l]{\rlap{\hspace{-2.35pt}\avgdividerstroke}\makebox[29.53192pt][c]{\resacell{\textbf{62.6}}}}}\\
\hspace{2pt}\gainentry{Gain\,$\uparrow$} &  &  &  &  &  &  &  &  &  &  &  & \gainentry{+4.4}\\
\bottomrule
\end{tabular}%
\end{lrbox}
\typeout{SUPP-COMPARISON-1-WIDTH=\the\wd\suppcomparisononeshotbox}
\typeout{SUPP-COMPARISON-TEXT-WIDTH=\the\textwidth}
\begin{minipage}{\textwidth}
\UseCaptionStatus{tab:supp_comparison_1shot}
\caption{\textbf{1-shot comparison across all ten datasets.} Foreground mIoU (\%). GF-SAM v2/v3 use DINOv2/v3 backbones. Avg weights all ten datasets equally; -- denotes unavailable scores or an incomplete average. Gray cells indicate task-training overlap (possible overlap for COSINE); they and averages containing them are excluded from ranking. Best/second best are \textbf{bold}/\underline{underlined}, including displayed ties. $^\dagger$: local evaluations (the whole row when attached to a method). Gain is the Avg margin over the best eligible competitor.}
\label{tab:supp_comparison_1shot}
\centering
\usebox{\suppcomparisononeshotbox}
\end{minipage}
\end{table*}
\subsection{Task-trained methods}

\paragraph{FSSAM.}
All twenty FSSAM scores are local official-code evaluations~\cite{fssam}.
COCO-$20^i$ uses the corresponding fold-trained checkpoint; other datasets use the COCO fold-0 checkpoint.
COCO-$20^i$ and PASCAL-$5^i$ are marked for task-training overlap; PASCAL shares target categories with the COCO training split.

\paragraph{LDIS-1.}
All twenty LDIS-1 scores are local evaluations of the released one-step model~\cite{ldis}.
We use the released LDIS-1 U-Net and the Stable Diffusion v1.5 VAE~\cite{ldm}\footnote{Stable Diffusion v1.5 model card: \url{https://huggingface.co/stable-diffusion-v1-5/stable-diffusion-v1-5}.} with AlphaCLIP visual conditioning~\cite{alphaclip}, preserving the native 1-shot path.
Our 5-shot extension concatenates the five references' conditioning tokens before a single query prediction.
Offline task training uses PASCAL, COCO, DAVIS16~\cite{davis}, and VSPW~\cite{vspw}, so the COCO-$20^i$ and PASCAL-$5^i$ cells and overall averages are gray and excluded from ranking.

\paragraph{COSINE.}
All twenty scores are local evaluations of the public multi-scale FSS 49-epoch checkpoint~\cite{cosine}.
The visual-prompt configuration combines frozen DINOv2-L/14 and CLIP ConvNeXt-L encoders~\cite{clip,convnext} with a task-trained segmentation decoder.
The CLIP encoder uses OpenCLIP~\cite{openclip}, with model \texttt{convnext\_large\_d\_320} and pretrained tag \texttt{laion2b\_s29b\_b131k\_ft\_soup}~\cite{openclipconvnext}.
We retain the native multi-reference semantic-prompt average, first-reference identity prompts, fixed reference order, and score threshold $0.7$.
The released training script lists COCO, PACO-LVIS, and Objects365-SAM~\cite{objects365,sam}; the checkpoint-specific training manifest is unavailable.
We mark LVIS-$92^i$, COCO-$20^i$, PASCAL-$5^i$, PACO-Part, and their averages as possible task-training overlap and exclude them from ranking; PASCAL-$5^i$ shares target categories with COCO.

\paragraph{SANSA.}
SANSA uses the generalist SAM2-L/Hiera-Large architecture trained on COCO and ADE20K~\cite{sansa,ade20k}.
The available entries are published results; COCO-$20^i$ cells are marked for task-training overlap.

\paragraph{Other task-trained entries.}
SINE's two FSS cells are local official-code evaluations~\cite{sine}.
DiffewS's FSS scores use the baseline results reported by SANSA~\cite{diffews,sansa}.
SegIC retains published 1-shot scores, with PASCAL-$5^i$ taken from a secondary evaluation~\cite{segic,frost}.
Its ten 5-shot entries are local evaluations using the public DINOv2-L 80k$\times$12e checkpoint.
Our five-reference extension averages the probabilities from five native single-reference passes at the query resolution and applies a fixed threshold of $0.5$.
SegGPT and DiffewS PASCAL-$5^i$ cells are gray because of task training on PASCAL VOC~\cite{seggpt,diffews}.

\begin{table*}[tbp]
\centering
\newsavebox{\suppcomparisonfiveshotbox}
\begin{lrbox}{\suppcomparisonfiveshotbox}%
\fontsize{8.5}{10}\selectfont%
\setlength{\tabcolsep}{1.65pt}%
\renewcommand{\arraystretch}{1.00}%
\begin{tabular}{@{}l@{\tablegapfill{7.7pt}\hspace{7.7pt}}l>{\centering\arraybackslash}p{29.53192pt}>{\centering\arraybackslash}p{29.53192pt}>{\centering\arraybackslash}p{29.53192pt}>{\centering\arraybackslash}p{36.33192pt}>{\centering\arraybackslash}p{36.33192pt}>{\centering\arraybackslash}p{29.53192pt}>{\centering\arraybackslash}p{29.53192pt}>{\centering\arraybackslash}p{29.53192pt}>{\centering\arraybackslash}p{29.53192pt}>{\centering\arraybackslash}p{29.53192pt}@{\avgdividerfill\hspace{4pt}{\color{black}\vrule width0.35pt}\hspace{2pt}}>{\centering\arraybackslash}p{29.53192pt}@{}}%
\toprule
\multirow{3}{*}[-2.7pt]{\textbf{Method}} & \multirow{3}{*}[-2.7pt]{\textbf{Venue}} & \multicolumn{4}{c}{\textbf{Natural objects}} & \multicolumn{2}{c}{\textbf{Object parts}} & \multicolumn{4}{c}{\textbf{Cross-domain}} & \multirow{3}{*}[-2.7pt]{\textbf{Avg}} \\
\cmidrule(lr){3-6}\cmidrule(lr){7-8}\cmidrule(lr){9-12}
 &  & \textbf{LVIS-} & \textbf{COCO-} & \textbf{FSS-} & \textbf{PASCAL-} & \textbf{PASCAL-} & \textbf{PACO-} & \multirow{2}{*}{\textbf{ISIC}} & \multirow{2}{*}{\textbf{SUIM}} & \multirow{2}{*}{\textbf{iSAID}} & \multirow{2}{*}{\textbf{X-Ray}} &  \\
 & & \textbf{92$^i$} & \textbf{20$^i$} & \textbf{1000} & \textbf{5$^i$} & \textbf{Part} & \textbf{Part} & & & & & \\
\midrule
\multicolumn{13}{@{}l}{\textbf{Task-specific fine-tuning:} \textit{Semantic + mask supervision}} \\
\hspace{2pt}SegGPT\cite{seggpt} & ICCV'23 & 25.4 & \textcolor{trainoverlap}{67.9} & 89.3 & \textcolor{trainoverlap}{89.8} & 42.8 & \textcolor{trainoverlap}{14.1} & 45.2 & 33.7 & \textcolor{trainoverlap}{35.9} & \underline{89.1} & \textcolor{trainoverlap}{53.3}\\
\hspace{2pt}SegIC\tableevalmark\cite{segic} & ECCV'24 & \textcolor{trainoverlap}{46.8} & \textcolor{trainoverlap}{74.4} & 87.3 & \textcolor{trainoverlap}{88.4} & 38.9 & 30.5 & 29.1 & 64.5 & 51.0 & 29.6 & \textcolor{trainoverlap}{54.0}\\
\hspace{2pt}SINE\cite{sine} & NeurIPS'24 & 35.5 & \textcolor{trainoverlap}{66.1} & 85.0\rlap{\tableevalmark} & \textcolor{trainoverlap}{86.2} & 36.4 & 25.4 & 28.6 & 54.8 & 40.5 & 40.6 & \textcolor{trainoverlap}{49.9}\\
\hspace{2pt}DiffewS\cite{diffews} & NeurIPS'24 & 35.4 & \textcolor{trainoverlap}{72.2} & 88.0 & \textcolor{trainoverlap}{87.8} & 39.7 & 26.1 & 32.7 & 49.8 & 48.0 & 45.1 & \textcolor{trainoverlap}{52.5}\\
\hspace{2pt}LDIS-1\tableevalmark\cite{ldis} & AAAI'25 & 20.4 & \textcolor{trainoverlap}{63.8} & 69.3 & \textcolor{trainoverlap}{86.0} & 32.9 & 20.7 & 24.5 & 26.8 & 21.8 & 34.7 & \textcolor{trainoverlap}{40.1}\\
\hspace{2pt}FSSAM\tableevalmark\cite{fssam} & ICML'25 & \underline{47.7} & \textcolor{trainoverlap}{71.0} & \textbf{89.6} & \textcolor{trainoverlap}{86.3} & 52.7 & 38.3 & 65.1 & 61.9 & 53.5 & 70.0 & \textcolor{trainoverlap}{63.6}\\
\hspace{2pt}COSINE\tableevalmark\cite{cosine} & ICCV'25 & \textcolor{trainoverlap}{40.8} & \textcolor{trainoverlap}{71.5} & 87.1 & \textcolor{trainoverlap}{88.3} & 36.0 & \textcolor{trainoverlap}{28.9} & 29.0 & 61.9 & 36.9 & 30.0 & \textcolor{trainoverlap}{51.0}\\
\hspace{2pt}SANSA\cite{sansa} & NeurIPS'25 & 42.5 & \textcolor{trainoverlap}{77.1} & 89.0 & -- & -- & -- & -- & -- & -- & -- & --\\
\midrule
\multicolumn{13}{@{}l}{\textbf{No offline task training:} \textit{Mask-supervised pre-training}} \\
\hspace{2pt}PerSAM\tableevalmark\cite{persam} & ICLR'24 & 24.3 & 39.1 & 72.8 & 56.0 & 32.5 & 28.5 & 38.9 & 28.8 & 31.1 & 31.3 & 38.3\\
\hspace{2pt}Matcher\cite{matcher} & ICLR'24 & 40.0 & 60.7 & \textbf{89.6} & 74.0 & 45.8 & 33.6 & 35.0 & 50.6 & 34.3 & 71.2 & 53.5\\
\hspace{2pt}GF-SAM v2\cite{gfsam} & NeurIPS'24 & 44.2 & \underline{66.8} & 88.9 & \underline{82.6} & 51.2 & 41.9 & 55.2 & 58.1 & 52.4 & 52.9 & 59.4\\
\hspace{2pt}GF-SAM v3\cite{gfsam} & NeurIPS'24 & 42.8 & 64.4 & 89.1\rlap{\tableevalmark} & 82.1\rlap{\tableevalmark} & 50.3 & 39.3 & 56.7 & 58.6 & 53.2 & 54.7 & 59.1\\
\hspace{2pt}GSM+Matcher\cite{gsm} & AAAI'26 & 41.9 & 62.7 & \textbf{89.6} & 75.7 & -- & -- & 44.5 & -- & 39.1 & -- & --\\
\hspace{2pt}RNS\tableevalmark\cite{rns} & CVPR'26 & 34.4 & 58.8 & 87.0 & 81.6 & 37.0 & 28.7 & 43.6 & 61.1 & 52.0 & 29.8 & 51.4\\
\hspace{2pt}PR-MaGIC\tableevalmark\cite{prmagic} & CVPR'26 & 24.5 & 48.2 & 88.6 & 69.8 & 41.4 & 32.4 & 41.5 & 43.8 & 34.3 & 74.3 & 49.9\\
\hspace{2pt}RPG-SAM\tableevalmark\cite{rpgsam} & MICCAI'26 & 39.7 & 53.1 & 77.9 & 64.0 & 48.3 & 42.5 & 63.7 & 44.0 & 47.4 & 49.6 & 53.0\\
\midrule
\multicolumn{13}{@{}l}{\textbf{No offline task training:} \textit{Self-supervised pre-training}} \\
\hspace{2pt}INSID3\cite{insid3} & CVPR'26 & 47.2 & 65.1 & 87.4\rlap{\tableevalmark} & 80.1\rlap{\tableevalmark} & 57.1 & 46.8 & 63.9 & 61.7 & 56.9 & 80.1 & 64.6\\
\hspace{2pt}HERA\tableevalmark\cite{hera} & CVPR'26 & 33.2 & 54.0 & 82.5 & 74.7 & 59.4 & 45.5 & 61.9 & 58.0 & 52.8 & 86.8 & 60.9\\
\hspace{2pt}RSRM\tableevalmark\cite{rsrm} & ECCV'26 & 39.9 & 61.8 & 82.8 & 79.4 & 60.2 & 48.5 & 63.9 & 62.2 & 57.1 & 89.0 & 64.5\\
\hspace{2pt}FROST\cite{frost} & arXiv'26 & 47.1 & 60.5 & 86.0\rlap{\tableevalmark} & 82.5 & \underline{64.9}\rlap{\tableevalmark} & \underline{52.1} & \underline{66.3} & \underline{68.8} & \underline{61.0}\rlap{\tableevalmark} & 85.5\rlap{\tableevalmark} & \underline{67.5}\\
\rowcolor{gainrow}
\multicolumn{1}{@{}>{\columncolor{gainrow}[0pt][\tabcolsep]}l}{\resacell{\hspace{2pt}ReSA (ours)}} & \resacell{This paper} & \resacell{\textbf{53.0}} & \resacell{\textbf{68.4}} & \resacell{\underline{89.5}} & \resacell{\textbf{88.9}} & \resacell{\textbf{66.3}} & \resacell{\textbf{57.3}} & \resacell{\textbf{68.2}} & \resacell{\textbf{69.0}} & \resacell{\textbf{66.2}} & \resacell{\textbf{90.7}} & \multicolumn{1}{@{}>{\columncolor{gainrow}[0pt][0pt]}c@{}}{\makebox[29.53192pt][l]{\rlap{\hspace{-2.35pt}\avgdividerstroke}\makebox[29.53192pt][c]{\resacell{\textbf{71.8}}}}}\\
\hspace{2pt}\gainentry{Gain\,$\uparrow$} &  &  &  &  &  &  &  &  &  &  &  & \gainentry{+4.3}\\
\bottomrule
\end{tabular}%
\end{lrbox}
\typeout{SUPP-COMPARISON-5-WIDTH=\the\wd\suppcomparisonfiveshotbox}
\typeout{SUPP-COMPARISON-TEXT-WIDTH=\the\textwidth}
\begin{minipage}{\textwidth}
\UseCaptionStatus{tab:supp_comparison_5shot}
\caption{\textbf{5-shot comparison across all ten datasets.} Foreground mIoU (\%). Avg weights all ten datasets equally. Conventions follow Table~\ref{tab:supp_comparison_1shot}.}
\label{tab:supp_comparison_5shot}
\centering
\usebox{\suppcomparisonfiveshotbox}
\end{minipage}
\end{table*}
\subsection{Methods using mask-pretrained models}

\paragraph{GF-SAM, Matcher, and PerSAM.}
GF-SAM v2 and v3 use DINOv2-L/14 and DINOv3-L/16 features, respectively, with the same SAM-H mask model~\cite{gfsam,insid3}.
The v2 row retains the published GF-SAM scores and the additional benchmark results reported by INSID3.
The v3 row uses the DINOv3 variant reported by INSID3, with our local FSS and PASCAL-$5^i$ evaluations marked by cell-level daggers.
For PASCAL-$5^i$, Matcher uses the result reported by GF-SAM~\cite{matcher,gfsam}.
PerSAM combines DINOv2-L localization with SAM-H prompting in both shot settings, without PerSAM-F optimization~\cite{persam}.
Both shot settings are locally evaluated on all ten datasets.
The 5-shot evaluation follows the paper's maximum-location rule across five references.

\paragraph{RPG-SAM.}
RPG-SAM is locally evaluated on all ten datasets with the shared episode lists and scoring~\cite{rpgsam}.
RPG-SAM uses DINOv2-L/14 and the original SAM2 Hiera-L checkpoint, retaining the authors' prototype mining, geometric threshold selection, and iterative prompting.
Full reference/query images are resized to $560\times560$; masks use nearest-neighbor resizing.
When SLIC~\cite{slic} foreground regions disappear on the feature grid, a single coverage-weighted foreground prototype is used.
An originally empty reference mask yields an all-background prediction and remains in the evaluation.
For 5-shot evaluation, we average the five independently computed reference priors before threshold selection and SAM2 refinement.

\paragraph{RNS.}
All twenty RNS entries are local evaluations with DINOv3 ViT-L/16 and SAM2.1-L across the ten datasets~\cite{rns}.
We disable the text branch and use all annotated references to build foreground and background feature banks, retaining the native region-level sliding-window inference.

\paragraph{PR-MaGIC and scoring convention.}
All twenty PR-MaGIC entries are local evaluations of the authors' Matcher-based variant with DINOv2-L/14 and SAM ViT-H~\cite{prmagic}.
We retain native multi-reference matching and select one of six candidate masks by support--query cosine similarity.
DINOv2 uses $518\times518$ inputs and SAM retains its native 1024-pixel preprocessing; final scoring follows the shared evaluation grids above.
The authors' implementation averages foreground and background IoU within each episode,
\begin{equation}
s_e^{\mathrm{PR}}=\tfrac12\left(
\frac{I_{e,\mathrm{fg}}}{U_{e,\mathrm{fg}}}+
\frac{I_{e,\mathrm{bg}}}{U_{e,\mathrm{bg}}}\right),
\end{equation}
then averages episodes and folds.
Our tables instead accumulate foreground intersections and unions by class before averaging classes and folds.
The reported ten-dataset averages, 48.3/49.9, use our foreground metric.
The original PR-MaGIC paper's ``Pascal'' benchmark denotes PASCAL-Part, distinct from the PASCAL-$5^i$ column added here.

\paragraph{GSM+Matcher.}
GSM+Matcher uses the published results of GSM applied to Matcher~\cite{gsm}.
It reports 1-shot/5-shot scores on LVIS-$92^i$, COCO-$20^i$, FSS, PASCAL-$5^i$, ISIC, and iSAID.
PASCAL-Part, PACO-Part, SUIM, and X-Ray are unreported; their entries and the incomplete group and ten-dataset averages are shown as --.


\begin{table}[!htbp]
\centering
\UseCaptionStatus{tab:supp_group_means}
\caption{\textbf{Group-wise comparison.} Equal-weight mean foreground IoU (\%) for objects (4 datasets), parts (2), and cross-domain benchmarks (4). --: incomplete groups. Gray: averages with task-training overlap, excluded from ranking. Conventions follow Table~\ref{tab:supp_comparison_1shot}.}
\label{tab:supp_group_means}
\fontsize{8.5}{10}\selectfont
\setlength{\tabcolsep}{1.65pt}
\renewcommand{\arraystretch}{1.00}
\newlength{\suppgroupmethodwidth}
\settowidth{\suppgroupmethodwidth}{\hspace{2pt}GSM+Matcher\cite{gsm}}
\newsavebox{\suppgroupmethodbox}
\sbox{\suppgroupmethodbox}{\hspace{2pt}GSM+Matcher\cite{gsm}}
\ifdim\wd\suppgroupmethodbox>\suppgroupmethodwidth\setlength{\suppgroupmethodwidth}{\wd\suppgroupmethodbox}\fi
\newlength{\suppgroupnumwidth}
\setlength{\suppgroupnumwidth}{\dimexpr(\columnwidth-\suppgroupmethodwidth-12\tabcolsep)/6\relax}
\begin{tabular}{@{}>{\raggedright\arraybackslash}p{\suppgroupmethodwidth}*{6}{>{\centering\arraybackslash}p{\suppgroupnumwidth}}@{}}
\toprule
\multirow{2}{*}[-2.7pt]{\textbf{Method}} & \multicolumn{2}{c}{\textbf{Objects}} & \multicolumn{2}{c}{\textbf{Parts}} & \multicolumn{2}{c}{\textbf{Cross-domain}} \\
\cmidrule(lr){2-3}\cmidrule(lr){4-5}\cmidrule(lr){6-7}
 & 1-shot & 5-shot & 1-shot & 5-shot & 1-shot & 5-shot \\
\midrule
\multicolumn{7}{@{}l}{\textbf{Task-specific fine-tuning:} \textit{Semantic + mask supervision}} \\
\hspace{2pt}SegGPT\cite{seggpt} & \textcolor{trainoverlap}{60.9} & \textcolor{trainoverlap}{68.1} & \textcolor{trainoverlap}{24.7} & \textcolor{trainoverlap}{28.5} & \textcolor{trainoverlap}{47.7} & \textcolor{trainoverlap}{51.0}\\
\hspace{2pt}SegIC\cite{segic} & \textcolor{trainoverlap}{73.3} & \textcolor{trainoverlap}{74.2} & 32.9 & 34.7 & 39.6 & 43.5\\
\hspace{2pt}SINE\cite{sine} & \textcolor{trainoverlap}{66.3} & \textcolor{trainoverlap}{68.2} & 29.8 & 30.9 & 38.7 & 41.1\\
\hspace{2pt}DiffewS\cite{diffews} & \textcolor{trainoverlap}{69.7} & \textcolor{trainoverlap}{70.9} & 28.4 & 32.9 & 41.5 & 43.9\\
\hspace{2pt}LDIS-1\cite{ldis} & \textcolor{trainoverlap}{56.4} & \textcolor{trainoverlap}{59.9} & 25.6 & 26.8 & 26.2 & 26.9\\
\hspace{2pt}FSSAM\cite{fssam} & \textcolor{trainoverlap}{66.9} & \textcolor{trainoverlap}{73.7} & 38.4 & 45.5 & 60.3 & 62.6\\
\hspace{2pt}COSINE\cite{cosine} & \textcolor{trainoverlap}{69.1} & \textcolor{trainoverlap}{71.9} & \textcolor{trainoverlap}{31.9} & \textcolor{trainoverlap}{32.5} & 38.4 & 39.4\\
\hspace{2pt}SANSA\cite{sansa} & -- & -- & 34.8 & -- & -- & --\\
\midrule
\multicolumn{7}{@{}l}{\textbf{No offline task training:} \textit{Mask-supervised pre-training}} \\
\hspace{2pt}PerSAM\cite{persam} & 44.3 & 48.0 & 28.3 & 30.5 & 30.7 & 32.5\\
\hspace{2pt}Matcher\cite{matcher} & 60.2 & 66.1 & 38.8 & 39.7 & 46.7 & 47.8\\
\hspace{2pt}GF-SAM v2\cite{gfsam} & \underline{63.5} & \underline{70.6} & 40.4 & 46.6 & 50.0 & 54.7\\
\hspace{2pt}GF-SAM v3\cite{gfsam} & 61.7 & 69.6 & 39.7 & 44.8 & 51.1 & 55.8\\
\hspace{2pt}GSM+Matcher\cite{gsm} & 62.2 & 67.5 & -- & -- & -- & --\\
\hspace{2pt}RNS\cite{rns} & 54.1 & 65.5 & 25.1 & 32.8 & 40.5 & 46.6\\
\hspace{2pt}PR-MaGIC\cite{prmagic} & 54.7 & 57.8 & 36.4 & 36.9 & 47.8 & 48.5\\
\hspace{2pt}RPG-SAM\cite{rpgsam} & 51.8 & 58.7 & 39.2 & 45.4 & 47.5 & 51.2\\
\midrule
\multicolumn{7}{@{}l}{\textbf{No offline task training:} \textit{Self-supervised pre-training}} \\
\hspace{2pt}INSID3\cite{insid3} & 63.1 & 70.0 & 44.6 & 52.0 & 60.1 & 65.7\\
\hspace{2pt}HERA\cite{hera} & 53.3 & 61.1 & 39.1 & 52.4 & 55.0 & 64.9\\
\hspace{2pt}RSRM\cite{rsrm} & 58.1 & 66.0 & 44.5 & 54.3 & \underline{61.5} & 68.0\\
\hspace{2pt}FROST\cite{frost} & 58.5 & 69.0 & \underline{46.1} & \underline{58.5} & 61.4 & \underline{70.4}\\
\rowcolor{gainrow}
\multicolumn{1}{@{}>{\columncolor{gainrow}[0pt][\tabcolsep]}l}{\resacell{\hspace{2pt}ReSA (ours)}} & \resacell{\textbf{66.0}} & \resacell{\textbf{74.9}} & \resacell{\textbf{50.7}} & \resacell{\textbf{61.8}} & \resacell{\textbf{65.2}} & \multicolumn{1}{>{\columncolor{gainrow}[\tabcolsep][0pt]}c@{}}{\resacell{\textbf{73.5}}}\\
\hspace{2pt}\gainentry{Gain\,$\uparrow$} & \gainentry{+2.5} & \gainentry{+4.3} & \gainentry{+4.6} & \gainentry{+3.3} & \gainentry{+3.7} & \gainentry{+3.1}\\
\bottomrule
\end{tabular}
\end{table}
\subsection{Methods using self-supervised features}

\paragraph{INSID3 and RSRM.}
INSID3's eight original benchmark scores come from its paper~\cite{insid3}.
Its FSS and PASCAL-$5^i$ entries are official-code evaluations using the same frozen lists as \method.
All RSRM entries are local official-code evaluations with DINOv3-L/16, whereas RSRM's paper reports B/16~\cite{rsrm}.

\paragraph{FROST.}
The FROST row uses frozen DINOv3-L/16 features without an auxiliary segmentation model~\cite{frost}.
Both shot settings on FSS-1000, PASCAL-Part, iSAID, and X-Ray are local evaluations and carry cell-level daggers; the other six datasets use published scores.
Local runs use $1024\times1024$ inputs and the scoring grids in Section~C.1; 5-shot uses the author's sequential reference interface.

\paragraph{HERA and episode isolation.}
All twenty HERA entries are local evaluations with DINOv3-L/16 and native $400\times400$ inputs~\cite{hera}.
We retain feature fusion, calibrated attention refinement, and support-only test-time optimization with Adam~\cite{adam} at learning rate $0.0013$, using two updates for 1-shot and four for 5-shot.
HERA optimizes selected backbone parameters and its fusion head at test time, with no offline task training.
For independent episodes, we reset both the backbone and fusion head to their initial states; the released code resets only the backbone and retains fusion-head state across episodes.
Our row uses this episode-isolated evaluation with the sampling and foreground-scoring protocol above.
PASCAL-$5^i$ uses the common four-fold episode lists and original-resolution scoring.

\subsection{Detailed benchmark comparisons}

ReSA achieves the highest ten-dataset mean among eligible comparisons in both shot settings (Tables~\ref{tab:supp_comparison_1shot} and~\ref{tab:supp_comparison_5shot}) and leads the part and cross-domain groups (Table~\ref{tab:supp_group_means}). Its part scores reach 50.7/61.8 mIoU and its cross-domain scores reach 65.2/73.5. Tables~\ref{tab:supp_comparison_1shot} and~\ref{tab:supp_comparison_5shot} additionally include SANSA and GF-SAM v3; the main-paper GF-SAM row is labeled v2 here.

\begin{figure*}[!tp]
\centering
\includegraphics[width=\textwidth]{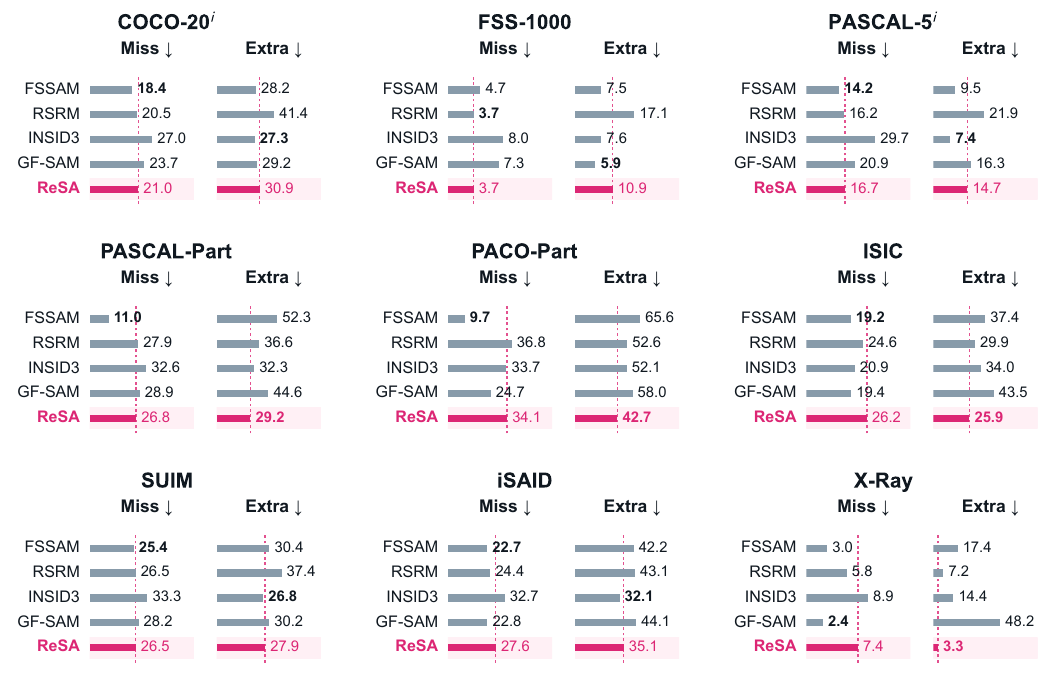}
\UseCaptionStatus{fig:supp_foreground_errors}
\caption{\textbf{1-shot foreground error rates beyond LVIS-$92^i$.}
Miss is $\mathrm{FN}/(\mathrm{TP}+\mathrm{FN})$; Extra is $\mathrm{FP}/(\mathrm{TP}+\mathrm{FP})$.
Counts are pooled within each class; classes and folds are averaged equally.
Bars report percentages (lower is better), with ReSA in pink; dashed lines mark ReSA's values.
Bold values indicate the lowest error.
GF-SAM uses our local DINOv2-L/14 + SAM-H reproduction.}
\label{fig:supp_foreground_errors}
\end{figure*}

\paragraph{Foreground selection.}
ReSA has the lowest extra-foreground rates among the methods displayed in Fig.~\ref{fig:supp_foreground_errors} on both PASCAL-Part (29.2\%) and PACO-Part (42.7\%), indicating more precise selection of the requested part. The figure extends the main-paper LVIS-$92^i$ analysis to the other nine benchmarks.

On COCO-$20^i$, ReSA reduces the miss rate from INSID3's 27.0\% to 21.0\%; the corresponding extra-foreground rates are 30.9\% and 27.3\%.

\begin{table*}[tbp]
\begingroup
\centering
\UseCaptionStatus{tab:supp_progression}
\caption{\textbf{Module combinations.} Foreground mIoU (\%) on 133,962 matched episodes. Base denotes the first-stage initialization $p_0$. Bold marks the highest displayed scores per dataset and shot, including ties.}
\label{tab:supp_progression}
\fontsize{8.5}{10}\selectfont
\setlength{\tabcolsep}{2pt}
\renewcommand{\arraystretch}{1.15}
\begin{tabular*}{\textwidth}{@{\extracolsep{\fill}}lcccccccccc@{}}
\toprule
\multirow{2}{*}[-2.7pt]{\textbf{Dataset}} & \multicolumn{5}{c}{\textbf{1-shot}} & \multicolumn{5}{c}{\textbf{5-shot}}\\
\cmidrule(lr){2-6}\cmidrule(lr){7-11}
 & \textbf{Base} & \textbf{RCA} & \textbf{RCA+RMS} & \textbf{RCA+TAR} & \textbf{Full} & \textbf{Base} & \textbf{RCA} & \textbf{RCA+RMS} & \textbf{RCA+TAR} & \textbf{Full}\\
\midrule
LVIS-$92^i$ & 37.5 & 44.1 & 43.2 & \textbf{45.2} & 44.9 & 47.0 & 52.4 & 50.6 & \textbf{54.8} & 53.0\\
COCO-$20^i$ & 50.2 & 55.3 & 58.0 & 55.8 & \textbf{58.8} & 63.3 & 66.8 & 67.2 & 67.9 & \textbf{68.4}\\
FSS-1000 & 81.3 & 84.6 & 84.6 & \textbf{86.4} & \textbf{86.4} & 86.2 & 87.9 & 87.6 & \textbf{89.7} & 89.5\\
PASCAL-$5^i$ & 66.6 & 69.8 & 73.4 & 69.9 & \textbf{73.8} & 85.2 & 87.6 & 88.2 & 88.1 & \textbf{88.9}\\
PASCAL-Part & 51.3 & 53.4 & 55.5 & 53.7 & \textbf{56.0} & 63.0 & 65.4 & 65.7 & 66.0 & \textbf{66.3}\\
PACO-Part & 39.5 & 41.0 & 44.9 & 41.3 & \textbf{45.5} & 52.8 & 55.2 & 56.6 & 55.9 & \textbf{57.3}\\
ISIC & 54.0 & 55.5 & 58.7 & 55.6 & \textbf{58.8} & 63.7 & 65.4 & 67.8 & 65.9 & \textbf{68.2}\\
SUIM & 54.8 & 56.5 & 58.0 & 57.0 & \textbf{58.8} & 65.4 & 67.1 & 67.7 & 68.2 & \textbf{69.0}\\
iSAID & 49.5 & 51.0 & 53.0 & 51.0 & \textbf{53.2} & 63.0 & 65.2 & 65.6 & 65.6 & \textbf{66.2}\\
X-Ray & 88.3 & 89.3 & 89.3 & \textbf{89.9} & 89.7 & 89.3 & 89.8 & 90.2 & 90.4 & \textbf{90.7}\\
\midrule
Mean & 57.3 & 60.0 & 61.9 & 60.6 & \textbf{62.6} & 67.9 & 70.3 & 70.7 & 71.3 & \textbf{71.8}\\
\bottomrule
\end{tabular*}

\endgroup
\end{table*}
\section{Additional ablation results}
\label{sec:supp_ablations}

Accuracy uses one decimal place unless otherwise specified; differences subtract displayed scores. Timing uses seconds to three decimal places.

\begin{table}[!htbp]
\begingroup
\centering
\UseCaptionStatus{tab:supp_ca_readout}
\caption{\textbf{Query adaptation under matched threshold fitting and refinement.} Ten-dataset mean mIoU (\%) on 133,962 matched episodes. All rows use the same reference-weighted KL threshold fit. For RCA, $\lambda$ weights query entropy; RMS selects between the two branches. Bold marks the best per shot.}
\label{tab:supp_ca_readout}
\fontsize{8.5}{10}\selectfont
\setlength{\tabcolsep}{2.5pt}
\renewcommand{\arraystretch}{1.13}
\begin{tabular*}{\columnwidth}{@{\extracolsep{\fill}}lcccc@{}}
\toprule
Prediction & Branch setting & TAR & 1-shot & 5-shot\\
\midrule
$p_0$ & Single prediction & -- & 57.7 & 67.1\\
$R$-only & Single prediction & -- & 57.8 & 68.1\\
RCA & Fixed $\lambda=1$ & -- & 61.5 & 70.2\\
RCA & RMS: $\lambda\in\{1,3\}$ & -- & 61.9 & 70.7\\
\midrule
$p_0$ & Single prediction & $\checkmark$ & 58.8 & 68.4\\
$R$-only & Single prediction & $\checkmark$ & 58.8 & 69.4\\
RCA & RMS: $\lambda\in\{1,3\}$ & $\checkmark$ & \textbf{62.6} & \textbf{71.8}\\
\bottomrule
\end{tabular*}

\endgroup
\end{table}
\subsection{Module configurations}

RMS improves the RCA-only mean by 1.9/0.4 mIoU points in 1-shot/5-shot, and TAR adds a further 0.7/1.1 points to RCA+RMS (Table~\ref{tab:supp_progression}).
The table follows the five module combinations in main-paper Table~3.
Base thresholds $p_0$ at $T=0.5$ immediately after reference fitting.
RCA adapts with the conditional objective, retaining the fixed $\lambda=1$ branch and $T=0.5$.
RCA+RMS fits each branch's threshold and then selects the branch with the larger relative gain; RCA+TAR refines the same RCA prediction at $T=0.5$.
Full adds TAR to RCA+RMS. All five configurations use 66,981 episodes per shot and the class/fold aggregation specified above.

Threshold controls vary the cutoff on the same RMS-selected probability map $p^*$; refinement controls also fix its coarse mask and fitted threshold. Query-objective and branch-selection controls use matched inputs and optimization settings.
Without TAR, token masks are bilinearly resized and binarized under the same evaluation protocol.

\paragraph{Adaptation improves the input to mask selection.}
Under the same single-candidate threshold fit, RCA exceeds continued reference supervision by 3.7/2.1 mIoU points without TAR (Table~\ref{tab:supp_ca_readout}). The $R$-only control resets AdamW after reference fitting and continues reference supervision for the same adaptation budget. Both controls share the features, kernel, and episodes. Here the $\lambda=1$ prediction uses a fitted threshold; the RCA-only configuration in Table~\ref{tab:supp_progression} uses $T=0.5$.

\begin{table}[tbp]
\begingroup
\centering
\UseCaptionStatus{tab:supp_ja}
\caption{\textbf{Single, pairwise, and full RCA objectives.} Ten-dataset mean mIoU (\%) on 133,962 matched episodes. $R$, $C$, and $H$ denote reference supervision, conditional KL, and query entropy. The upper block uses $\lambda=1$ and $T=0.5$. The lower block includes RMS and TAR and reports two decimals. Bold marks the best per shot within each block, including ties.}
\label{tab:supp_ja}
\fontsize{8.5}{10}\selectfont
\setlength{\tabcolsep}{3pt}
\renewcommand{\arraystretch}{1.15}
\begin{tabular*}{\columnwidth}{@{\extracolsep{\fill}\hspace{5pt}}lcc@{\hspace{5pt}}}
\toprule
\textbf{Objective} & \textbf{1-shot} & \textbf{5-shot}\\
\midrule
$R$ & 56.8 & 68.3\\
$C$ & 57.3 & 67.9\\
$H$ & 48.9 & 64.0\\
$R+C$ & 55.2 & 67.5\\
$R+H$ & 53.9 & 69.7\\
$C+H$ & \textbf{60.0} & \textbf{70.4}\\
$R+C+H$ & \textbf{60.0} & 70.3\\
\midrule
\multicolumn{3}{@{\hspace{5pt}}l@{\hspace{5pt}}}{\textit{Full pipeline (RMS + TAR)}}\\
$C+H$ & 62.47 & \textbf{71.75}\\
$R+C+H$ & \textbf{62.60} & \textbf{71.75}\\
\bottomrule
\end{tabular*}
\endgroup
\end{table}

\subsection{RCA objectives and error patterns}
\label{sec:supp_ca_controls}

\paragraph{Complementary query objectives.}
With reference supervision $R$ retained, combining conditional KL $C$ and query entropy $H$ raises mean mIoU from 56.8/68.3 to 60.0/70.3 in 1-shot/5-shot (Table~\ref{tab:supp_ja}). Adding either query term alone gives 55.2/67.5 for $R+C$ and 53.9/69.7 for $R+H$, highlighting the benefit of combining the two terms. The $C+H$ control reaches 60.0/70.4 under the fixed cutoff. With RMS and TAR, $C+H$ and $R+C+H$ reach 62.47/71.75 and 62.60/71.75, respectively.

Here $R=\mathcal L_{\rm ref}$, $C=\mathcal L_{\rm cond}$, and $H=\overline H$. All variants start from the same reference-fitted $p_0$. Fixed-threshold controls share the adaptation budget, use $\lambda=1$ and $T=0.5$, and omit RMS and TAR; retained query terms keep their $1/S$ scaling. Both full-pipeline variants use the same two-branch configuration and the same RMS and TAR procedures.

\paragraph{Complementary reductions in foreground and background errors.}
Adding $C$ to $R+H$ reduces foreground misses from 37.1/17.9\% to 26.5/14.9\%, while adding $H$ to $R+C$ reduces background false-positive rates from 7.8/5.4\% to 5.7/4.4\% in 1-shot/5-shot. Relative to $R$ alone, the full objective also reduces foreground-area error from 7.6/4.9 to 7.1/4.5 percentage points. Main-paper Fig.~6(a) and Fig.~\ref{fig:supp_ja_examples} visualize the corresponding probability changes.

These diagnostics use the matched loss controls in Table~\ref{tab:supp_ja}. Foreground miss rate is $\mathrm{FN}/(\mathrm{TP}+\mathrm{FN})$, and background false-positive rate is $\mathrm{FP}/(\mathrm{FP}+\mathrm{TN})$. Counts are pooled within each class before equal class, fold, and dataset averaging. Foreground-area error is the absolute difference between predicted and annotated foreground fractions, computed per episode and averaged over classes, folds, and datasets.

\begin{table}[!htbp]
\centering
\UseCaptionStatus{tab:supp_initial_confidence}
\caption{\textbf{Foreground recall by initial confidence.} Recall (\%) in fixed $p_0$ intervals after the same TAR refinement for each objective; cells list ten-dataset means for 1-shot/5-shot. All variants use $\lambda=1$ and $T=0.5$. Objectives follow Table~\ref{tab:supp_ja}.}
\label{tab:supp_initial_confidence}
\fontsize{8.5}{10}\selectfont
\setlength{\tabcolsep}{3pt}
\renewcommand{\arraystretch}{1.12}
\begin{tabular*}{\columnwidth}{@{\extracolsep{\fill}}lccc@{}}
\toprule
\textbf{Initial $p_0$} & $R$ & $R+H$ & $R+C+H$\\
\midrule
$[0,0.3]$ & 0.1/0.3 & 3.2/3.2 & 1.6/1.2\\
$(0.3,0.5]$ & 10.1/13.5 & 17.2/23.0 & 20.1/20.7\\
$(0.5,0.7]$ & 74.4/76.0 & 44.4/60.5 & 69.6/73.3\\
$(0.7,1]$ & 99.7/99.8 & 85.1/96.8 & 98.5/99.5\\
\bottomrule
\end{tabular*}
\end{table}
\paragraph{Foreground retention by initial confidence.}
Adding $C$ to $R+H$ gives the largest recall gains in bins with $p_0>0.5$ (Table~\ref{tab:supp_initial_confidence}). Recall rises from 44.4/60.5 to 69.6/73.3 in $(0.5,0.7]$ and from 85.1/96.8 to 98.5/99.5 in $(0.7,1]$, showing improved retention of target pixels already supported by reference fitting.

We group query pixels by their reference-fitted foreground probability $p_0$, using fixed boundaries at $0.3$, $0.5$, and $0.7$ before query adaptation.
The three objectives share initialization, budgets, $\lambda=1$, $T=0.5$, and the same TAR refinement.
For grouping, continuous $p_0$ is bilinearly resized to the scoring grid; final predictions follow the dataset's scoring convention.
The analysis covers 133,690 episodes with probability outputs.
Within each bin, recall pools foreground counts per class and then averages defined class rates, folds, and datasets. All ten datasets contribute to every bin; class--fold groups without foreground in a bin are omitted from that bin's average.

\setcounter{dbltopnumber}{2}
\subsection{Mask selection controls}
\label{sec:supp_jms_controls}
\begin{table}[!t]
\centering
\UseCaptionStatus{tab:reference_controls}
\caption{\textbf{Reference weights in RMS.} Ten-dataset mean mIoU (\%), without TAR. Both rows use prototype-response KL, fitted thresholds, and relative-gain selection.}
\label{tab:reference_controls}
\fontsize{8.5}{10}\selectfont
\setlength{\tabcolsep}{3pt}
\renewcommand{\arraystretch}{1.10}
\begin{tabular*}{0.78\columnwidth}{@{\extracolsep{\fill}}lcc@{}}
\toprule
RMS weights & 1-shot & 5-shot\\
\midrule
Uniform & 61.6 & 69.6\\
Reference & \textbf{61.9} & \textbf{70.7}\\
\bottomrule
\end{tabular*}
\end{table}

\paragraph{Reconstruction weights.}
Reference weighting improves mask selection with the adapted candidate maps held fixed: mean mIoU rises from 61.6/69.6 to 61.9/70.7 (Table~\ref{tab:reference_controls}). The control replaces uniform weights, $w_i=1/N$, with reference-kernel weights, $w_i=(2S)^{-1}\sum_rK_{ri}$, throughout threshold fitting and relative-gain scoring.

\begin{table*}[!t]
\begingroup
\centering
\UseCaptionStatus{tab:supp_thresholds}
\caption{\textbf{Thresholds on fixed RMS-selected probabilities.} Foreground mIoU (\%) on 133,962 matched episodes, without TAR. Parentheses report differences from the RMS threshold using displayed one-decimal scores; bold marks the best per dataset and shot, including ties.}
\label{tab:supp_thresholds}
\fontsize{8.5}{10}\selectfont
\setlength{\tabcolsep}{2pt}
\renewcommand{\arraystretch}{1.15}
\begin{tabular*}{\textwidth}{@{\extracolsep{\fill}}lcccccccc@{}}
\toprule
\multirow{2}{*}[-2.7pt]{\textbf{Dataset}} & \multicolumn{4}{c}{\textbf{1-shot}} & \multicolumn{4}{c}{\textbf{5-shot}}\\
\cmidrule(lr){2-5}\cmidrule(lr){6-9}
 & \textbf{\boldmath $T=0.3$} & \textbf{\boldmath $T=0.5$} & \textbf{\boldmath $T=0.7$} & \textbf{RMS threshold} & \textbf{\boldmath $T=0.3$} & \textbf{\boldmath $T=0.5$} & \textbf{\boldmath $T=0.7$} & \textbf{RMS threshold}\\
\midrule
LVIS-$92^i$ & 44.5\suppdelta{+1.3} & \textbf{44.6}\suppdelta{+1.4} & 43.1\suppdelta{-0.1} & 43.2 & 51.4\suppdelta{+0.8} & 53.4\suppdelta{+2.8} & \textbf{54.4}\suppdelta{+3.8} & 50.6\\
COCO-$20^i$ & 56.6\suppdelta{-1.4} & 55.8\suppdelta{-2.2} & 53.9\suppdelta{-4.1} & \textbf{58.0} & 66.5\suppdelta{-0.7} & 67.3\suppdelta{+0.1} & \textbf{67.4}\suppdelta{+0.2} & 67.2\\
FSS-1000 & 83.9\suppdelta{-0.7} & 84.7\suppdelta{+0.1} & \textbf{85.0}\suppdelta{+0.4} & 84.6 & 87.0\suppdelta{-0.6} & 87.8\suppdelta{+0.2} & \textbf{88.3}\suppdelta{+0.7} & 87.6\\
PASCAL-$5^i$ & 71.8\suppdelta{-1.6} & 70.3\suppdelta{-3.1} & 67.7\suppdelta{-5.7} & \textbf{73.4} & 88.1\suppdelta{-0.1} & 88.0\suppdelta{-0.2} & 87.3\suppdelta{-0.9} & \textbf{88.2}\\
PASCAL-Part & 54.4\suppdelta{-1.1} & 53.4\suppdelta{-2.1} & 51.0\suppdelta{-4.5} & \textbf{55.5} & 64.8\suppdelta{-0.9} & 65.6\suppdelta{-0.1} & 65.3\suppdelta{-0.4} & \textbf{65.7}\\
PACO-Part & 42.6\suppdelta{-2.3} & 40.9\suppdelta{-4.0} & 37.6\suppdelta{-7.3} & \textbf{44.9} & 55.1\suppdelta{-1.5} & 55.2\suppdelta{-1.4} & 53.7\suppdelta{-2.9} & \textbf{56.6}\\
ISIC & 57.9\suppdelta{-0.8} & 56.7\suppdelta{-2.0} & 54.7\suppdelta{-4.0} & \textbf{58.7} & 67.4\suppdelta{-0.4} & 66.8\suppdelta{-1.0} & 65.1\suppdelta{-2.7} & \textbf{67.8}\\
SUIM & 57.3\suppdelta{-0.7} & 56.6\suppdelta{-1.4} & 55.3\suppdelta{-2.7} & \textbf{58.0} & 67.1\suppdelta{-0.6} & 67.3\suppdelta{-0.4} & 66.8\suppdelta{-0.9} & \textbf{67.7}\\
iSAID & 51.8\suppdelta{-1.2} & 51.1\suppdelta{-1.9} & 49.8\suppdelta{-3.2} & \textbf{53.0} & 64.9\suppdelta{-0.7} & 65.3\suppdelta{-0.3} & 65.2\suppdelta{-0.4} & \textbf{65.6}\\
X-Ray & 89.3\suppdelta{+0.0} & \textbf{89.4}\suppdelta{+0.1} & 87.8\suppdelta{-1.5} & 89.3 & 88.4\suppdelta{-1.8} & 90.0\suppdelta{-0.2} & 89.7\suppdelta{-0.5} & \textbf{90.2}\\
\midrule
Mean & 61.0\suppdelta{-0.9} & 60.3\suppdelta{-1.6} & 58.6\suppdelta{-3.3} & \textbf{61.9} & 70.1\suppdelta{-0.6} & \textbf{70.7}\suppdelta{+0.0} & 70.3\suppdelta{-0.4} & \textbf{70.7}\\
\bottomrule
\end{tabular*}

\endgroup
\end{table*}

\begin{table*}[!t]
\begingroup
\centering
\UseCaptionStatus{tab:supp_amg}
\caption{\textbf{Branch selection in RMS.} Foreground mIoU (\%) on 133,962 matched episodes with fixed probabilities and threshold fits, without TAR. Parentheses subtract Relative gain using displayed scores; bold marks the best per setting, including ties.}
\label{tab:supp_amg}
\fontsize{8.5}{10}\selectfont
\setlength{\tabcolsep}{2pt}
\renewcommand{\arraystretch}{1.15}
\begin{tabular*}{\textwidth}{@{\extracolsep{\fill}}lcccccccc@{}}
\toprule
\multirow{2}{*}[-2.7pt]{\textbf{Dataset}} & \multicolumn{4}{c}{\textbf{1-shot}} & \multicolumn{4}{c}{\textbf{5-shot}}\\
\cmidrule(lr){2-5}\cmidrule(lr){6-9}
 & \textbf{\boldmath $\lambda=1$} & \textbf{\boldmath $\lambda=3$} & \textbf{\boldmath Min $D$} & \textbf{Relative gain} & \textbf{\boldmath $\lambda=1$} & \textbf{\boldmath $\lambda=3$} & \textbf{\boldmath Min $D$} & \textbf{Relative gain}\\
\midrule
LVIS-$92^i$ & 42.5\suppdelta{-0.7} & 42.8\suppdelta{-0.4} & 42.5\suppdelta{-0.7} & \textbf{43.2} & 49.4\suppdelta{-1.2} & \textbf{51.1}\suppdelta{+0.5} & 50.7\suppdelta{+0.1} & 50.6\\
COCO-$20^i$ & 57.5\suppdelta{-0.5} & 56.8\suppdelta{-1.2} & 56.7\suppdelta{-1.3} & \textbf{58.0} & 66.7\suppdelta{-0.5} & \textbf{67.3}\suppdelta{+0.1} & 67.1\suppdelta{-0.1} & 67.2\\
FSS-1000 & 84.5\suppdelta{-0.1} & 84.5\suppdelta{-0.1} & 84.4\suppdelta{-0.2} & \textbf{84.6} & \textbf{87.6}\suppdelta{+0.0} & \textbf{87.6}\suppdelta{+0.0} & \textbf{87.6}\suppdelta{+0.0} & \textbf{87.6}\\
PASCAL-$5^i$ & 73.1\suppdelta{-0.3} & 72.2\suppdelta{-1.2} & 72.5\suppdelta{-0.9} & \textbf{73.4} & 87.8\suppdelta{-0.4} & \textbf{88.3}\suppdelta{+0.1} & \textbf{88.3}\suppdelta{+0.1} & 88.2\\
PASCAL-Part & 55.4\suppdelta{-0.1} & 54.1\suppdelta{-1.4} & 54.2\suppdelta{-1.3} & \textbf{55.5} & 65.4\suppdelta{-0.3} & 65.5\suppdelta{-0.2} & 65.4\suppdelta{-0.3} & \textbf{65.7}\\
PACO-Part & 44.7\suppdelta{-0.2} & 43.1\suppdelta{-1.8} & 43.0\suppdelta{-1.9} & \textbf{44.9} & 56.2\suppdelta{-0.4} & 55.9\suppdelta{-0.7} & 55.8\suppdelta{-0.8} & \textbf{56.6}\\
ISIC & 57.6\suppdelta{-1.1} & 57.6\suppdelta{-1.1} & 57.6\suppdelta{-1.1} & \textbf{58.7} & 66.5\suppdelta{-1.3} & 67.5\suppdelta{-0.3} & 67.3\suppdelta{-0.5} & \textbf{67.8}\\
SUIM & 57.7\suppdelta{-0.3} & 57.5\suppdelta{-0.5} & 57.5\suppdelta{-0.5} & \textbf{58.0} & 67.3\suppdelta{-0.4} & 67.6\suppdelta{-0.1} & 67.5\suppdelta{-0.2} & \textbf{67.7}\\
iSAID & \textbf{53.0}\suppdelta{+0.0} & 52.1\suppdelta{-0.9} & 52.3\suppdelta{-0.7} & \textbf{53.0} & 65.3\suppdelta{-0.3} & 65.4\suppdelta{-0.2} & 65.5\suppdelta{-0.1} & \textbf{65.6}\\
X-Ray & \textbf{89.3}\suppdelta{+0.0} & 89.2\suppdelta{-0.1} & \textbf{89.3}\suppdelta{+0.0} & \textbf{89.3} & 90.1\suppdelta{-0.1} & \textbf{90.2}\suppdelta{+0.0} & \textbf{90.2}\suppdelta{+0.0} & \textbf{90.2}\\
\midrule
Mean & 61.5\suppdelta{-0.4} & 61.0\suppdelta{-0.9} & 61.0\suppdelta{-0.9} & \textbf{61.9} & 70.2\suppdelta{-0.5} & 70.6\suppdelta{-0.1} & 70.5\suppdelta{-0.2} & \textbf{70.7}\\
\bottomrule
\end{tabular*}

\endgroup
\end{table*}

\paragraph{Threshold fitting on the selected branch.}
Fitting the cutoff improves 1-shot mean mIoU by 1.6 points over $T=0.5$, reaching 61.9; both give 70.7 after rounding in 5-shot (Table~\ref{tab:supp_thresholds}). The comparison holds the RMS-selected probability map fixed and varies only the cutoff, without TAR. All masks use $p_i>T$.

On 5-shot LVIS-$92^i$, the fitted cutoff shifts the coverage--background balance: foreground misses decrease from 19.2\% to 14.0\%, while background false positives increase from 1.9\% to 2.5\%. The resulting mIoU is 50.6, compared with 53.4 at $T=0.5$.

\paragraph{Branch-selection rule.}
Relative gain provides a stronger branch-selection criterion than absolute reconstruction error: it improves mIoU by 0.9/0.2 points with candidates and fitted thresholds held fixed (Table~\ref{tab:supp_amg}). Min $D$ selects $\lambda^*=\arg\min_{\lambda\in\{1,3\}}D_\lambda^*$; RMS maximizes $1-D_\lambda^*/D_{\lambda,\rm const}$. The normalization measures how much each two-level fit improves on its own constant baseline, allowing comparison across maps with different probability variation.

\begin{table}[!htbp]
\centering
\UseCaptionStatus{tab:supp_random_selection}
\caption{\textbf{Branch-selection controls.} Ten-dataset mean mIoU (\%) on 133,962 matched episodes, before TAR. Candidate probabilities and fitted thresholds are fixed; the random controls use $\lambda\in\{1,3\}$. Randomized controls average 20 fixed seeds.}
\label{tab:supp_random_selection}
\fontsize{8.5}{10}\selectfont
\setlength{\tabcolsep}{3pt}
\renewcommand{\arraystretch}{1.15}
\begin{tabular*}{\columnwidth}{@{\extracolsep{\fill}}lcc@{}}
\toprule
\textbf{Selection rule} & \textbf{1-shot} & \textbf{5-shot}\\
\midrule
Single branch: $\lambda=1$ & 61.5 & 70.2\\
Single branch: $\lambda=3$ & 61.0 & 70.6\\
Min $D$: $\lambda\in\{1,3\}$ & 61.0 & 70.5\\
Random $\lambda\in\{1,3\}$ & 61.2 & 70.4\\
50\% correct (GT-assisted) & 61.3 & 70.4\\
RMS: $\lambda\in\{1,3\}$ & 61.9 & 70.7\\
RMS: $\lambda\in\{1,2,3\}$ & 61.8 & 70.7\\
\midrule
GT oracle: $\lambda\in\{1,3\}$ & 63.2 & 71.7\\
GT oracle: $\lambda\in\{1,2,3\}$ & 63.3 & 71.7\\
\bottomrule
\end{tabular*}
\end{table}

\paragraph{Selection within a fixed candidate set.}
RMS reaches 61.9/70.7 mIoU, compared with 61.2/70.4 for random branch selection (Table~\ref{tab:supp_random_selection}). The GT-guided selector reaches 63.2/71.7 by choosing the candidate with the higher IoU for each episode.

Both randomized controls average 20 seeds (20261115--20261134). The GT-assisted ``50\% correct'' control sets branch-choice correctness to 50\% among non-tied episodes. All selections are made at the episode level before class-level pooling.

\subsection{Backbone and debiasing controls}

\begin{table*}[!t]
\begingroup
\centering
\UseCaptionStatus{tab:supp_backbones}
\caption{\textbf{Backbone and debiasing results by dataset.} Foreground mIoU (\%) on 133,962 matched episodes per configuration; cells list 1-shot/5-shot. Base thresholds the first-stage $p_0$ at $0.5$; Full is RCA+RMS+TAR. Each pair shares the encoder, debias setting, episodes, and reference initialization. Bold marks the better score within each pair. Gains subtract displayed means.}
\label{tab:supp_backbones}
\fontsize{8.5}{10}\selectfont
\setlength{\tabcolsep}{2pt}
\renewcommand{\arraystretch}{1.12}
\begin{tabular*}{\textwidth}{@{\extracolsep{\fill}}l*{10}{c}@{}}
\toprule
\multicolumn{1}{@{}l}{\multirow{3}{*}[-2.4pt]{\textbf{Dataset}}} & \multicolumn{2}{c}{\textbf{DINOv2-L/14}} & \multicolumn{2}{c}{\textbf{DINOv3-S/16}} & \multicolumn{2}{c}{\textbf{DINOv3-B/16}} & \multicolumn{2}{c}{\textbf{DINOv3-L/16}} & \multicolumn{2}{c}{\textbf{DINOv3-L/16}}\\
 & \multicolumn{2}{c}{\textbf{Debias $\checkmark$}} & \multicolumn{2}{c}{\textbf{Debias $\checkmark$}} & \multicolumn{2}{c}{\textbf{Debias $\checkmark$}} & \multicolumn{2}{c}{\textbf{Debias --}} & \multicolumn{2}{c}{\textbf{Debias $\checkmark$}}\\
\cmidrule(lr){2-3}\cmidrule(lr){4-5}\cmidrule(lr){6-7}\cmidrule(lr){8-9}\cmidrule(lr){10-11}
 & \textbf{Base} & \textbf{Full} & \textbf{Base} & \textbf{Full} & \textbf{Base} & \textbf{Full} & \textbf{Base} & \textbf{Full} & \textbf{Base} & \textbf{Full}\\
\midrule
LVIS-$92^i$ & 25.9/35.8 & \textbf{36.8}/\textbf{46.0} & 23.5/31.8 & \textbf{26.1}/\textbf{33.5} & 28.8/39.1 & \textbf{35.4}/\textbf{44.7} & 38.1/48.4 & \textbf{43.2}/\textbf{53.2} & 37.5/47.0 & \textbf{44.9}/\textbf{53.0}\\
COCO-$20^i$ & 41.1/54.1 & \textbf{51.5}/\textbf{63.1} & 41.7/55.1 & \textbf{48.6}/\textbf{59.3} & 47.0/60.6 & \textbf{56.8}/\textbf{66.4} & 48.6/63.0 & \textbf{56.9}/\textbf{68.4} & 50.2/63.3 & \textbf{58.8}/\textbf{68.4}\\
FSS-1000 & 76.4/80.5 & \textbf{82.5}/\textbf{85.9} & 80.7/85.6 & \textbf{86.7}/\textbf{89.0} & 80.7/86.1 & \textbf{86.8}/\textbf{89.9} & 82.9/86.8 & \textbf{87.6}/\textbf{90.2} & 81.3/86.2 & \textbf{86.4}/\textbf{89.5}\\
PASCAL-$5^i$ & 60.0/75.6 & \textbf{68.5}/\textbf{83.3} & 64.2/81.2 & \textbf{71.6}/\textbf{84.7} & 68.9/84.5 & \textbf{75.6}/\textbf{88.3} & 60.6/83.2 & \textbf{67.6}/\textbf{86.9} & 66.6/85.2 & \textbf{73.8}/\textbf{88.9}\\
PASCAL-Part & 47.5/57.6 & \textbf{52.9}/\textbf{63.2} & 43.7/55.1 & \textbf{46.9}/\textbf{57.4} & 47.9/59.7 & \textbf{52.3}/\textbf{63.4} & 49.7/63.0 & \textbf{53.9}/\textbf{66.3} & 51.3/63.0 & \textbf{56.0}/\textbf{66.3}\\
PACO-Part & 33.5/46.1 & \textbf{38.5}/\textbf{52.3} & 29.1/41.1 & \textbf{33.9}/\textbf{44.7} & 33.4/47.2 & \textbf{40.2}/\textbf{52.4} & 39.1/53.2 & \textbf{44.6}/\textbf{57.5} & 39.5/52.8 & \textbf{45.5}/\textbf{57.3}\\
ISIC & 47.6/62.3 & \textbf{51.2}/\textbf{68.1} & 48.1/60.7 & \textbf{56.2}/\textbf{67.6} & 50.0/62.6 & \textbf{56.9}/\textbf{68.7} & 49.0/63.1 & \textbf{51.9}/\textbf{66.9} & 54.0/63.7 & \textbf{58.8}/\textbf{68.2}\\
SUIM & 48.9/58.4 & \textbf{54.6}/\textbf{64.3} & 48.1/60.3 & \textbf{52.0}/\textbf{63.7} & 52.4/63.8 & \textbf{57.2}/\textbf{67.3} & 51.0/64.2 & \textbf{54.5}/\textbf{67.7} & 54.8/65.4 & \textbf{58.8}/\textbf{69.0}\\
iSAID & 42.4/54.6 & \textbf{46.3}/\textbf{59.8} & 40.1/53.0 & \textbf{44.0}/\textbf{54.2} & 41.7/56.2 & \textbf{45.9}/\textbf{58.8} & 47.0/62.1 & \textbf{50.6}/\textbf{65.6} & 49.5/63.0 & \textbf{53.2}/\textbf{66.2}\\
X-Ray & 81.7/83.5 & \textbf{87.3}/\textbf{89.1} & 85.4/86.2 & \textbf{88.1}/\textbf{88.3} & 87.4/88.1 & \textbf{89.3}/\textbf{89.6} & 87.8/88.9 & \textbf{89.7}/\textbf{90.5} & 88.3/89.3 & \textbf{89.7}/\textbf{90.7}\\
\midrule
Mean & 50.5/60.8 & \textbf{57.0}/\textbf{67.5} & 50.5/61.0 & \textbf{55.4}/\textbf{64.2} & 53.8/64.8 & \textbf{59.6}/\textbf{68.9} & 55.4/67.6 & \textbf{60.1}/\textbf{71.3} & 57.3/67.9 & \textbf{62.6}/\textbf{71.8}\\
\textbf{Gain} &  & +6.5/+6.7 &  & +4.9/+3.2 &  & +5.8/+4.1 &  & +4.7/+3.7 &  & +5.3/+3.9\\
\bottomrule
\end{tabular*}

\endgroup
\end{table*}

\begin{figure}[!htbp]
\begingroup
\centering
\includegraphics[width=\columnwidth]{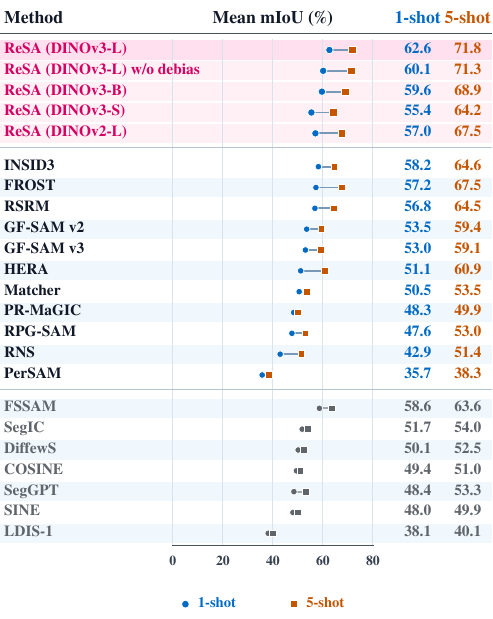}
\UseCaptionStatus{fig:supp_debias_method_means}
\caption{\textbf{Ten-dataset mean comparison.}
ReSA uses DINOv2-L/14 and DINOv3-S/B/L/16; the no-debias control uses DINOv3-L/16. Table~\ref{tab:supp_backbones} gives the matched results.
Methods without a complete ten-dataset mean are omitted.
Gray indicates task-training overlap or possible overlap; these results are excluded from ranking.
Circles/squares denote 1-shot/5-shot.}
\label{fig:supp_debias_method_means}

\endgroup
\end{figure}

\begin{table*}[!t]
\begingroup
\centering
\UseCaptionStatus{tab:supp_tar}
\caption{\textbf{Matched refinement controls.} No TAR evaluates the RCA+RMS coarse mask; the other variants retain matched bands and pairwise terms. Foreground mIoU (\%). Each setting covers 133,962 matched episodes. Parentheses give changes from Full TAR, calculated from the displayed one-decimal scores. Bold marks the highest displayed scores per dataset and shot, including ties.}
\label{tab:supp_tar}
\fontsize{8.5}{10}\selectfont
\setlength{\tabcolsep}{2pt}
\renewcommand{\arraystretch}{1.15}
\begin{tabular*}{\textwidth}{@{\extracolsep{\fill}}lcccccccc@{}}
\toprule
\multirow{2}{*}[-2.7pt]{\textbf{Dataset}} & \multicolumn{4}{c}{\textbf{1-shot}} & \multicolumn{4}{c}{\textbf{5-shot}}\\
\cmidrule(lr){2-5}\cmidrule(lr){6-9}
 & \textbf{No TAR} & \textbf{Unaligned} & \textbf{No evidence} & \textbf{Full TAR} & \textbf{No TAR} & \textbf{Unaligned} & \textbf{No evidence} & \textbf{Full TAR}\\
\midrule
LVIS-$92^i$ & 43.2\suppdelta{-1.7} & 44.8\suppdelta{-0.1} & 44.4\suppdelta{-0.5} & \textbf{44.9} & 50.6\suppdelta{-2.4} & \textbf{53.7}\suppdelta{+0.7} & 52.5\suppdelta{-0.5} & 53.0\\
COCO-$20^i$ & 58.0\suppdelta{-0.8} & 58.3\suppdelta{-0.5} & 58.5\suppdelta{-0.3} & \textbf{58.8} & 67.2\suppdelta{-1.2} & \textbf{68.5}\suppdelta{+0.1} & 68.1\suppdelta{-0.3} & 68.4\\
FSS-1000 & 84.6\suppdelta{-1.8} & 86.5\suppdelta{+0.1} & \textbf{86.6}\suppdelta{+0.2} & 86.4 & 87.6\suppdelta{-1.9} & \textbf{89.7}\suppdelta{+0.2} & \textbf{89.7}\suppdelta{+0.2} & 89.5\\
PASCAL-$5^i$ & 73.4\suppdelta{-0.4} & 72.9\suppdelta{-0.9} & 73.5\suppdelta{-0.3} & \textbf{73.8} & 88.2\suppdelta{-0.7} & 88.7\suppdelta{-0.2} & 88.5\suppdelta{-0.4} & \textbf{88.9}\\
PASCAL-Part & 55.5\suppdelta{-0.5} & 55.6\suppdelta{-0.4} & 55.9\suppdelta{-0.1} & \textbf{56.0} & 65.7\suppdelta{-0.6} & \textbf{66.5}\suppdelta{+0.2} & 66.3\suppdelta{+0.0} & 66.3\\
PACO-Part & 44.9\suppdelta{-0.6} & 44.8\suppdelta{-0.7} & 45.3\suppdelta{-0.2} & \textbf{45.5} & 56.6\suppdelta{-0.7} & 57.1\suppdelta{-0.2} & 57.2\suppdelta{-0.1} & \textbf{57.3}\\
ISIC & 58.7\suppdelta{-0.1} & 58.1\suppdelta{-0.7} & 58.7\suppdelta{-0.1} & \textbf{58.8} & 67.8\suppdelta{-0.4} & 67.9\suppdelta{-0.3} & \textbf{68.2}\suppdelta{+0.0} & \textbf{68.2}\\
SUIM & 58.0\suppdelta{-0.8} & 58.4\suppdelta{-0.4} & 58.7\suppdelta{-0.1} & \textbf{58.8} & 67.7\suppdelta{-1.3} & \textbf{69.0}\suppdelta{+0.0} & 68.9\suppdelta{-0.1} & \textbf{69.0}\\
iSAID & 53.0\suppdelta{-0.2} & 52.8\suppdelta{-0.4} & 52.9\suppdelta{-0.3} & \textbf{53.2} & 65.6\suppdelta{-0.6} & \textbf{66.2}\suppdelta{+0.0} & 65.7\suppdelta{-0.5} & \textbf{66.2}\\
X-Ray & 89.3\suppdelta{-0.4} & \textbf{89.8}\suppdelta{+0.1} & 89.0\suppdelta{-0.7} & 89.7 & 90.2\suppdelta{-0.5} & \textbf{90.7}\suppdelta{+0.0} & 89.9\suppdelta{-0.8} & \textbf{90.7}\\
\midrule
Mean & 61.9\suppdelta{-0.7} & 62.2\suppdelta{-0.4} & 62.4\suppdelta{-0.2} & \textbf{62.6} & 70.7\suppdelta{-1.1} & \textbf{71.8}\suppdelta{+0.0} & 71.5\suppdelta{-0.3} & \textbf{71.8}\\
\bottomrule
\end{tabular*}

\endgroup
\end{table*}

Table~\ref{tab:supp_backbones} compares DINOv2 ViT-L/14 at $896\times896$ and DINOv3 ViT-S/B/L/16 at $1024\times1024$, all producing $64\times64$ tokens.
The controls share the scoring grids, 133,962 episodes, reference sampling, and dataset/shot head settings.
All four backbone configurations apply the positional-subspace removal inherited from INSID3~\cite{insid3}: 500 components for both L backbones, 188 for DINOv3-S ($d=384$), and 375 for DINOv3-B ($d=768$).
The DINOv3 rank follows $r=\operatorname{round}(500d/1024)$, fixed before evaluation, where $d$ is the feature dimension; each basis is computed for its corresponding encoder.
Full retains the same adaptation, RMS, and TAR settings.

For the additional DINOv3-L control, we disable the post-encoding positional projection in Base and Full, retaining L2 normalization and the same initialization, episodes, and inference settings.
The encoder's native positional encoding is unchanged.

\paragraph{Encoder and debiasing effects.}
ReSA improves Base on every dataset and shot setting across all four encoders. Mean gains are 4.9/3.2, 5.8/4.1, and 5.3/3.9 points for DINOv3-S/B/L, and 6.5/6.7 for DINOv2-L (Table~\ref{tab:supp_backbones}). These results show that the adaptation and mask-formation pipeline benefits different feature capacities and encoder generations. With DINOv3-L fixed, positional debiasing adds 2.5/0.5 points. Figure~\ref{fig:supp_debias_method_means} compares the resulting configurations with other methods.

\section{Boundary recovery and threshold alignment}
\label{sec:supp_mechanisms}

\paragraph{Boundary refinement and threshold alignment.}
TAR improves the RCA+RMS coarse mask by 0.7/1.1 mIoU points in 1-shot/5-shot (Table~\ref{tab:supp_tar}). Relative to the no-evidence control, TAR gains 0.2/0.3 points. Relative to unaligned evidence, it gains 0.4 points in 1-shot; both variants round to 71.8 in 5-shot.

Refinement controls share the selected probabilities, coarse masks, boundary bands, pairwise terms, and solver settings. Unaligned refinement uses $\operatorname{logit}(p)$, while the no-evidence control sets both boundary unary scores to zero.

\paragraph{Boundary metric.}
Boundary IoU~\cite{boundaryiou} uses an inner boundary width of $\max(1,\operatorname{round}(0.02\,d))$ pixels, where $d$ is the scoring-image diagonal.
Ignored pixels are excluded from boundary intersections and unions.
Overall scores pool boundary intersections and unions within each class, then average classes, folds, and datasets.
Threshold-deviation bins instead use matched per-episode scores as specified below.

\begin{figure}[!htbp]
\begingroup
\centering
\includegraphics[width=\columnwidth]{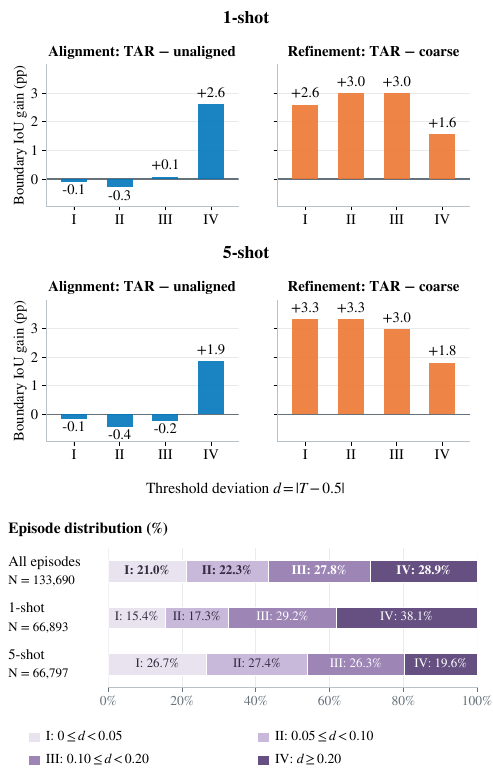}
\UseCaptionStatus{fig:supp_threshold_refinement}
\caption{\textbf{Boundary refinement across threshold deviations.}
The upper two rows show 1-shot/5-shot Boundary IoU gains in percentage points.
Left: TAR minus unaligned CRF; right: TAR minus the coarse mask.
Gains are averaged over matched episodes within each dataset, then over datasets.
The lower panel shows the distribution of 133,690 valid-threshold episodes, normalized within each row.
Groups I--IV share the $d=|T-0.5|$ intervals defined in the legend.}
\label{fig:supp_threshold_refinement}

\endgroup
\end{figure}

\paragraph{Threshold-deviation groups.}
Figure~\ref{fig:supp_threshold_refinement} separates the 1-shot and 5-shot results pooled in main-paper Figure~7. Both use four $|T-0.5|$ groups, averaging matched per-episode Boundary IoUs within each dataset, then equally over represented datasets. The lower panel gives episode proportions for each shot setting and their union.
Probability and threshold diagnostics cover 133,690 episodes; full-benchmark mIoU covers all 133,962. The remaining 272 episodes (88 in 1-shot and 184 in 5-shot) use the empty-candidate or missing-reference-class path in Section~A.6, which produces masks directly.

\paragraph{Alignment gains concentrate at large threshold shifts.}
When $|T-0.5|\geq0.2$, TAR improves Boundary IoU over unaligned refinement by 2.61/1.85 points in 1-shot/5-shot (Fig.~\ref{fig:supp_threshold_refinement}). This group contains 38.1\% of the valid-threshold 1-shot episodes and 19.6\% of the valid-threshold 5-shot episodes. TAR aligns the refinement evidence with the coarse-mask cutoff, whereas unaligned evidence retains the $0.5$ cutoff. Across all 133,690 diagnostic episodes, class-pooled Boundary IoU is 40.44/45.97 for the coarse masks, 42.32/48.86 for unaligned refinement, and 42.64/48.69 for TAR.

\setcounter{dbltopnumber}{1}
\section{Inference cost and benchmark sensitivity}
\label{sec:supp_inference_efficiency}

\paragraph{Measurement protocol.}
Runtime measurements use one H100 GPU and one process, with 100 episodes per shot: ten from each dataset. Timings cover complete inference, including both ReSA candidates and final refinement.

\paragraph{Candidate count and inference cost.}
Two branches achieve 62.6/71.8 mean mIoU at 0.205/0.420 seconds per episode (Table~\ref{tab:supp_branch_timing}). Adding $\lambda=2$ changes mIoU to 62.6/71.7 and runtime to 0.210/0.450 seconds. Feature extraction and reference fitting are shared; each additional branch repeats only lightweight adaptation and threshold search.

\begin{table}[!tbp]
\centering
\UseCaptionStatus{tab:supp_branch_timing}
\caption{\textbf{Complete-pipeline accuracy and time by branch count.} mIoU (\%) uses all 133,962 benchmark episodes; time (s/episode) uses the same 200 inputs and measurement protocol in Section~\ref{sec:supp_inference_efficiency}, on an H100. All four configurations include TAR. Bold marks the highest displayed mIoU per shot.}
\label{tab:supp_branch_timing}
\fontsize{8.5}{10}\selectfont
\setlength{\tabcolsep}{2pt}
\renewcommand{\arraystretch}{1.12}
\begin{tabular*}{\columnwidth}{@{\extracolsep{\fill}}lcccc@{}}
\toprule
\multirow[c]{2}{*}[-2.7pt]{\textbf{Branch setting}} & \multicolumn{2}{c}{\textbf{1-shot}} & \multicolumn{2}{c}{\textbf{5-shot}}\\
\cmidrule(lr){2-3}\cmidrule(lr){4-5}
 & mIoU & Time & mIoU & Time\\
\midrule
Single: $\lambda=1$ & 62.3 & 0.176 & 71.3 & 0.396\\
Single: $\lambda=3$ & 61.7 & 0.177 & 71.7 & 0.384\\
Dual: $\lambda\in\{1,3\}$ & \textbf{62.6} & 0.205 & \textbf{71.8} & 0.420\\
Three: $\lambda\in\{1,2,3\}$ & \textbf{62.6} & 0.210 & 71.7 & 0.450\\
\bottomrule
\end{tabular*}
\end{table}

\begin{table}[!htbp]
\begingroup
\centering
\UseCaptionStatus{tab:supp_inference_timing}
\caption{\textbf{Measured inference time (s/episode).} H100 measurements use 100 episodes per shot, ten from each dataset, with complete inference timed as specified in Section~\ref{sec:supp_inference_efficiency}. Groups distinguish offline task training and test-time optimization. Bold marks the lowest time within each group and shot.}
\label{tab:supp_inference_timing}
\fontsize{8.5}{10}\selectfont
\setlength{\tabcolsep}{4pt}
\renewcommand{\arraystretch}{1.12}
\begin{tabular*}{\columnwidth}{@{\extracolsep{\fill}}lrr@{}}
\toprule
\textbf{Method} & \textbf{1-shot} & \textbf{5-shot} \\
\midrule
\multicolumn{3}{@{}l}{\textit{Offline task training}} \\
\hspace{2pt}FSSAM & 0.100 & 0.167\\
\hspace{2pt}SINE & 0.300 & 0.830\\
\hspace{2pt}SegIC & 0.267 & 1.281\\
\hspace{2pt}COSINE & 0.502 & 1.068\\
\hspace{2pt}LDIS-1 & \textbf{0.060} & \textbf{0.161}\\
\midrule
\multicolumn{3}{@{}l}{\textit{No task training or test-time optimization}} \\
\hspace{2pt}INSID3 & 0.397 & 0.643\\
\hspace{2pt}FROST & 0.264 & 1.807\\
\hspace{2pt}RSRM & \textbf{0.087} & \textbf{0.307}\\
\hspace{2pt}PerSAM & 0.519 & 1.610\\
\hspace{2pt}RPG-SAM & 0.468 & 1.892\\
\midrule
\multicolumn{3}{@{}l}{\textit{Test-time adaptation / optimization}} \\
\hspace{2pt}HERA & 1.616 & 3.214\\
\hspace{2pt}RNS & 2.802 & 9.429\\
\hspace{2pt}PR-MaGIC & 4.531 & 9.112\\
\hspace{2pt}\textbf{\method{} (ours)} & \textbf{0.205} & \textbf{0.420}\\
\bottomrule
\end{tabular*}

\endgroup
\end{table}

\paragraph{Comparison groups.}
Table~\ref{tab:supp_inference_timing} separates methods with offline segmentation-task training, reference-conditioned inference without gradient-based test-time optimization, and test-time adaptation or optimization. The last group includes parameter fitting and gradient-based feature refinement with frozen model weights. Method-specific inference configurations are given in Section~D.

\paragraph{Sensitivity to benchmark composition.}
\label{sec:supp_bootstrap}
Paired dataset-bootstrap intervals are positive for all five 1-shot comparisons in Table~\ref{tab:supp_bootstrap}. In 5-shot, the Full-versus-Base and RCA-versus-$R$-only intervals remain positive, while the threshold-fitting and alignment intervals include zero.

We use 20,000 bootstrap draws of ten datasets with replacement (seed 20261108). Each draw averages unrounded paired mIoU differences; the 2.5th and 97.5th percentiles define the 95\% interval. Point gains use the displayed means.

\paragraph{Compared settings.}
\textit{Full vs. Base}: complete ReSA versus $p_0$ at $T=0.5$.
\textit{RCA vs. $R$-only}: adding constraints and entropy to reference supervision (matched budgets, $T=0.5$, no RMS/TAR; RCA: $\lambda=1$).
\textit{Fitted $T$ vs. $0.5$}: fitted versus fixed cutoff on the same map.
\textit{Gain vs. min. $D$}: relative-gain versus minimum-error selection with fixed candidates and thresholds. Both omit TAR.
\textit{TAR vs. unaligned}: changing only evidence alignment.

\begin{table}[!htbp]
\begingroup
\centering
\UseCaptionStatus{tab:supp_bootstrap}
\caption{\textbf{Paired benchmark-bootstrap intervals.} Mean mIoU gain in points, followed by its 95\% percentile interval. All pairs use matched configurations and episodes.}
\label{tab:supp_bootstrap}
\fontsize{8.5}{10}\selectfont
\setlength{\tabcolsep}{2pt}
\renewcommand{\arraystretch}{1.12}
\begin{tabular*}{\columnwidth}{@{\extracolsep{\fill}}lcc@{}}
\toprule
Comparison & 1-shot & 5-shot\\
\midrule
Full vs. Base & $+5.3\;[+4.1,+6.5]$ & $+3.9\;[+3.1,+4.6]$\\
RCA vs. $R$-only & $+3.2\;[+2.4,+4.2]$ & $+2.0\;[+1.4,+2.4]$\\
Fitted $T$ vs. $0.5$ & $+1.6\;[+0.6,+2.5]$ & $+0.0\;[-0.7,+0.6]$\\
Gain vs. min. $D$ & $+0.9\;[+0.5,+1.2]$ & $+0.2\;[+0.0,+0.4]$\\
TAR vs. unaligned & $+0.4\;[+0.2,+0.6]$ & $+0.0\;[-0.2,+0.1]$\\
\bottomrule
\end{tabular*}

\endgroup
\end{table}

\newpage
\section{Further visual results}

\label{sec:supp_visuals}
Figures~\ref{fig:supp_module_progression}--\ref{fig:supp_comparisons} show progressive module predictions, RCA loss controls, matched boundary refinements, and comparisons with other methods, respectively. The visualizations use recorded outputs from the configuration in Section~\ref{sec:supp_inference}. One annotated reference is displayed for each query.
Probability maps share the $[0,1]$ scale. Cyan and magenta fills denote annotations and predictions, respectively, with white contours. IoUs are evaluated on the full query, excluding ignored pixels; cropped comparisons use the same region across methods.

\paragraph{Progressive module ablation.}
Figure~\ref{fig:supp_module_progression} follows the same query through adaptation, mask selection, and refinement. Base thresholds the first-stage reference initialization $p_0$ at $0.5$; RCA uses the adapted $\lambda=1$ branch at the same cutoff. RMS then fits thresholds and selects between the two adapted branches; TAR refines the selected coarse mask.

\clearpage
\begin{figure*}[!tp]
\phantomcaption
\label{fig:supp_module_progression}
\begingroup\centering
\includegraphics[page=1,width=0.988\textwidth]{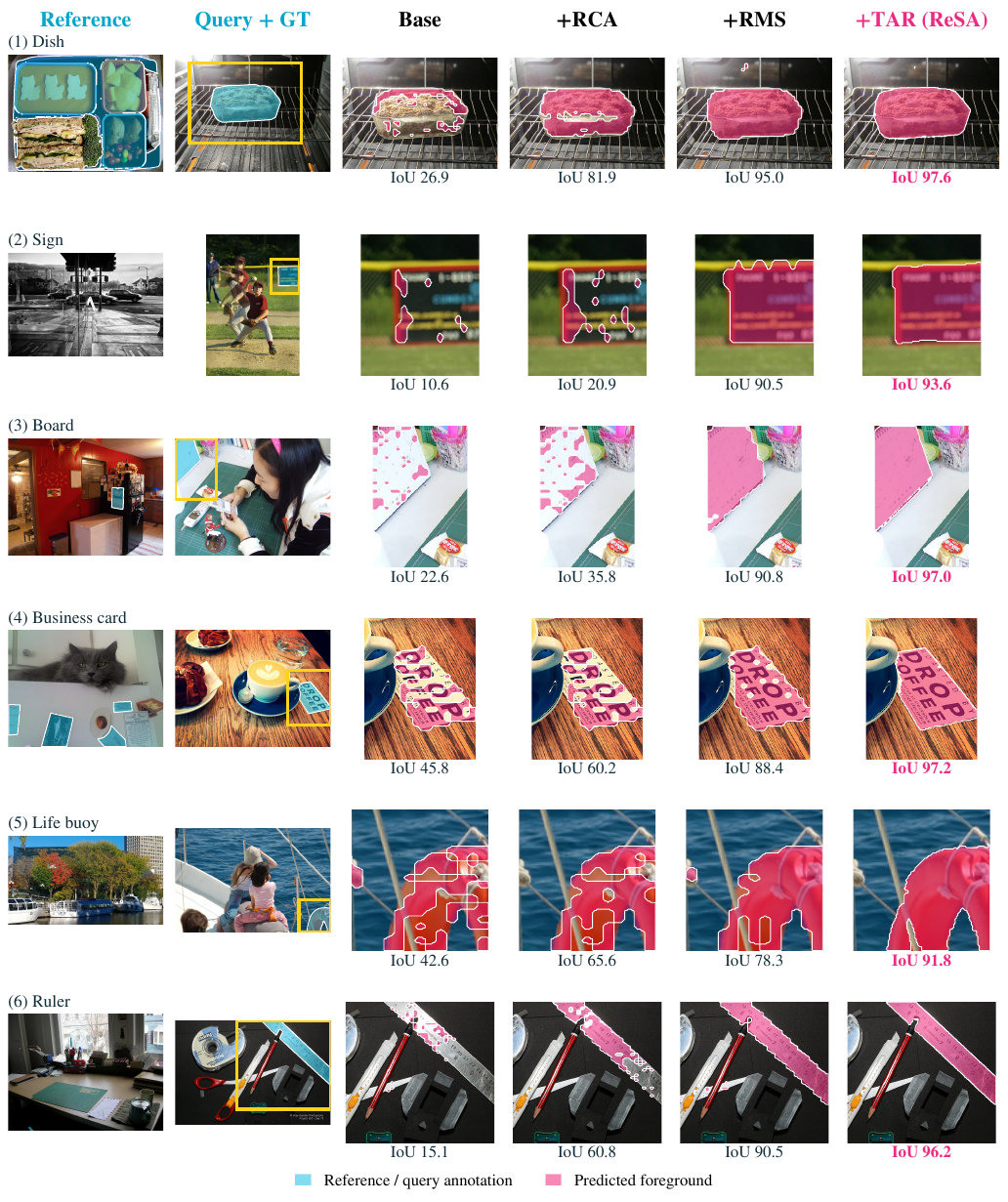}\par
\endgroup
\end{figure*}

\begin{figure*}[!p]
\ContinuedFloat
\begingroup\centering
\includegraphics[page=2,width=0.988\textwidth]{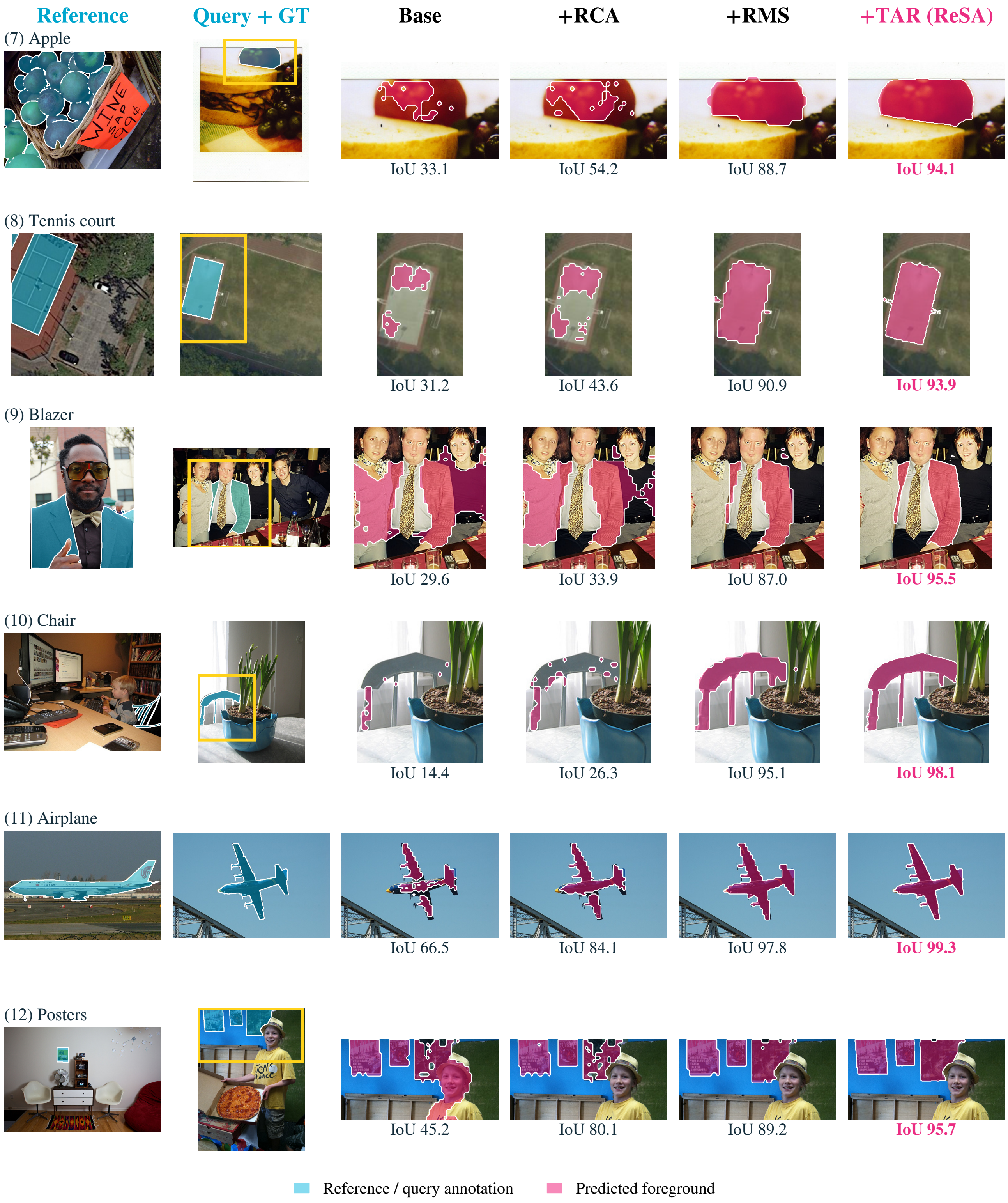}\par
\endgroup
\UseCaptionStatus{fig:supp_module_progression}
\caption{\textbf{Progressive effects of ReSA components.}
RCA recovers target regions missed by the initial prediction; RMS improves foreground selection; TAR refines the resulting contours. The successive predictions show how these stages contribute to more complete target masks and fewer extraneous regions. The figure presents 12 examples. Yellow boxes indicate regions enlarged consistently across prediction columns; IoUs are measured on the full query.}
\end{figure*}

\begin{figure*}[!p]
\phantomcaption
\label{fig:supp_ja_examples}
\begingroup\centering
\includegraphics[page=1,width=\textwidth]{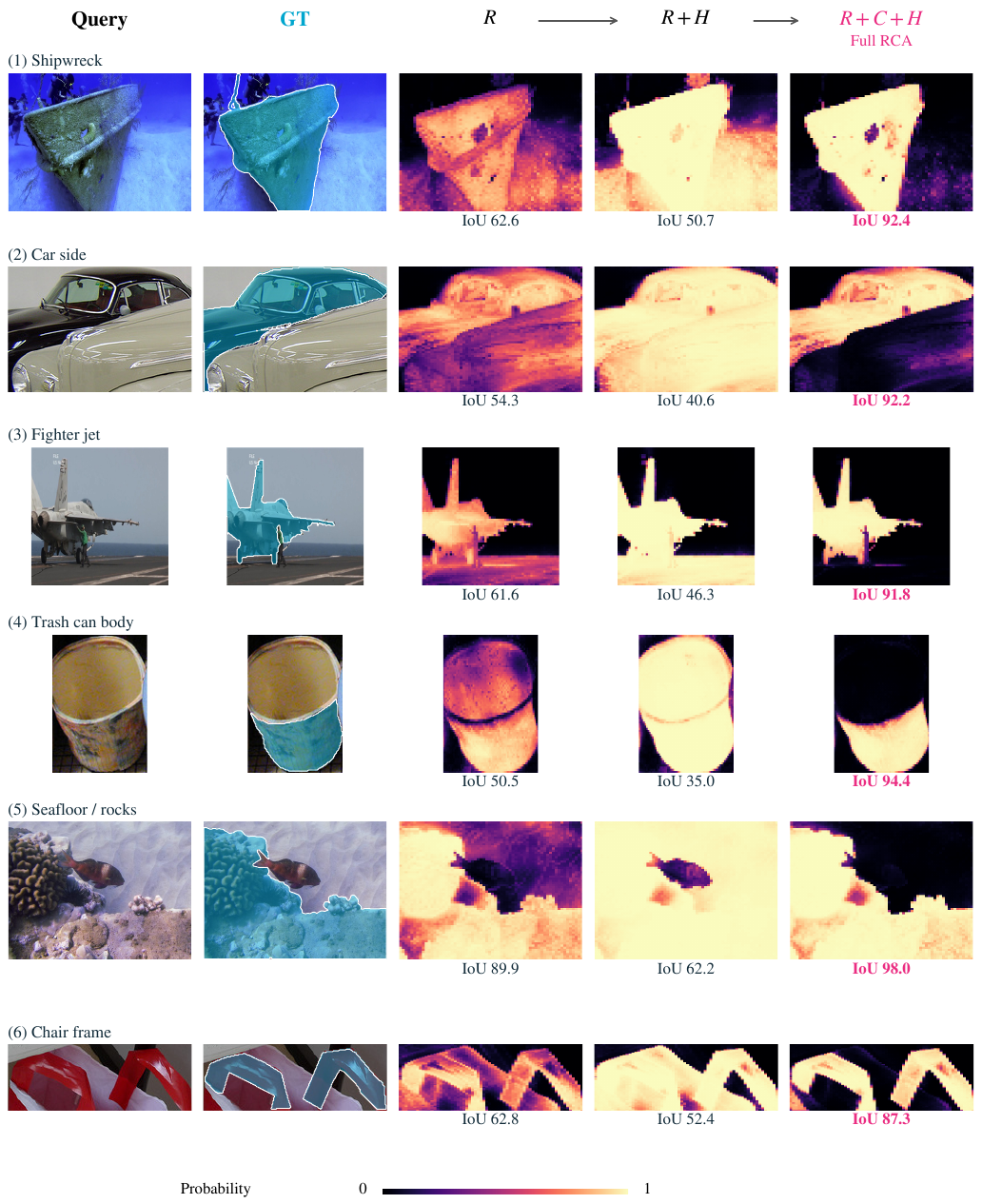}\par
\endgroup
\end{figure*}

\begin{figure*}[!p]
\ContinuedFloat
\begingroup\centering
\includegraphics[page=2,width=\textwidth]{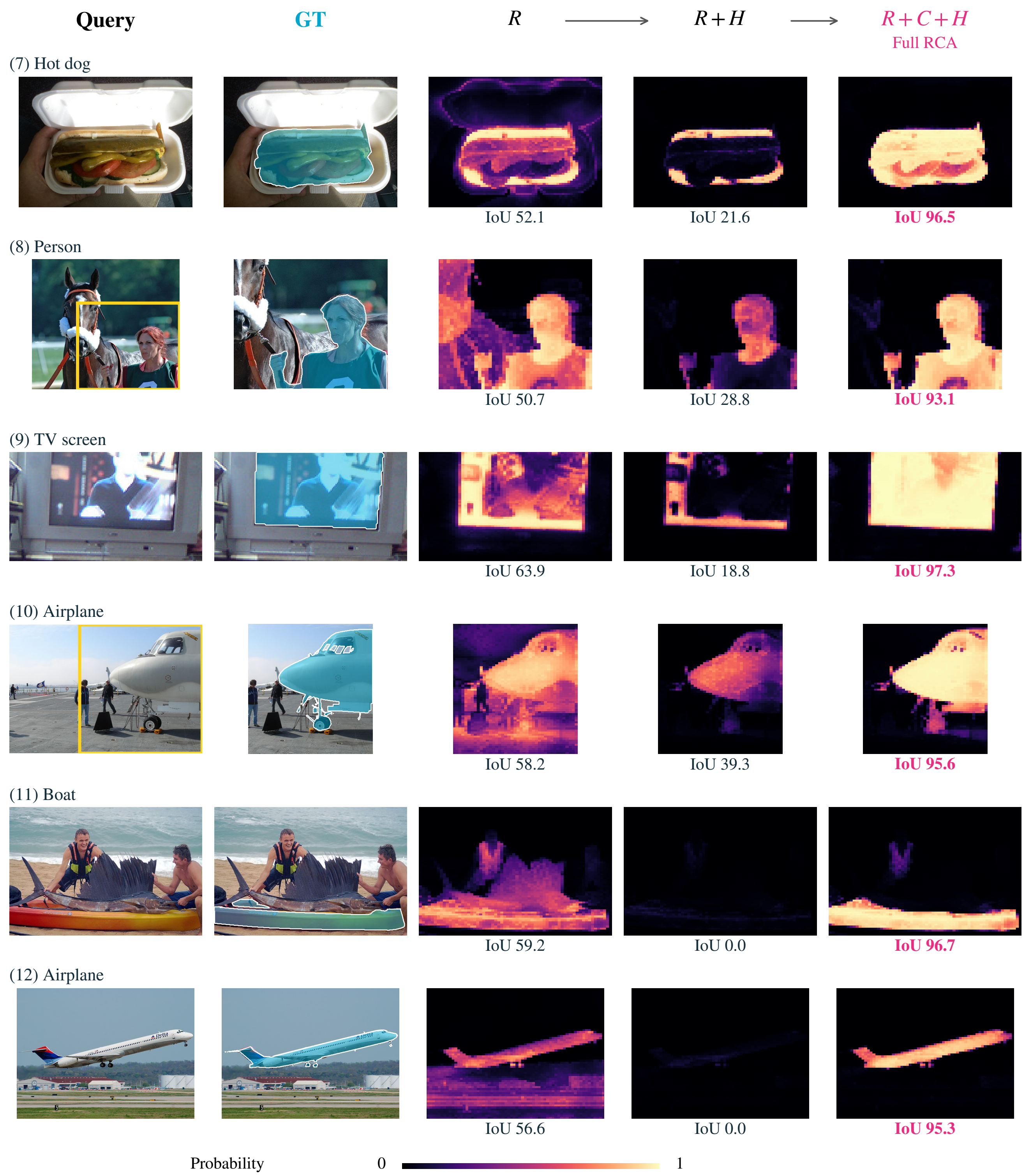}\par
\endgroup
\UseCaptionStatus{fig:supp_ja_examples}
\caption{\textbf{Response constraints guide entropy-based adaptation.}
Adding the entropy term $H$ to reference supervision can sharpen responses toward either foreground or background, amplifying distracting regions or suppressing target regions. The response constraint $C$ guides this sharpening to retain the reference-specified foreground and suppress distracting responses. The figure presents 12 comparisons of $R$, $R+H$, and $R+C+H$, using the same initialization and optimization budgets, with $\lambda=1$, $T=0.5$, and no RMS or TAR. Probability maps share a common color scale; yellow boxes indicate the enlarged regions.}
\end{figure*}

\begin{figure*}[!p]
\phantomcaption
\label{fig:supp_tar_crops}
\begingroup\centering
\includegraphics[page=1,width=\textwidth]{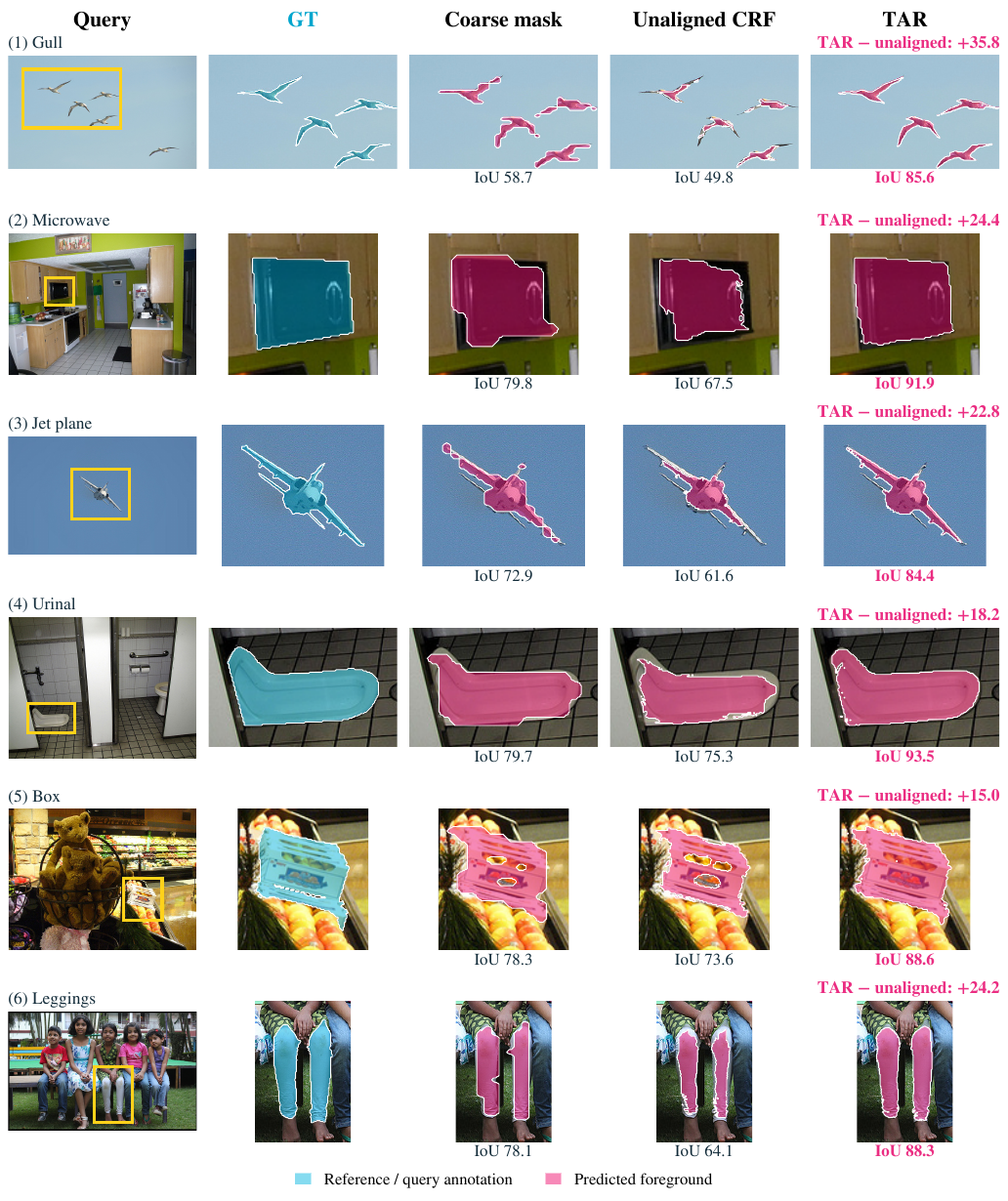}\par
\endgroup
\end{figure*}

\begin{figure*}[!p]
\ContinuedFloat
\begingroup\centering
\includegraphics[page=2,width=\textwidth]{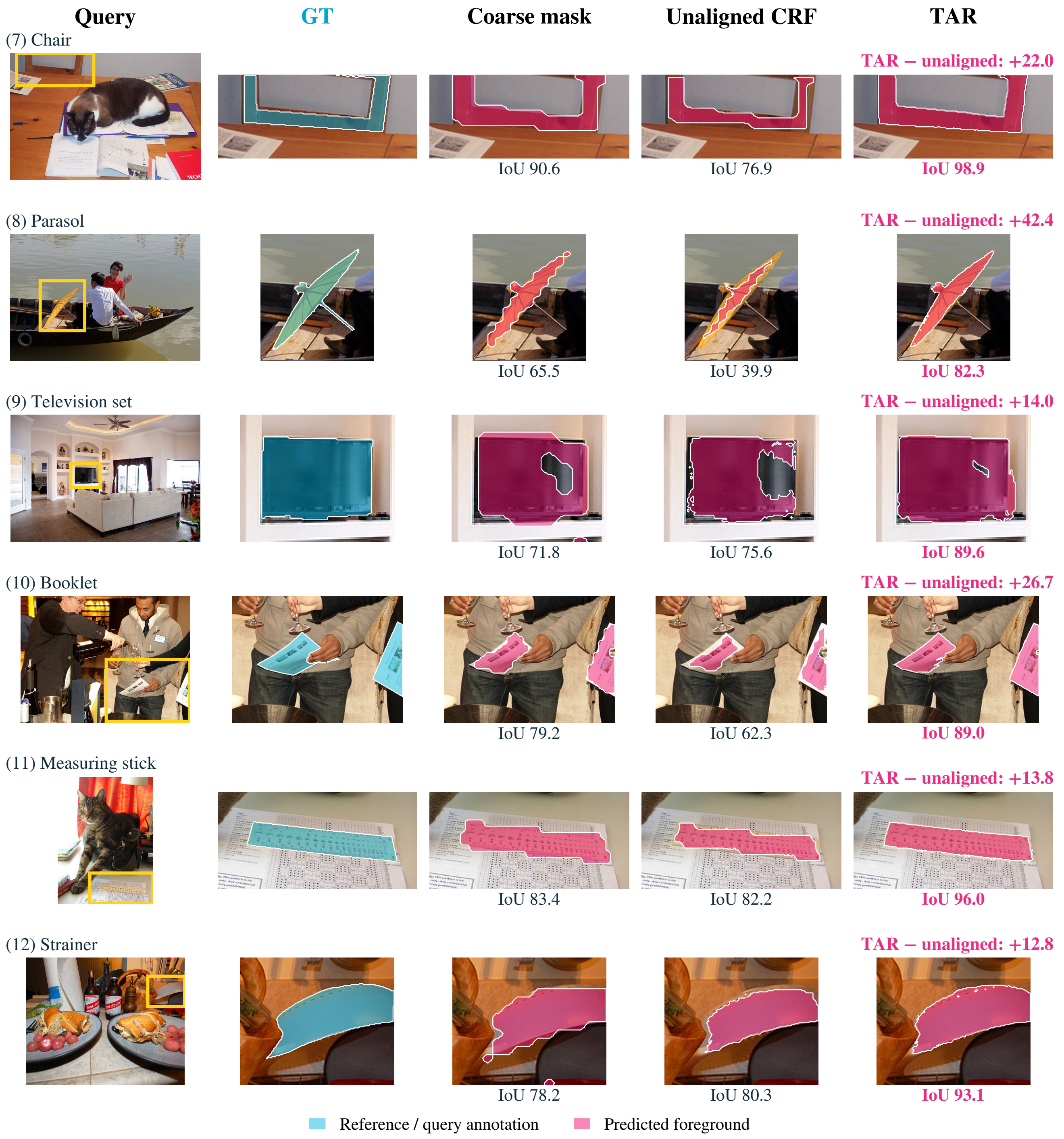}\par
\endgroup
\UseCaptionStatus{fig:supp_tar_crops}
\caption{\textbf{Threshold alignment improves boundary refinement.}
Aligning the refinement scores with the selected threshold helps preserve target structures that unaligned CRF removes or distorts, improving the resulting boundaries. Matched crops highlight these differences at thin structures and object contours. The figure presents 12 examples, with both refiners using the same input probabilities and coarse mask. Yellow boxes indicate the enlarged regions; the reported gain is the full-image IoU difference between TAR and unaligned CRF.}
\end{figure*}

\begin{figure*}[!p]
\phantomcaption
\label{fig:supp_comparisons}
\label{fig:supp_comparison_one}
\begingroup\centering
\includegraphics[width=\textwidth]{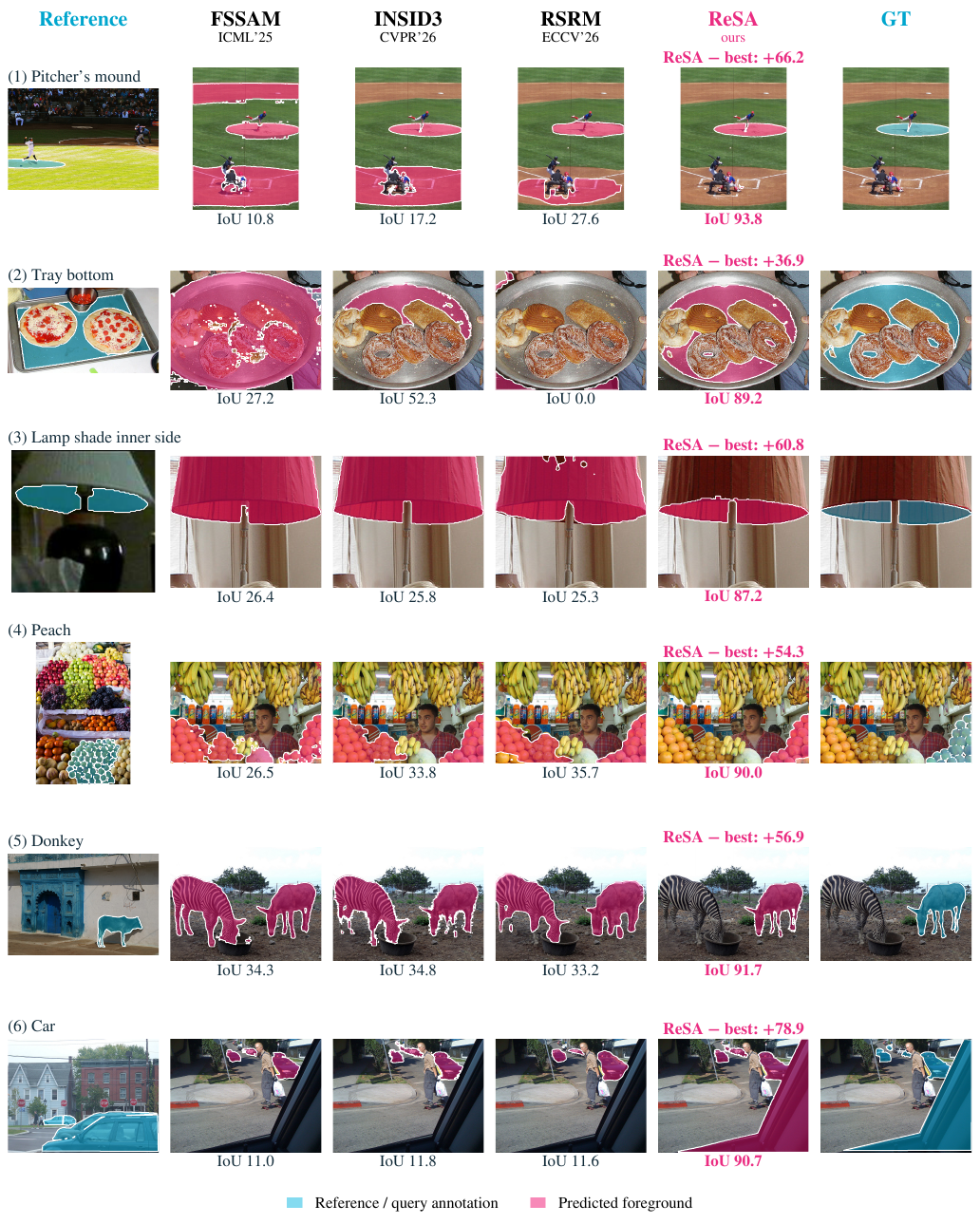}\par
\endgroup
\end{figure*}

\begin{figure*}[!p]
\ContinuedFloat
\phantomcaption
\label{fig:supp_comparison_two}
\begingroup\centering
\includegraphics[width=\textwidth]{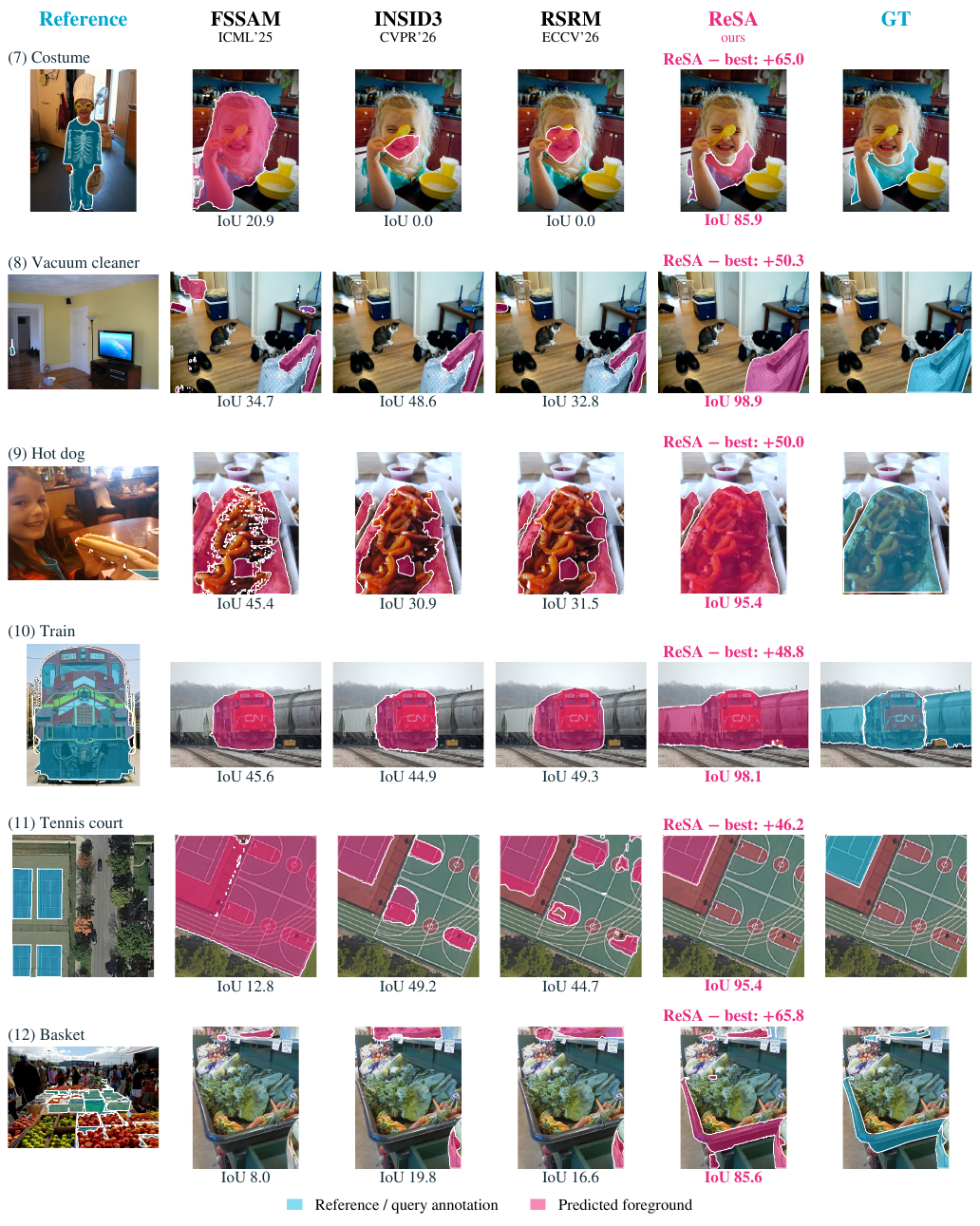}\par
\endgroup
\end{figure*}

\begin{figure*}[!p]
\ContinuedFloat
\phantomcaption
\label{fig:supp_comparison_three}
\begingroup\centering
\includegraphics[width=\textwidth]{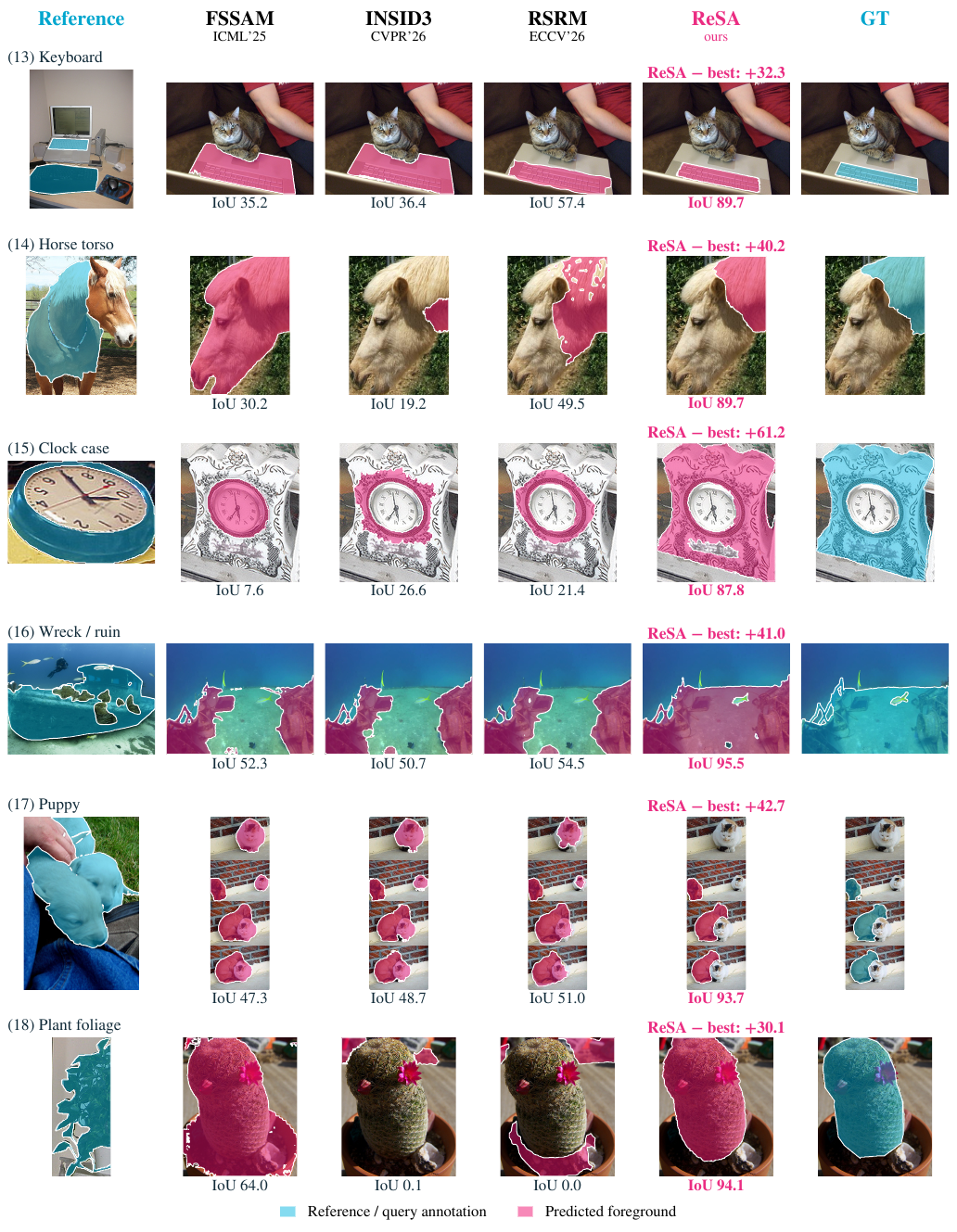}\par
\endgroup
\end{figure*}

\begin{figure*}[!p]
\ContinuedFloat
\begingroup\centering
\includegraphics[width=\textwidth]{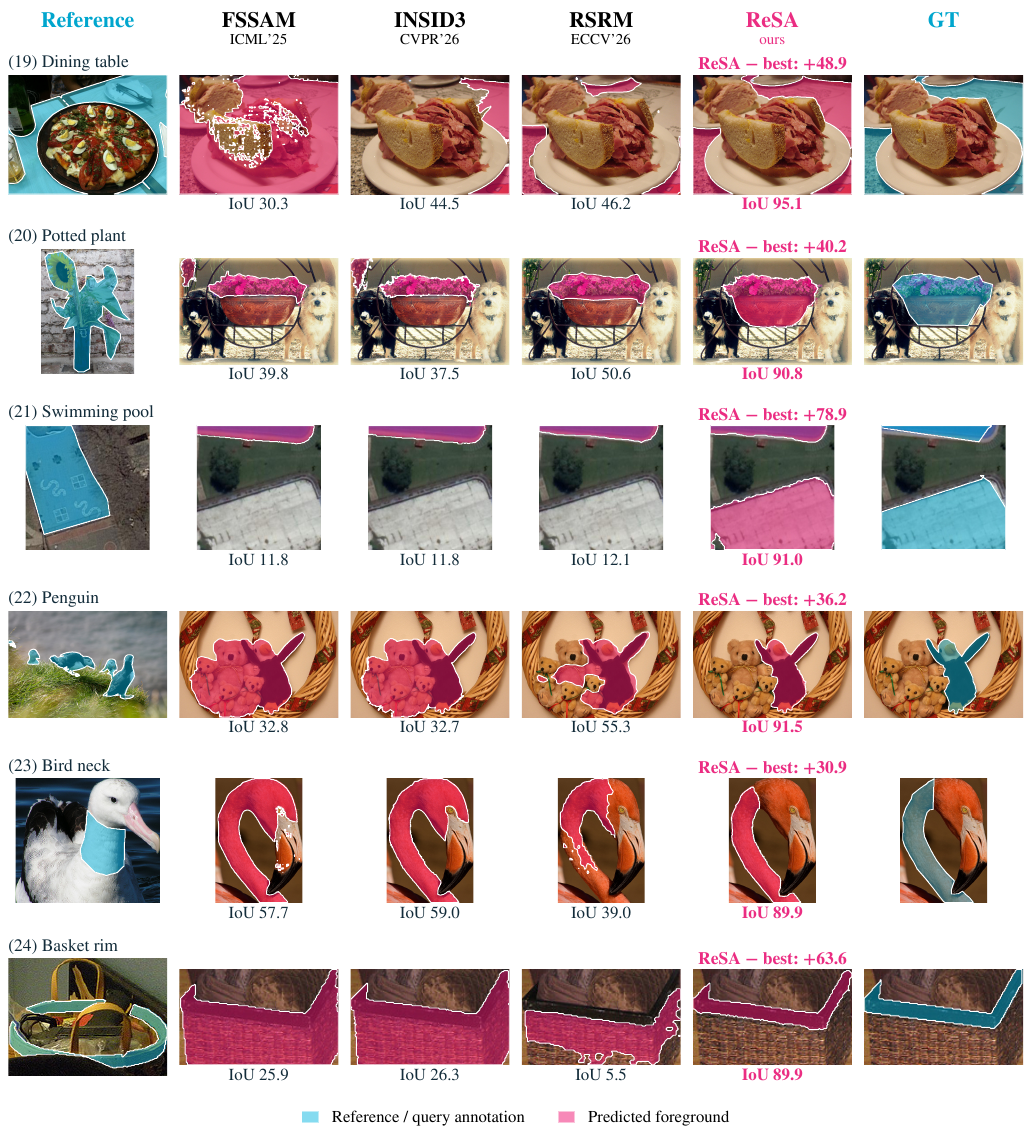}\par
\endgroup
\UseCaptionStatus{fig:supp_comparisons}
\captionsetup{skip=8pt}
\caption{\textbf{Qualitative comparisons with methods with and without offline task training.}
The comparisons show how ReSA retains the reference-specified target extent while rejecting nearby objects and surrounding regions. Whole-object and part examples highlight the role of reference guidance in distinguishing the desired foreground from visually similar content. The figure presents 24 five-shot examples from seven datasets. All methods receive the same query and five annotated references; one reference is shown. ``ReSA $-$ best'' reports the IoU gain over the highest-scoring method among FSSAM, INSID3, and RSRM for each example.}
\label{fig:supp_comparison_four}
\end{figure*}

\end{document}